\documentclass[11pt,twoside]{article}

\usepackage{fullpage}

\input{command-template}

\begin{document}

\begin{center}

  {\bf{\LARGE{Next-token functional estimation}}}

\vspace*{.2in}

{\large{
\begin{tabular}{ccc}
Milind Nakul$^{\star}$, Vidya Muthukumar$^{\dagger,\star}$, Ashwin Pananjady$^{\star, \dagger}$
\end{tabular}
}}
\vspace*{.2in}

\begin{tabular}{c}
Schools of $^\star$Industrial and Systems Engineering and $^\dagger$Electrical and Computer Engineering \\
Georgia Institute of Technology
\end{tabular}

\vspace*{.2in}

\today

\vspace*{.2in}

\begin{abstract}
Suppose we observe the first $n$ points of a sequence of random variables having length $n+1$, and wish to estimate a functional of the unobserved final point and the empirical measure of the $n$ observed training points. Such \emph{next-token functionals} include the probability that the next token is novel (also known as the surprise probability), the tail probability of the minimum distance between the next token and training points, and the test error of a classifier trained on the observed points. All of these quantities are classically estimated by the leave-one-out method, which is inconsistent under temporal dependence. We propose a \emph{leave-a-window-out} estimator, which deletes a window of length $\tau$ after each index before forming the empirical measure and reduces to leave-one-out at $\tau = 1$. Under natural assumptions, we show that the error of our estimator decays at a parametric rate for any stationary $\beta$-mixing process that also admits a Marton coupling. Our results thus cover several natural functionals on a large class of stochastic processes.
We complement these upper bounds with a sharp minimax lower bound for estimating the surprise probability on mixing Markov chains. Simulations on Markov chains, moving-average processes, and autoregressive processes show that our estimator succeeds in many scenarios where leave-one-out and add-constant baselines fail.
\end{abstract}
\end{center}

\section{Introduction}

Let
\(
X^{n+1} = (X_1,\ldots,X_n,X_{n+1})
\)
denote a sequence of random variables, each of which takes values in a space \(\Xcal\). Suppose only the first \(n\) random variables are observed and the final point \(X_{n+1}\) (or next token) is unobserved. 
We consider the problem of estimating \emph{next-token functionals}
\begin{align} \label{eq:NT-intro}
\EE\bigl[f(X_{n+1};\measure_n)\bigr],
\end{align}
where \(\measure_n\) is the empirical measure of the observed trajectory \(X^n\). The function \(f\) can be chosen so as to recover classical quantities such as discovery probabilities, sample coverage, and the predictive performance of learning algorithms.

Estimating next-token functionals is a basic problem in statistics and machine learning. It arises whenever the first \(n\) observations serve as a training sample and the final point plays the role of a test observation. 
Such problems therefore appear in a broad range of applications spanning learning theory~\citep{bousquet2002stability}, ecology~\citep{fisher1943relation}, genomics~\citep{favaro2012new,lijoi2007bayesian}, and natural language processing~\citep{church1991probability,chen1999empirical,ney1994structuring}. To set the stage, it is useful to see two concrete examples:

\paragraph{Example 1: Surprise probability in a stochastic process.} Let $X^{n+1}$ be a trajectory from a stochastic process. The \emph{surprise probability} is the probability that the $(n+1)$-th point has not been seen before, i.e., $\Prob(X_{n+1} \notin \{X_{1}, \ldots, X_n\})$. This quantity has been 
studied for ergodic Markov chains~\citep{norris2014surprise}, and depends on the joint distribution of the entire trajectory. It is a next-token functional of the form~\eqref{eq:NT-intro}, obtained by taking $f(X_{n+1};\measure_n) = \ind{X_{n+1} \notin \mathsf{supp}(\mathsf{dP}_n)}$. Here we denote the support of a measure $\mu$ as $\text{supp}(\mu)$. 

\vspace{-4mm}

\paragraph{Example 2: Error on a future ``test" point.} Suppose we have a trajectory of covariates and labels denoted as $Z^{n+1} = (Z_1,\ldots,Z_{n+1})$ where $Z_i = (X_i,Y_i)$, the $X_i$'s are the covariates and the $Y_i$'s are the labels. 
We train a classifier on the first $n$ points and wish to quantify the test error at $Z_{n+1}$. This is again a next-token functional of the form~\eqref{eq:NT-intro} if we take $f$ to be the loss incurred by the classifier, and the training algorithm is invariant to the labeling of the data points (and so depends on the training data $Z^n$ only through its empirical measure). 

\medskip

A large body of work has studied the question of next-token functional estimation in the i.i.d.\ setting. In this case, the entire sample is symmetric, and so estimation is often carried out using ideas based on the leave-one-out principle. In brief, the main idea of this principle is to estimate the functional~\eqref{eq:NT-intro} using just the $n$ training points with 
\begin{align} \label{eq:LOO}
\frac{1}{n} \sum_{i = 1}^n f(X_i, \mathsf{dP}_{-i}).
\end{align}
Here $X_i$ is treated as the stand-in for the unobserved test point, and $\mathsf{dP}_{-i}$ denotes the empirical measure of the $n - 1$ training points with the $i$-th point left out.
In the above two examples, the leave-one-out estimator~\eqref{eq:LOO} is given by the Good--Turing estimator~\citep{good} for surprise probability (which is equivalent to the expected ``missing mass" in the i.i.d. setting) and the leave-one-out or ``jackknife" estimator for predictive risk~\citep{rogers1978finite,vapnik2006estimation}. The Good--Turing estimator has been analyzed extensively in the i.i.d.\ setting, beginning with the pioneering work of~\citet{mcallester2000convergence}, while leave-one-out error estimates for stable learning algorithms have also been the subject of substantial study~\citep{bousquet2002stability,feldman2018generalization,feldman2019high,bousquet2020sharper}.

The main motivation of this paper is that in many modern applications, observations are not independent. Temporal dependence is intrinsic in sequential data such as text, biological sequences, and time series, where successive observations may be strongly correlated. In such settings, estimators designed for i.i.d.\ samples can fail: for instance, the naive use of Good--Turing is no longer valid even for Markovian data~\citep{chandra2021good}. More broadly, temporal dependence changes both the statistical difficulty of the problem and the algorithmic structure of good estimators, necessitating the use of new tools that explicitly account for local correlations~\citep{hao2018learning,skorski2020missing,chandra2021good}.

Classical work on resampling under temporal dependence recognized that observation-wise leave-one-out procedures must be modified to account for local correlations. In particular, \citet{kunsch1989jackknife} introduced the block jackknife and block bootstrap procedures for stationary observations, replacing the deletion or resampling of individual observations by that of
contiguous blocks. 
Other such classical works include those by~\citet{liu1992moving} and~\citet{politis1991circular}.
In the specific context of the two examples discussed above, \citet{pananjady2024just} introduced a ``leave-a-window-out" version of Good--Turing for estimating the stationary missing mass of a Markov chain and showed that it achieves minimax-optimal mean-squared error. Related ideas have also been used for estimating the stationary distribution of a mixing process~\citep{pmlr-v291-nakul25a}. On a parallel front, \citet{fu2023sharper} studied the estimation of predictive accuracy for learning algorithms under mixing assumptions on the stochastic process, while \citet{chatterjee2024generalization} obtained generalization bounds via online-to-batch conversion without relying on uniform stability. Another recent example includes the application of these ideas to predictive inference~\citep{jiang2026leave}. In contrast to these specific studies, our goal is to provide a general treatment of the next-token functional estimation problem under general assumptions on the function $f$ and the stochastic process. 

\subsection{Contributions and organization}

Our main contributions are summarized below:
\begin{itemize}
    \item \textbf{Estimator.} We propose a general estimator (Eq.~\eqref{eq:estimator}) for next-token functionals of dependent data, which leaves \emph{windows} of data out to compute proxies for the empirical measure, instead of just a single point as in Eq.~\eqref{eq:LOO}.
    The estimator is designed to mitigate short-range dependence, and extends the windowed Good--Turing philosophy of~\citet{pananjady2024just} beyond stationary missing-mass estimation. We recover the classical leave-one-out estimator for the specific choice of window size $\tau = 1$.

    \item \textbf{General upper bound.} In Theorem~\ref{theorem:MSE_theorem}, we establish a mean-squared error bound for the proposed estimator under two conditions on the functional and the estimator: out-of-sample deletion stability (Assumption~\ref{assmp:stability}) and a suitable bounded-differences property (Assumption~\ref{assmp:bd}). Our bound cleanly isolates bias from variance, and each term is controlled by a different mechanism. In particular, the bias is controlled provided the stochastic process is suitably mixing, and
    we control variance using a McDiarmid-style concentration inequality for dependent data based on Marton couplings~\citep{paulin2015concentration}.

    \item \textbf{Concrete functionals.} In Section~\ref{sec:f_example}, we specialize this bound to four functionals that are usually studied separately, covering the two examples above. These include the surprise probability, count surprise, nearest-neighbor tail probability, and classification test error. In all cases, our bound produces parametric rates of estimation under natural assumptions on the stochastic process that cover canonical assumptions under which these specific problems have been studied.

    \item \textbf{Matching lower bound.} In Proposition~\ref{thm:minimax_risk} we prove that no estimator of the surprise probability can achieve risk smaller than order $\Tmix/n$ over the class of stationary finite-state ergodic Markov chains with mixing time at most $\Tmix$. In conjunction with Corollary~\ref{corollary:surprise_probability}, which shows that our estimator attains this rate, our lower bound showcases the minimax optimality of the leave-a-window-out principle.
\end{itemize}

The rest of the paper is organized as follows. Notation is presented in Section~\ref{sec:notation}. Section~\ref{sec:background} formalizes the estimation problem, including our assumptions on the stochastic process. Section~\ref{sec:estimator} explains why leave-one-out fails under dependence and introduces the leave-a-window-out estimator. We also state our assumptions on $f$ as well as our main upper bound, Theorem~\ref{theorem:MSE_theorem}. Section~\ref{sec:f_example} presents corollaries for four concrete functionals. We also present the minimax lower bound for estimating surprise probability in Section~\ref{sec:surprise}.  Section~\ref{sec:numerical_results} reports numerical results and Section~\ref{sec:proof_MSE_theorem} proves Theorem~\ref{theorem:MSE_theorem}. Section~\ref{sec:discussion} discusses implications and open questions. The remaining proofs---of corollaries and lower bounds---are collected in Appendix~\ref{sec:proofs}. Further examples and numerical experiments 
are presented in the remaining appendices.

\subsection{Notation} \label{sec:notation}
We use $\Pcal(\Xcal)$ to denote the set of all probability measures on $\Xcal$ (atoms are allowed). For any $x\in \Xcal$ we use $\delta_x$ to denote a distribution which places all its mass on the point $x$. We use $\Delta(S)$ to denote the set of all probability mass functions (PMFs) on a finite set $S$. For a sequence $Z \in \Xcal^{\mathbb{N}}$ and $x \in \Xcal$, we let $N_x(Z) = \#\{i: Z_i = x \}$ denote the number of occurrences of $x$ in $Z$. We frequently use the shorthand $N_x = N_x(Z)$. For two real numbers $a$ and $b$, let $a \wedge b = \min\{a, b\}$ and $a \vee b = \max\{a, b\}$. Let $[n]$ denote the set of natural numbers less than or equal to $n$. For an index set $P \subseteq [n]$, we use two related objects, distinguished by their typeface: the \emph{sequence} $X_P = (X_i)_{i \in P}$, ordered canonically, and the \emph{multiset} $\bX_P = \{X_i\}_{i \in P}$. Boldface thus indicates that only the collection of values matters and not the order in which they appear; in particular, $\bX_{[n]}$ is the set of all random variables occurring in the trajectory $X^n$. We write membership statements in terms of the set and reserve the sequence for those quantities that depend on the ordering, such as total variation distances between trajectories. Occurrence counts are insensitive to this choice, since $N_x(X_P) = N_x(\bX_P)$ for every $x \in \Xcal$. 
We use the notation $f(u) \lesssim g(u)$ to mean that there exists some absolute positive constant $C$ that is independent of all problem parameters, such that $f(u) \le C \cdot g(u)$ for all $u$ in the domain of $f$ and $g$. We use the notation $f(u) \gtrsim g(u)$ when $g(u) \lesssim f(u)$. We write $f(u) \asymp g(u)$ if both relations $f(u) \gtrsim g(u)$ and $g(u) \gtrsim f(u)$ hold. Logarithms are taken to the base $e$. We use $(c, C)$ to denote universal positive constants that could be different in each instantiation. For two probability measures $\mu,\nu$, sharing $\sigma$-field $\mathcal{F}$, their total variation distance is
\[
\TV(\mu,\nu)
:=
\sup_{A\in \mathcal{F}} |\mu(A)-\nu(A)|.
\]
If $\mu, \nu$ are PMFs, then this is equal to half the $\ell_1$ distance between them.
For a matrix $\Gamma$ we denote its operator norm by $\opnorm{\Gamma}$.

\section{Background and problem formulation}\label{sec:background}

In this section, we formally define our estimation problem
as well as assumptions on our stochastic process.

\subsection{Next-token functionals}\label{sec:joint_functionals}
Denote the estimand by
\begin{align}\label{eq:joint_functional}
\estimand_f \defn \EE\big[f(X_{n+1};\measure_n)\big],
\end{align}
where \(\measure_n \defn \frac{1}{n}\sum_{i\in [n]}\delta_{X_i}\) denotes the empirical measure of \(X^n\), and \(f: \Xcal \times \Pcal(\Xcal) \to \mathbb{R}\) is a known real-valued function that takes as input a point \(x \in \Xcal\) and a probability measure on the state space $\Xcal$. Throughout the paper we assume that $f$ is uniformly bounded, with sup norm $\|f\|_{\infty}$. The expectation in \eqref{eq:joint_functional} is taken with respect to the joint distribution of $X^{n+1}$. As mentioned before, the estimator has access to \(X^n\), which plays the role of the training data, while \(X_{n+1}\) is the unavailable test point. When the context is clear, we will drop $f$ from our notation $\estimand_f$, and write just $\estimand$ instead.

Our goal is to estimate \(\estimand\) using only the observed sequence \(X^n\). Specifically, we seek an estimator \(\estimator : \Xcal^n \to \mathbb{R}\), and we assess its performance through the mean squared error
\begin{align}
\MSE \big(\estimator,\estimand\big)
:=
\EE \left[\big(\estimand - \estimator\big)^2\right].
\end{align}

\subsection{Assumptions on the stochastic process}\label{sec:mixing_SP}
We present our modeling assumptions on the stochastic process $X^{n + 1}$ through two definitions: $\beta$-mixing and Martonizability.

The $\beta$-mixing coefficient of $X^{n+1}$ at lag $\tau$ is defined as follows:
\begin{definition}[$\beta$-mixing coefficient]\label{defn:beta_mixing}
Let $\widetilde X^{n+1} = (\widetilde X_1,\ldots,\widetilde X_{n+1}) \in \Xcal^{n+1}$
denote an i.i.d.\ copy of $X^{n+1}$. For a given time lag $\tau \in [n]$, the
$\beta$-mixing coefficient of the sequence $X^{n+1}$ is defined as
\citep{doukhan1995mixing}
\begin{align*}
    \beta(\tau) \defn \max_{i \in [n+1-\tau]} \TV\big( (X_1,\ldots,X_{i},X_{i+\tau},\ldots,X_{n+1}), (X_1,\ldots,X_{i},\widetilde X_{i+\tau},\ldots,\widetilde X_{n+1}) \big).
\end{align*}
\end{definition}
In words, if $\beta(\tau)$ is small then the segments $(X_1,\ldots,X_{i})$ and $(X_{i+\tau},\ldots,X_{n+1})$ are nearly independent. For an i.i.d.\ process, $\beta(\tau) = 0$ for all $\tau \geq 1$.

Our second assumption on the stochastic process relies on the Marton-coupling framework
of \citet{marton2003measure,paulin2015concentration}. A Marton coupling for a dependent sequence $X^n$ defines how a change at a particular time index propagates to future time indices \citep[Definition~2.1]{paulin2015concentration}. Suppose we perturb $X_i$ and we wish to quantify the effect of this change at a future time index $j>i $. Marton coupling allows us to do so by defining an upper triangular matrix $\Gamma$ which we refer to as the mixing matrix. $\Gamma_{ij}$ records the worst-case probability that the original sequence and perturbed trajectory disagree at time $j$ after a perturbation at $i$. The diagonal entries of $\Gamma$ are set as $1$. Thus, smaller off-diagonal entries of
$\Gamma$ correspond to weaker dependence; for an i.i.d.\ process $\Gamma$ reduces
to the identity matrix.

\begin{definition}[Martonizable process]\label{defn:martonizable}
Let $X^n \defn (X_1,\ldots,X_n)$ be a vector of random variables taking values in
$\Xcal^n$, and let $\gamma \geq 1$ and $s \in \mathbb{N}$. We say that $X^n$ is
\emph{$(\gamma,s)$-Martonizable} if the \emph{uniform block partition}
\[
    \widehat{X}^{m}
    \defn
    \Big(
    \underbrace{(X_1,\ldots,X_{s})}_{\widehat X_1},\;
    \underbrace{(X_{s+1},\ldots,X_{2s})}_{\widehat X_2},\;
    \ldots,\;
    \underbrace{(X_{(m-1)s+1},\ldots,X_n)}_{\widehat X_m}
    \Big),
    \qquad m = \lceil n/s\rceil,
\]
into $m$ contiguous blocks of equal length $s$ (with a possibly shorter final
block) admits a Marton coupling in the sense of
\citet[Definition~2.1]{paulin2015concentration} whose mixing matrix
$\Gamma \in \mathbb{R}^{m\times m}$ satisfies $\opnorm{\Gamma} \leq \gamma$.
\end{definition}

\begin{remark}
Definition~\ref{defn:martonizable} is a special case of the partition-based
Marton-coupling framework of \citet[Definition~2.3]{paulin2015concentration},
which permits an arbitrary partition of $\{X_1,\ldots,X_n\}$ into disjoint,
non-empty subsets. We restrict attention to \emph{contiguous blocks of a common
size $s$}, so that partition size (in the notation of~\citet{paulin2015concentration}) is exactly $s$ and the number of blocks is exactly $m = \lceil n/s\rceil$. 
\end{remark}

Several natural stochastic processes are both $\beta$-mixing and Martonizable. We give two examples below.

\paragraph{Ergodic Markov chains:}\label{sec:ergodic_MC}
Suppose $X^{n+1}$ is a time-homogeneous Markov chain
on $(\Xcal,\mathcal{F})$ with transition kernel $P$ and stationary distribution
$\pi$. For $t\in\mathbb N$, let $P^t(x,\cdot)$ denote the $t$-step transition law
started from $x$. For $\epsilon\in(0,1/2]$, the mixing time of the chain is given by
\begin{align}\label{defn:mixing_time}
\tmix(\epsilon)
\defn
\min\left\{
t\in\mathbb N:
\sup_{x\in\Xcal} \TV \bigl(P^t(x,\cdot),\pi\bigr)\le \epsilon
\right\},
\end{align}
with the convenient shorthand $\Tmix \defn \tmix(1/4)$.
If $X^{n+1}$ is additionally stationary, it is known~\citep[e.g.][]{wolfer2024optimistic} that we have
\begin{align}
    \label{eq:mixing_time_relation}
    \beta(\tau) \leq \epsilon, \text{ for all } \tau \geq \tmix(\epsilon).
\end{align}
In addition, ergodic Markov chains are also known to be Martonizable. Partition the trajectory into
consecutive blocks of length $\Tmix$, so that we have
$s =  \Tmix$. By
\citet[Proposition~2.4]{paulin2015concentration} this block process admits a
Marton coupling, and by \citet[Remark~2.5]{paulin2015concentration} its mixing
matrix satisfies
\(
\opnorm{\Gamma} \leq \frac{2-1/4}{1-1/4} < 3
\). Hence $X^n$ is
$(3,\Tmix)$-Martonizable with $m \asymp n/\Tmix$.

\paragraph{\texorpdfstring{$\ell$}{l}-dependent processes:}\label{sec:memory_chains}

We say that $X^{n+1}$ is $\ell$-dependent if for each $1 \leq i \leq n+1-\ell$ the
blocks $(X_1,\ldots,X_i)$ and $(X_{i+\ell},\ldots,X_{n+1})$ are independent. By definition, this implies that $\beta(\tau) = 0$ for all $\tau \geq \ell$.
To verify Martonizability, partition the trajectory into
consecutive blocks of length $\ell$, so that $s = \ell$ and the
number of blocks is $m = \lceil n/\ell\rceil$. By
\citet[Example~2.15]{paulin2015concentration} this block process admits a Marton
coupling, and its mixing matrix satisfies $\opnorm{\Gamma} \leq 2$. Hence $X^n$ is
$(2,\ell)$-Martonizable with $m \asymp n/\ell$. We adopt the convention that for an $\ell$-dependent process, $\Tmix = \ell$.

\section{Methodology and general theory}\label{sec:estimator}

Having presented our assumptions on the stochastic process, we now turn
to the estimation problem set up above and
propose a consistent estimator for the next-token functional. Before presenting our estimator, let us revisit the leave-one-out approach~\eqref{eq:LOO} and inspect what goes wrong with it.
 
\subsection{Failure of leave-one-out estimation beyond i.i.d.\ setting}
As mentioned above (see the discussion surrounding Eq.~\eqref{eq:LOO}), the classical leave-one-out estimator with i.i.d.\ data treats each observed $X_i$ as a
stand-in for the test point $X_{n+1}$ and uses the empirical measure $\mathsf{dP}_{-i}$ as the corresponding stand-in for $\mathsf{dP}_n$.
Averaging over $i$, we then obtain expression~\eqref{eq:LOO}. The estimator is not generally unbiased (since $\mathsf{dP}_{-i}$ is not equal in distribution to $\mathsf{dP}_n$), but in many situations the bias decays with $n$.

\begin{figure}
		\centering
            \subfigure[Surprise probability]{\label{fig:loo_surprise}\includegraphics[width=7.85cm]{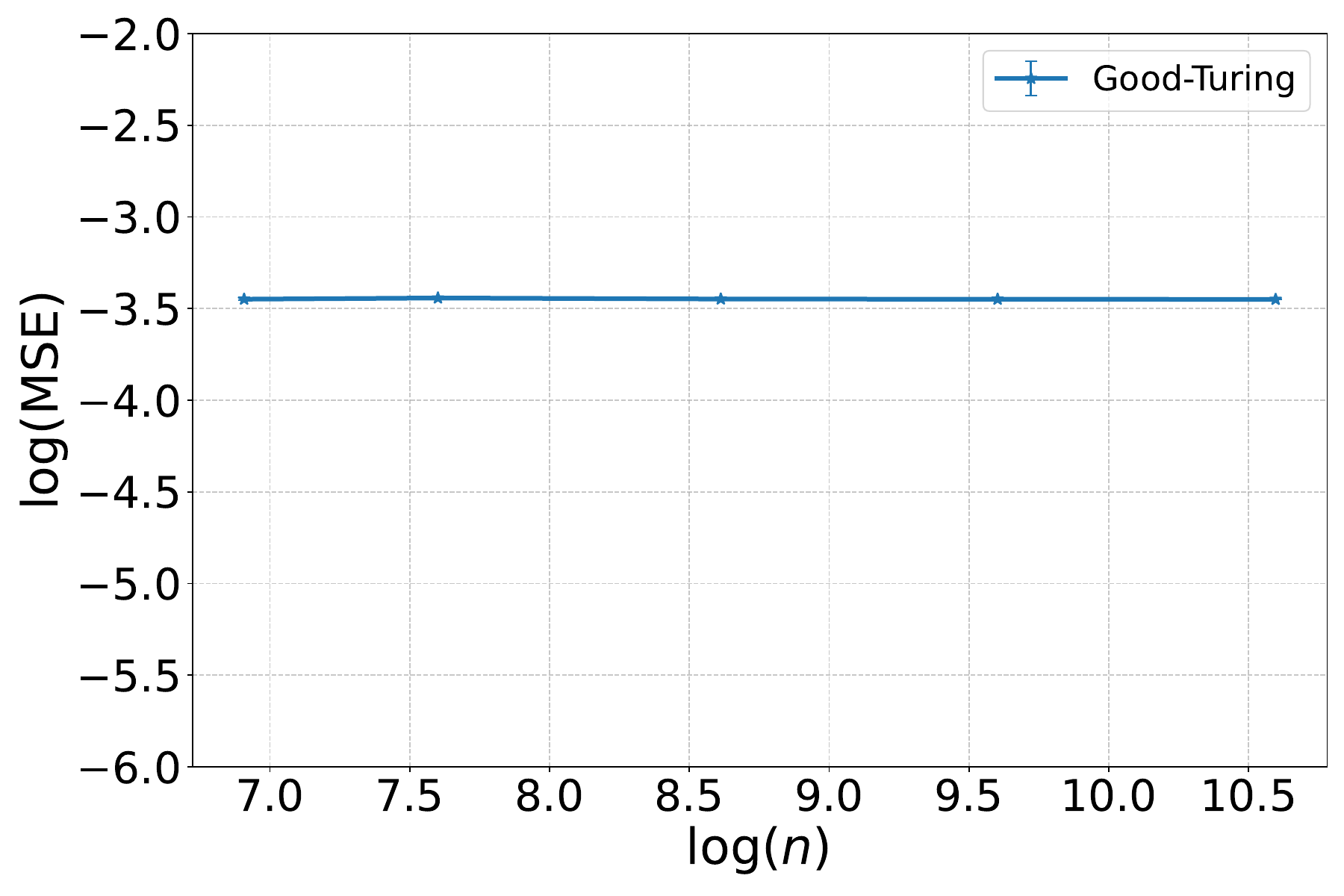}}
            \subfigure[Classification test error]{\label{fig:loo_cte}\includegraphics[width=7.85cm]{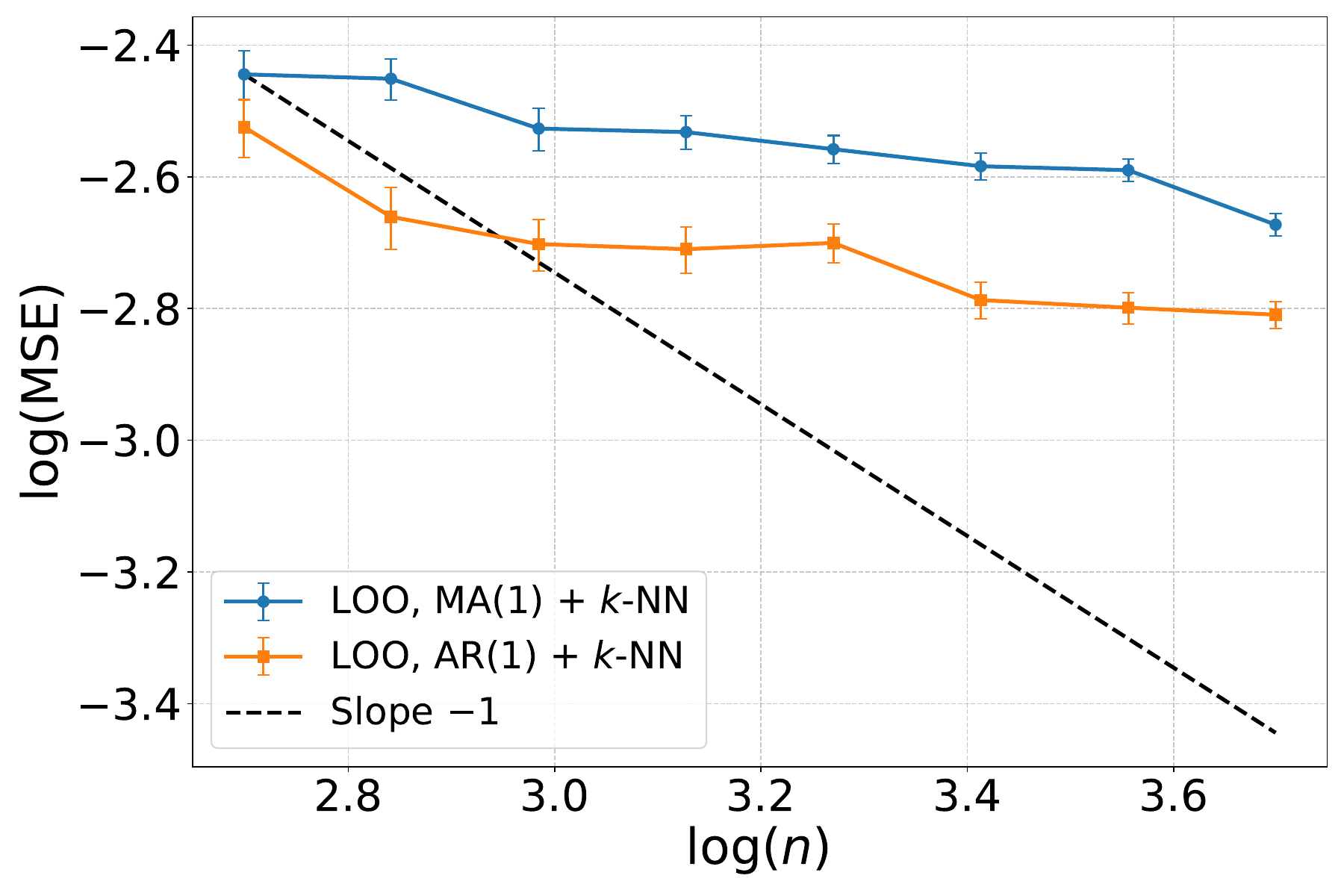}}
        \caption{Failure of leave-one-out estimation under temporal dependence.
        (a) The Good--Turing estimator, which is the leave-one-out estimator for surprise probability, used on a sticky Markov chain with $\Tmix=4$. The $\MSE$ remains constant as $n$ grows. (b) The leave-one-out test error
        estimator with the $k$-nearest neighbor rule for classification. The dependent covariates are generated using the moving average (MA($1$)) and autoregressive (AR($1$)) process. The labels are obtained using the sign of the first coordinate of the covariates. The $\MSE$ decreases at a very slow rate with $n$. The black dotted line plots the parametric $n^{-1}$ rate. See Section~\ref{sec:numerical_results} for more details.}
        \label{fig:LOO_failure}
	\end{figure}

Unfortunately, under temporal dependence, this substitution breaks down: $X_i$ is correlated with
its neighbors $X_{i+1},X_{i+2}, \dots$, which remain in the sample after
deleting a single point, so $X_i$ is far more
predictable from $\mathsf{dP}_{-i}$ than the test point $X_{n+1}$ is from $\mathsf{dP}_n$.
An observed point is therefore no longer a valid stand-in for the test point.
Appendix~\ref{example:1} makes this concrete, exhibiting a simple dependent process on which the leave-one-out estimator for surprise probability (i.e. Good--Turing) suffers a bias that does not vanish with $n$.
Figure~\ref{fig:LOO_failure} shows this failure empirically for two concrete functionals. For the surprise probability
(Fig.~\ref{fig:loo_surprise}), the error of the Good--Turing estimator is
essentially constant as $n$ increases. For the classification test error
(Fig.~\ref{fig:loo_cte}), the leave-one-out estimator does improve with $n$, but
at a slow rate. 

\subsection{The leave-a-window-out estimator}
To circumvent the issue of temporal dependence, we adopt a ``leave-a-window-out'' strategy. The core intuition is to continue to use $X_i$ as a proxy for the test point $X_{n+1}$, but to omit  $\tau$ points after index $i$ before computing a proxy for the empirical measure. 
Formally, for each index $i \in [n]$ in the sequence, we partition the data by defining the following index sets:
\begin{align} \label{eq:index-sets}
\Dset_i = \{k \in [i,(i+\tau-1) \wedge n]  \} \quad \text{ and } \quad \Iset_i = [n] \setminus \Dset_i.
\end{align}
Here, $\Dset_i$ captures the highly dependent local window starting at index $i$, and $\Iset_i$ contains those indices that are in the past as well as those that are far enough into the future from index $i$. For each subset \(\Iset \subseteq [n]\), let $\measure_{\Iset} := \frac{1}{|\Iset|}\sum_{j\in \Iset}\delta_{X_j}$ denote the empirical measure on the points corresponding to index set $\mathcal{I}$. With this notation, the $i$-th proxy is then given by
\begin{align*}
\singleestimator(X^n,\tau):= f(X_i;\measure_{\Iset_i}).
\end{align*} 
The final estimator for the next-token functional is then obtained by taking a simple average of the above estimates across the trajectory, i.e.,
\begin{align}\label{eq:estimator}
\estimator(X^n,\tau) = \frac{1}{n} \sum_{i=1}^{n} \singleestimator(X^n,\tau).
\end{align}
The proposed estimator is parameterized by the observed trajectory $X^n$ and the window length $\tau$. For ease of notation, we drop the dependence on $X^n$ when it is clear from context and write $\estimator(\tau)$.
Note that if $\tau = 1$, then we recover the leave-one-out estimator designed for i.i.d. data. 

Our estimator is
inspired by the recent Windowed Good--Turing (WingIt) estimator
\citep{pananjady2024just}, originally proposed for estimating stationary missing mass with
Markovian data. However, there are two noteworthy differences. First, our window is chosen only into the future (and not the past) of each proxy test point. Second, the analysis techniques that we use to develop our general theory are distinct from the specific analysis used in prior work. 

Having presented our estimator~\eqref{eq:estimator}, we now turn to stating our theoretical guarantees on its performance. Before doing so, however, we present assumptions under which these guarantees will hold.

\subsection{Assumptions on the functional}

While our assumptions can be stated in several ways, it is convenient to impose regularity conditions on the functional \(f\) and on the estimator \(\estimator(X^n,\tau)\). 
Recall our notation
$\measure_{\Iset}$,
\(\Iset_i\),
and $\Dset_i$
from Eq.~\eqref{eq:index-sets}.

\begin{assumption}[Out-of-sample deletion stability]
\label{assmp:stability}
There exists a deterministic function
\(
r_f : \mathbb{N}\times\mathbb{N} \to [0,\infty)
\)
such that for every \(n\), and every window length \(\tau \in \{1,\ldots,n-1\}\),
\[
\frac{1}{n}\sum_{i=1}^{n}
\left|
\EE \left[f(X_{n+1};\measure_n)\right]
-
\EE \left[f(X_{n+1};\measure_{\Iset_i})\right]
\right|
\le r_f(\tau,n).
\]
\end{assumption}

\noindent
In words, deleting a block from the original sequence does not substantially change the population functional, on average over the possible deletion locations. We use this assumption to control the bias of our estimator.

Our next assumption on $f$ enters through the estimator $\estimator(X^n,\tau)$. For the sequence \(X^n=(X_1,\ldots,X_n)\in\Xcal^n\), \(i\in[n]\), and \(a\in\Xcal\), we define the replacement vector
\[
X^{(i)}(a) := (X_1,\ldots,X_{i-1},a,X_{i+1},\ldots,X_n).
\]

\begin{assumption}[Bounded differences on the functional estimator]
\label{assmp:bd}
There exists a numerical constant \(B_f>0\) such that for every \(n\), every \(X^n\in\Xcal^n\), and every window size $\tau \in \{1,\ldots,n-1\}$,
\[
\sup_{a,b \in \Xcal}\sup_{i \in [n]}\left|
\estimator \big(X^{(i)}(a),\tau\big)
-
\estimator \big(X^{(i)}(b),\tau\big)
\right|
\le \frac{B_f}{n}.
\]
\end{assumption}

Recall that $\estimator \big(X^{(i)}(a),\tau\big)$ is the estimator from Eq.~\eqref{eq:estimator} calculated on the replacement vector $X^{(i)}(a)$. This is a replace-one stability condition: altering a single observation should change the estimator by at most \(B_f/n\). This assumption enables a McDiarmid-type concentration argument under dependence \citep{paulin2015concentration} to bound the variance of our estimator.

\subsection{The general upper bound}

Having stated our assumptions, we now provide an upper bound on the risk of our estimator $\estimator(\tau)$ defined in Eq.~\eqref{eq:estimator}.

\begin{theorem}
    \label{theorem:MSE_theorem}
    Let $X^{n+1}$ denote a sequence of random variables from a
    stationary stochastic process with $\beta$-mixing coefficient $\beta(\tau)$. Assume that $X^{n}$ is
    $(\gamma,s)$-Martonizable. If the
    functional $f$ and estimator $\estimator(\tau)$ satisfy
    Assumptions~\ref{assmp:stability}--\ref{assmp:bd}, then
    \begin{align*}
        \MSE\big(\estimator(\tau),\estimand \big)
        \leq
        2 \cdot \Bigl(4\|f\|_{\infty}\,\Bigl( \beta(\tau)+ \frac{\tau}{n}\Bigr)+r_f(\tau,n)\Bigr)^2
        +
         \frac{2 \cdot \gamma^2\cdot B_f^2 \cdot s  }{n}.
    \end{align*}
\end{theorem}
Theorem~\ref{theorem:MSE_theorem} is proved in Section~\ref{sec:proof_MSE_theorem}.

A few remarks are in order. Theorem~\ref{theorem:MSE_theorem} shows an upper bound consisting of two terms. The first corresponds to the squared \emph{bias} and is captured by the term
$\left(\|f\|_{\infty}\,\Bigl( \beta(\tau)+ \frac{\tau}{n}\Bigr) + r_f(\tau,n)\right)^2$. This term depends only on the stability of the functional and the $\beta$-mixing coefficients of the process. The second is a \emph{variance} term that is governed by the Martonizability of the process, and is proportional to $\tfrac{\gamma^2 B_f^2 s}{n}$. 

Note that an i.i.d.\ process is $(1,1)$-Martonizable \citep{paulin2015concentration}, so that $\gamma=1$ and $s=1$. Recall that for such a process $\beta(\tau) = 0$ for all $\tau \ge 1$. Provided $r_f(\tau,n)$ is a lower-order term, taking the window size $\tau =1$, yields $\MSE = \mathcal{O}(1/n)$, recovering the typical i.i.d.\ guarantee.
It is also useful to specialize the bound for the two families of stochastic processes discussed in Section~\ref{sec:memory_chains}.
For an ergodic Markov chain initialized at stationarity, the chain is
$(3,\Tmix)$-Martonizable, so that $\gamma = 3$ and
$s = \Tmix$. Choosing
$\tau \geq \tmix(\epsilon)$ for any $\epsilon \in (0,1/2]$ gives
$\beta(\tau) \leq \epsilon$, and thus we obtain
\begin{align}
\label{eq:mse_ergodic}
     \MSE\big(\estimator(\tau),\estimand\big)
     \leq
     2\Bigl(4\|f\|_{\infty}\,\Bigl( \epsilon+ \frac{\tau}{n}\Bigr) + r_f(\tau,n)\Bigr)^2
     + \frac{18\cdot B_f^2 \cdot \Tmix}{n}.
\end{align}
The variance term is thus $\mathcal{O}(\Tmix/n)$ for \emph{every} choice
of $\tau$. To make the bias term of the same order, take
$\epsilon = \sqrt{\Tmix/n}$, for which $\tau = \tmix(\epsilon) \lesssim \Tmix \log(n/\Tmix)$
suffices. Provided $r_f(\tau,n)$ is a lower-order term, Ineq.~\eqref{eq:mse_ergodic} then yields
\(
\MSE = \mathcal{O}(\Tmix/n).
\)

For an $\ell$-dependent process initialized at
stationarity, the process is $(2,\ell)$-Martonizable, so that $\gamma = 2$ and
$s = \ell$. Choosing $\tau \geq \ell$ gives $\beta(\tau) = 0$ exactly, and thus we obtain
\begin{align}
    \MSE\big(\estimator(\tau),\estimand\big)
    \lesssim
    \left( \frac{\tau \|f\|_{\infty}}{n} + r_f(\tau,n) \right)^2 + \frac{B_f^2 \cdot \ell}{n}.
\end{align}
Once again, the RHS is $\mathcal{O} \big(\frac{B_f^2\,\ell}{n}\big)$ provided $r_f$ is lower-order.

To summarize, in both cases above, the $\MSE$ is of order
$\mathcal{O} \big(\tfrac{\Tmix}{n}\big)$,
 and so we obtain the usual parametric rate but with a reduction in effective sample size by a multiplicative factor $\Tmix$.

\section{Consequences for specific functionals} \label{sec:f_example}

In this section we derive consequences of Theorem~\ref{theorem:MSE_theorem} for some specific examples of next-token functional estimation.

\subsection{Surprise probability}\label{sec:surprise}

The surprise probability of the sequence $X^{n+1}$ is given by
\begin{align}\label{eq:surprise_probability}
\surprise := \Prob( X_{n + 1} \notin \bX_{[n]} ).
\end{align}
Recalling our definition for the support of a measure, the surprise probability can be written as a next-token functional with
\begin{align*}
    f(X_{n+1};\measure_n) \defn \ind{X_{n+1} \notin \text{supp}(\measure_n)}.
\end{align*}
Existing work provides theoretical control of surprise probabilities in natural classes of stochastic processes~\citep{norris2014surprise}, but to the best of our knowledge, does not study explicit estimators for $\surprise$. Estimating surprise probabilities is motivated by applications such as anomaly detection—where one aims to flag rare or unexpected temporal events—and sample coverage assessment, where one wishes to quantify how much of the state space has been seen. Recent work also relates the phenomenon of hallucinations in large language models to surprise \citep{kalai2024calibrated}. 

For estimating the surprise probability, we again use the index sets defined in Section~\ref{sec:estimator}. For each index \(i \in [n]\), we say that \(X_i\) is a \emph{local surprise} with respect to the decorrelated subsequence \(X_{\Iset_i}\) if it does not appear in the retained portion of the trajectory. Accordingly, we define the estimator as
\begin{align*}
    \Shat_{(i)}(\tau) := \mathbb{I}\{X_i \notin \bX_{\Iset_i}\} \text{ and } \Shat(\tau) = \frac{1}{n}\sum_{i=1}^n \Shat_{(i)}(\tau).
\end{align*}

Note that when \(\tau = 1\), this estimator reduces to the usual leave-one-out estimator, which coincides with the Good--Turing estimator. A linear-time implementation of the above estimator (for general $\tau$) follows from the implementations provided in \citet{pananjady2024just,pmlr-v291-nakul25a}.
We now specialize Theorem~\ref{theorem:MSE_theorem} for the surprise
probability functional. 
\begin{corollary}\label{corollary:surprise_probability}
    Let $X^{n+1}$ denote a sequence of random variables from a
    stationary stochastic process with $\beta$-mixing coefficient $\beta(\tau)$. Assume that $X^{n}$ is
    $(\gamma,s)$-Martonizable. Then
    our estimator achieves the bound
    \begin{align*}
        \MSE\big(\Shat(\tau),\surprise\big)
        \leq
        64 \cdot \Bigl( \beta(\tau)+ \frac{\tau}{n}\Bigr)^2
        + 4 \cdot \Big( \frac{\tau}{n} \Big)^2
        +  \frac{18 \cdot \gamma^2 \cdot s}{n}.
    \end{align*}
\end{corollary}
Corollary~\ref{corollary:surprise_probability} is proved in
Appendix~\ref{sec:proof_corr_surprise_probability}. Recall that an ergodic Markov chain is
$(3,\Tmix)$-Martonizable and an $\ell$-dependent process is
$(2,\ell)$-Martonizable, so in both cases the last term is of order
$\gamma^2 s/n \asymp \Tmix/n$. Choosing $\tau \asymp \Tmix \log \left( n/\Tmix \right)$, we recover an $\MSE$ of
$\mathcal{O}(\Tmix/n)$. The above bound is information-theoretically optimal, as shown by the following proposition.

\begin{proposition}[Minimax lower bound for surprise probability]
\label{thm:minimax_risk}
Let $\Pcal_{\mathrm{MC}}(\Tmix; n+1)$ denote the class of all
stationary finite-state ergodic Markov chains of length $n+1$ with mixing time at most
$\Tmix$. For every integer $\Tmix\ge2$ and $n\ge2\Tmix$, we have
\[
    \inf_{\widetilde\surprise}
    \sup_{X^{n+1}\in\Pcal_{\mathrm{MC}}(\Tmix)}
    \EE \left[
        \big(
            \widetilde\surprise-\surprise
        \big)^2
    \right]
    \ge
    \frac{\Tmix}{1024n}.
\]
\end{proposition}
We prove Proposition~\ref{thm:minimax_risk} in Appendix~\ref{sec:proof_thm_minimax}.
The lower bound is established by constructing a structured hard subfamily of
stationary finite-state ergodic Markov chains. The construction is based on
independent latent labels, with the temporal dependence arranged so that a
length-$n$ trajectory reveals only order $n/\Tmix$ distinct labels. While we do not present one in this paper, we note that a similar lower bound can be constructed over the class of $\ell$-dependent processes.

\subsection{Count surprise probability}\label{sec:count_surprise}
Our second next-token functional is the probability that the unobserved test point
appears at most $\zeta \ge 0$ times in the observed sample
\begin{align}\label{eq:count_surprise_probability}
\Ncal := \Prob\big( N_{X_{n + 1}}(X^n) \leq \zeta \big), \;\text{for some integer } \zeta \ge 0,
\end{align}
where the probability is taken with respect to the distribution of $X^{n+1}$.
This can be framed as in Eq.~\eqref{eq:joint_functional} by defining $f$ as
\begin{align*}
    f(X_{n+1};\measure_n) \defn \ind{N_{X_{n + 1}}(X^n) \leq \zeta}.
\end{align*}
This functional is a natural extension of the surprise probability defined in
Section~\ref{sec:surprise} (since we recover the usual surprise probability when $\zeta = 0$). For each index $i \in [n]$, we say that $X_i$ is a
\emph{local surprise count} with respect to the decorrelated subsequence $X_{\Iset_i}$
if it appears at most $\zeta$ times in the retained portion of the trajectory.
Accordingly, we define the estimator as
\begin{align*}
    \Nhatlocal(\tau) := \mathbb{I}\{N_{X_i}(X_{\Iset_i}) \leq \zeta\}\text{ and } \Nhat(\tau) = \frac{1}{n}\sum_{i=1}^n \Nhatlocal(\tau).
\end{align*}
The following corollary specializes the bound in Theorem~\ref{theorem:MSE_theorem} for the count
surprise probability functional.
\begin{corollary}\label{corollary:count_surprise_probability}
    Let $X^{n+1}$ denote a sequence of random variables from a
    stationary stochastic process with $\beta$-mixing coefficient $\beta(\tau)$, and assume that $X^{n}$ is
    $(\gamma,s)$-Martonizable. Then our estimator achieves the bound
    \begin{align*}
        \MSE\big(\Nhat(\tau),\Ncal\big)
        \leq
        64 \cdot \Bigl( \beta(\tau)+ \frac{\tau}{n}\Bigr)^2
        + 4 \cdot \left( \frac{(\zeta+1)\tau}{n} \right)^2
        +  \frac{2 \cdot (2\zeta+5)^2 \cdot \gamma^2 \cdot s}{n}.
    \end{align*}
\end{corollary}
Corollary~\ref{corollary:count_surprise_probability} is proved in
Appendix~\ref{sec:proof_corr_count_surprise_probability}. Once again, one may specialize to ergodic Markov chains and $\ell$-dependent processes to obtain an
$\MSE$ of $\mathcal{O} \big( \frac{(2\zeta+5)^2\,\Tmix}{n} \big)$. Given Proposition~\ref{thm:minimax_risk} above, this bound
is optimal in its dependence on $n$ and $\Tmix$; obtaining a linear (rather than
quadratic) dependence on the count $\zeta$ remains open.

\subsection{Nearest-neighbor tail probability}\label{sec:nns}
Suppose $\Xcal \subset \mathbb{R}$, and suppose we are interested in the population-level probability that
the unobserved test point $X_{n+1}$ lies at distance greater than $\delta$ from
every point in the observed sample. This quantity is naturally interpreted as the
tail probability of the $1$-nearest-neighbor distance of $X_{n+1}$ to $X^n$, a
basic object in nearest-neighbor search, distance-based outlier detection, and
novelty detection \citep{clarkson2006nearest,chung2025conformalized}. We refer to
this quantity as the \emph{nearest-neighbor tail probability}, and define
\begin{align}
    \label{eq:defn_nns}
    \NS(\delta)
    \defn
    \Prob\left(\min_{i\in[n]} |X_{n+1}-X_i| > \delta \right),
    \qquad \text{ for some } \delta > 0.
\end{align}
Clearly, $\NS(\delta)$ can be
written as a special case of the general functional in
\eqref{eq:joint_functional} by taking
\begin{align*}
    f(X_{n+1};\measure_n)
    \defn
    \ind{\min_{i\in[n]} |X_{n+1}-X_i|>\delta}.
\end{align*}
Using the same leave-a-window-out construction, for each $i \in [n]$ we define the
estimator as
\begin{align*}
    \NShat_{(i)}(\tau)
    :=
    \mathbb{I}\left\{
        \min_{j \in \Iset_i} |X_i - X_j| > \delta
    \right\} \text{ and } \NShat(\tau)
    =
    \frac{1}{n}\sum_{i=1}^n \NShat_{(i)}(\tau).
\end{align*}
Thus $\NShat_{(i)}(\tau)$ records whether $X_i$ lies outside the
$\delta$-neighborhood of all points in the decorrelated subsequence $X_{\Iset_i}$,
after removing its locally-dependent window from the immediate future. 
We now apply Theorem~\ref{theorem:MSE_theorem} to this estimator.
\begin{corollary}
\label{corollary:nearest_surprise}
Let $\Xcal \subset \mathbb{R}$, and let $X^{n+1}$ denote a sequence of random variables from a
    stationary stochastic process with $\beta$-mixing coefficient $\beta(\tau)$. Assume that $X^{n}$ is
    $(\gamma,s)$-Martonizable. Then
    our estimator achieves the bound
\[
\MSE \big(\NShat(\tau), \NS(\delta)\big)
\le
64 \cdot \Bigl( \beta(\tau)+ \frac{\tau}{n}\Bigr)^2
+
4 \cdot \Big(\frac{\tau}{n}\Big)^2
+
  \frac{50 \cdot \gamma^2 \cdot s}{n}.
\]
\end{corollary}
Corollary~\ref{corollary:nearest_surprise} is proved in
Appendix~\ref{sec:proof_corr_nearest_surprise}. 

\subsection{Classification test error}\label{sec:cte}

Suppose we are given a trajectory
$Z^{n+1} := (Z_1,\ldots,Z_{n+1})$, where each observation takes the form
$Z_i = (X_i,Y_i)\in \Zcal$. Here $X_i$ denotes the covariates and $Y_i$ denotes
the class label. As in the general framework introduced earlier, we regard
$Z_{n+1}$ as an unobserved test point generated jointly with the observed
training trajectory $Z^n := (Z_1,\ldots,Z_n)$.
Working with the product space
$
\Zcal := \Xcal \times \{-1,1\},
$
empirical measures are formed on $\Zcal$, i.e.\ for $\Iset \subseteq [n]$, we write
$
\measure_n^Z \defn \frac{1}{n}\sum_{i=1}^n \delta_{Z_i}$ and $\measure_{\Iset}^Z \defn \frac{1}{|\Iset|}\sum_{i\in \Iset}\delta_{Z_i}.
$

Let $\widehat g_{Z^n} : \Xcal \to \mathbb{R}$ denote a classifier trained on $Z^n$. The predicted label at a point $x\in\Xcal$ is given by the sign of
$\widehat g_{Z^n}(x)$. To measure test performance, we consider a bounded loss
function
\[
\ell(\widehat g_{Z^n}, z)\in[0,1],
\qquad \text{ for } z=(x,y)\in\Zcal.
\]
Our target is the population test error
\begin{align}
\label{eq:test_error}
\testError
:=
\EE \left[\ell(\widehat g_{Z^n}, Z_{n+1})\right].
\end{align}
When the training algorithm is symmetric, $\widehat g_{Z^n}$ depends on the training sequence $Z^n$ only through its empirical measure, so $\testError$ can be
written as a special case of the general functional in
\eqref{eq:joint_functional}.

To address temporal dependence in the observed trajectory, we again adopt a
leave-a-window-out strategy, defining (in parallel to the above) the estimators
\begin{align}\label{eq:test-error-estimator}
\testErrorhat_{(i)}(\tau)
:=
\ell(\widehat g_{Z_{\Iset_i}}, Z_i) \text{ and } \testErrorhat(\tau)
=
\frac{1}{n}\sum_{i=1}^n \testErrorhat_{(i)}(\tau).
\end{align}

Note that when $\tau=1$, this estimator reduces to the usual leave-one-out
cross-validation estimator. Even for general $\tau$, the computational complexity of our estimator is typically comparable to that of the classical leave-one-out estimator.
To state our guarantee on this procedure, we assume that the loss function $\ell$ is uniformly
stable and the training algorithm is symmetric.

\begin{assumption} \label{assmp:uniform_stability}
   The training algorithm is symmetric, and the loss function $\ell$ is uniformly stable with respect to this
    algorithm, i.e., there is a numerical constant $C$, independent of the sample size,
    such that for every sample size $m \geq 2$, we have
    \[
    \sup_{z^m \in \Zcal^m}\sup_{z\in \Zcal}\sup_{i \in [m]}|\ell(\widehat g_{z^m},z)-\ell(\widehat g_{z^{[m] \setminus i}},z)| \leq \frac{C}{m}.
    \]
\end{assumption}
Uniform stability for the classifier ensures that our estimator satisfies the
bounded differences property (Assumption~\ref{assmp:bd}). For bounded loss
functions, various classification algorithms such as soft-margin SVMs and
regularized regression satisfy the uniform stability condition (see
\citet{bousquet2002stability} for further details).
Under uniform stability, the estimator $\testErrorhat(\tau)$ satisfies
Assumption~\ref{assmp:bd} and we obtain the following corollary. 

\begin{corollary}\label{corollary:test_error}
    Let $Z^{n+1} := (Z_1,\ldots,Z_{n+1})$ denote a sequence of random variables from a
    stationary stochastic process with $\beta$-mixing coefficient $\beta(\tau)$, where $Z_i = (X_i,Y_i) \in \Xcal \times \{-1,1\}$. Assume that $Z^n$ is
    $(\gamma,s)$-Martonizable. Then under
    Assumption~\ref{assmp:uniform_stability}, our estimator with  any $\tau \leq n/2$ achieves
    \begin{align*}
        \MSE\big(\testErrorhat(\tau),\testError\big)
        \leq
        64 \cdot \Bigl( \beta(\tau)+ \frac{\tau}{n}\Bigr)^2
        + 4 \cdot \left( \frac{C\tau}{n-\tau+1} \right)^2
        +   \frac{2 \cdot (4C+1)^2 \cdot \gamma^2 \cdot s}{n}.
    \end{align*}
\end{corollary}
Corollary~\ref{corollary:test_error} is proved in
Appendix~\ref{sec:proof_corr_test_error}. It is related to the stability-based
generalization literature. In the
i.i.d.\ setting, \citet{bousquet2002stability} show that the empirical training error of a uniformly stable algorithm estimates the population risk at the parametric rate. \citet{mohri2010stability} study stability-based generalization bounds for $\beta$-mixing sequences. They consider the empirical training error (without leaving out any points) and show that it consistently estimates the generalization error even in the presence of dependence. However, the rate is slow: converting the high-probability bound\footnote{The bound in \citet[Corollary 20]{mohri2010stability} is obtained using a blocking argument by dividing the entire trajectory into consecutive blocks of length $a$ and $b$. One way  to obtain the stated slow-rate bound is to  consider $\ell$ dependent processes, set $b=\Tmix = \ell$ and optimize the bound with respect to $a$.} from \citet[Corollary 20]{mohri2010stability}, yields an $\MSE$ guarantee of $\mathcal{O}(\Tmix/\sqrt{n})$, which is slower than the parametric rate $n^{-1}$ obtained above. 

One important point to note here is that the simple plug-in estimator is consistent due to the uniform stability assumption. Due to this stability, deleting the additional observations in a window only changes the learned predictor by a vanishing amount as $n$ grows. This behavior appears in the  regularized logistic-regression experiments of Appendix~\ref{sec:appendix_lr}. Here the predictor is uniformly stable and the behavior of our estimator and that of the leave-one-out estimator coincide. However, experiments with the $k$-nearest-neighbor algorithm in Section~\ref{sec:numerical_results} (which does not satisfy uniform stability) demonstrate that our approach can succeed in regimes where the leave-one-out estimator fails. Thus, it seems that Assumption~\ref{assmp:uniform_stability} here is sufficient rather than necessary. Studying our procedure under weaker stability assumptions (such as those satisfied by nearest-neighbors) is left for future work.

\section{Numerical results}\label{sec:numerical_results}

In this section, we provide a set of simulations on synthetically constructed dependent processes in order to demonstrate the effectiveness of the proposed estimator. 

\subsection{Surprise probability}
We evaluate surprise probability on two families of dependent processes:
\begin{enumerate}
    \item \textbf{Repeated-block chain}: Fix $\ell\ge3$ and
consider the repeated-block process,
where the i.i.d.\ base sequence $(\widetilde X_i)_{i\ge1}$ is drawn uniformly
from a finite state space $\Xcal=[k]$. For each of these labels we draw $L_i$ which denotes the number of times $\widetilde X_i$ is repeated until the trajectory reaches length $n$. $L_1$ is drawn from a law to ensure stationarity\footnote{In particular, this PMF is given by $\Prob(L_1 = r) = \frac{\Prob(\widetilde L \ge r)}{\EE[\widetilde L]} =\frac{2\Prob(\widetilde L \ge r)}{\ell+2}$, where $\widetilde L\ \sim\UNIF(\{2,\ldots,\ell\}) $.} and the remaining block lengths
$(L_i)_{i\ge2}$ are drawn independently as
\(
    L_i\sim\UNIF(\{2,\ldots,\ell\}).
\)
The block lengths are independent of the base sequence, and the process is
initialized at stationarity. The mixing time of this process is $\Tmix \leq \ell+1$.

    \item \textbf{Sticky Markov chain}: We consider a sticky Markov chain on $\Xcal = [k]$ with uniform stationary distribution and transition kernel
    \[
    P = p \mathbf{1}\pi^{\top} + (1-p)I.
    \]
    Its mixing time, $\Tmix$, is proportional to $1/p$ \citep{pananjady2024just,pmlr-v291-nakul25a}. 
\end{enumerate}

We compare the proposed estimator with the Good--Turing estimator and several add-constant estimators: Laplace \citep{laplace1814essai}, Krichevsky--Trofimov \citep{krichevsky1981performance}, and Braess--Sauer \citep{braess2004bernstein}.

\begin{figure}
		\centering
            \subfigure[Our estimator for different choices of $\tau$]{\label{fig:1}\includegraphics[width=7.85cm]{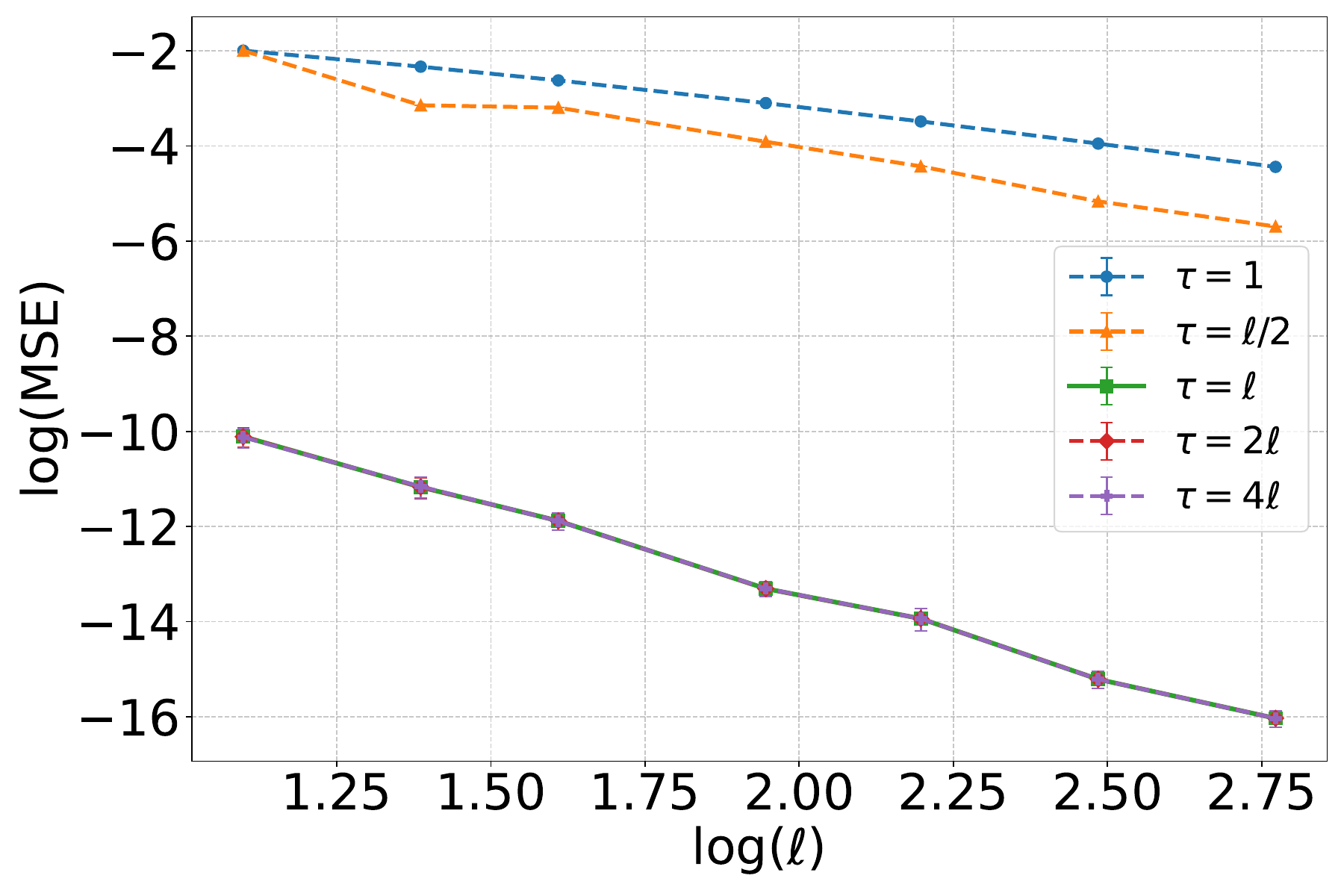}}
            \subfigure[Comparison with baselines]{\label{fig:2}\includegraphics[width=7.85cm]{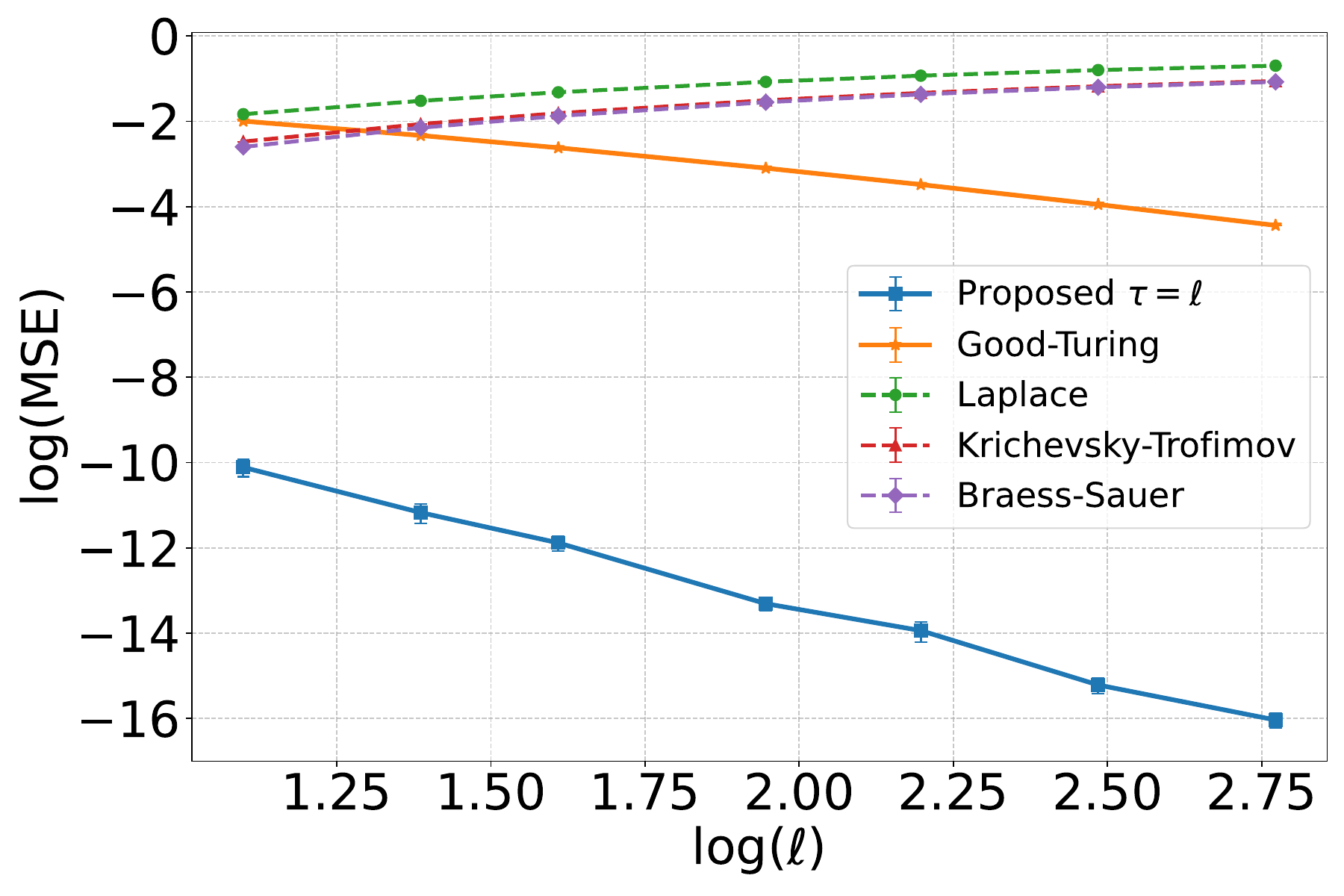}}
		\label{fig:estimation_plots1}
        \caption{Performance of the proposed estimator on the repeated-block chain, averaged over $64$ instances. In all cases, we set trajectory length $n = 64\ell^3$ and state space size $k = 5n$.}
        \label{fig:Repeat}
	\end{figure}

\begin{figure}
		\centering
            \subfigure[Our estimator for different choices of $\tau$]{\label{fig:3}\includegraphics[width=7.85cm]{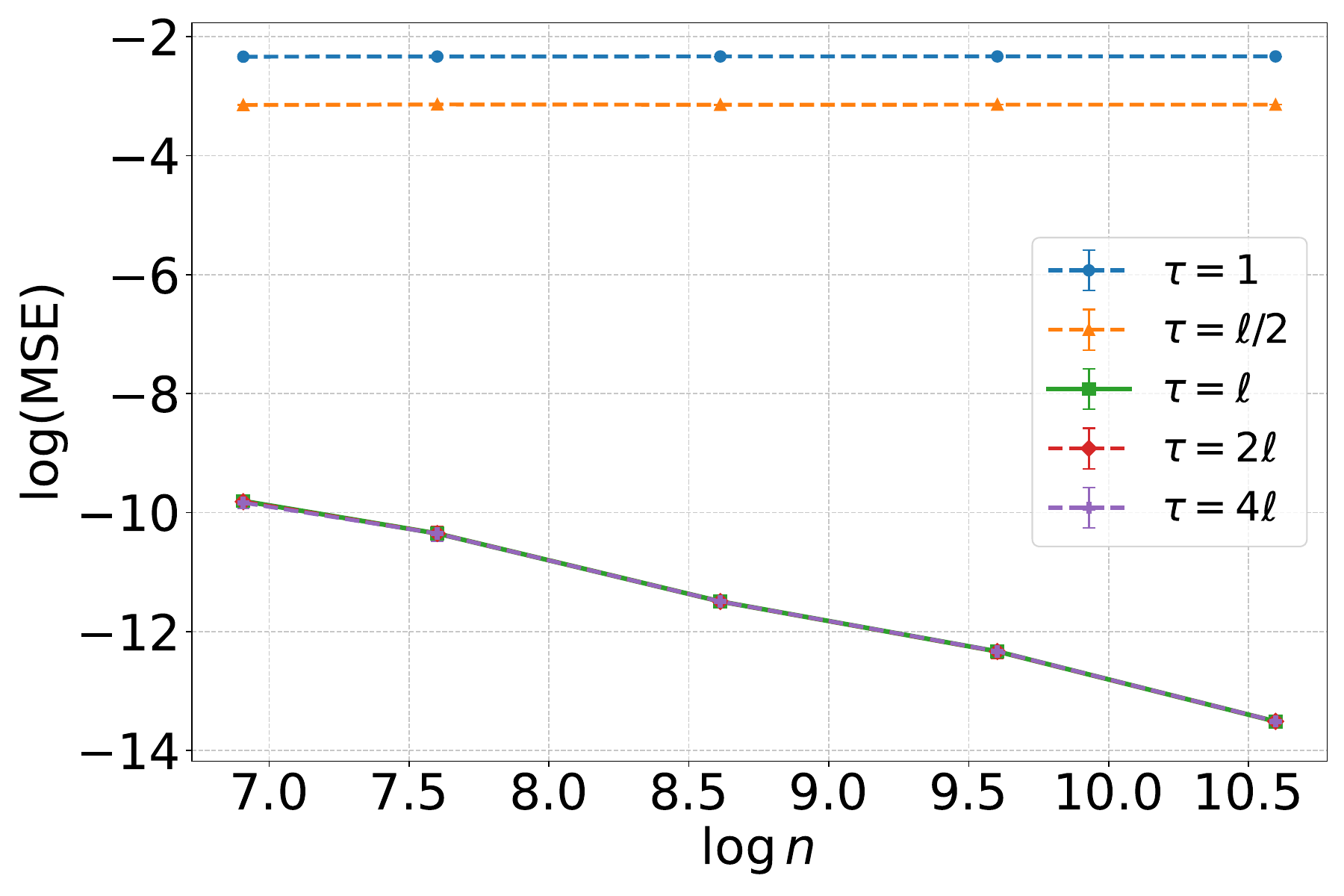}}
            \subfigure[Comparison with baselines]{\label{fig:4}\includegraphics[width=7.85cm]{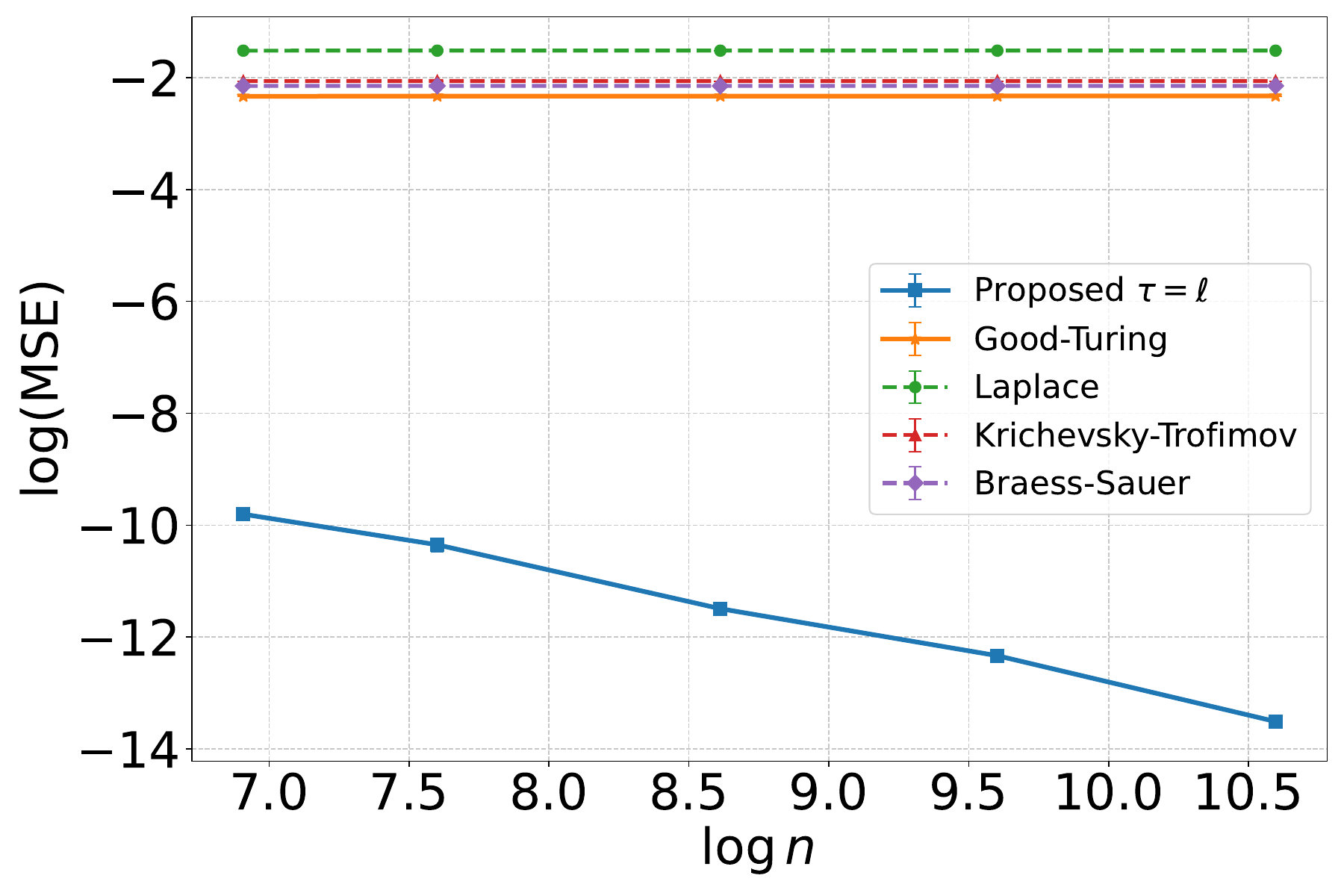}}
		\label{fig:estimation_plots2}
        \caption{Performance of the proposed estimator on the repeated-block chain, averaged over $200$ instances.  The block-length parameter is fixed as $\ell = 4$ and the state space size is $k = 5n$.}
        \label{fig:Repeat_vary_n}
	\end{figure}

For the repeated-block chain, Fig.~\ref{fig:1} shows the MSE of the proposed estimator, for different choices of the window size $\tau$, plotted against a range of block lengths $\ell\in \{3,4,5,7,9,12,16\}$. We set the trajectory length to $n = 64\ell^3$ and the state-space size to $k = 5n$. Once $\tau \ge \ell$, we observe that all the plots overlap. The plot also shows the robustness of the proposed estimator to the choice of $\tau$, as overestimating $\tau$ does not degrade performance. Fig.~\ref{fig:2} compares the proposed estimator with the baseline methods as we vary the dependence parameter $\ell$. The proposed estimator performs the best.
In Fig.~\ref{fig:3}, we plot the MSE of the proposed estimator for different choices of the window size $\tau$ against the trajectory length $n \in \{1000, 2000,$ $ 5500, 14800,40000\}$. We set $\ell=4$ and $k=5n$. Again, overestimating $\tau$ does not degrade performance. Fig.~\ref{fig:4} compares the proposed estimator with the baseline methods as we vary the trajectory length $n$. The proposed estimator performs the best while the other estimators fail to perform consistent estimation.

\begin{figure}
		\centering
            \subfigure[Our estimator for different choices of $\tau$ ]{\label{fig:5}\includegraphics[width=7.85cm]{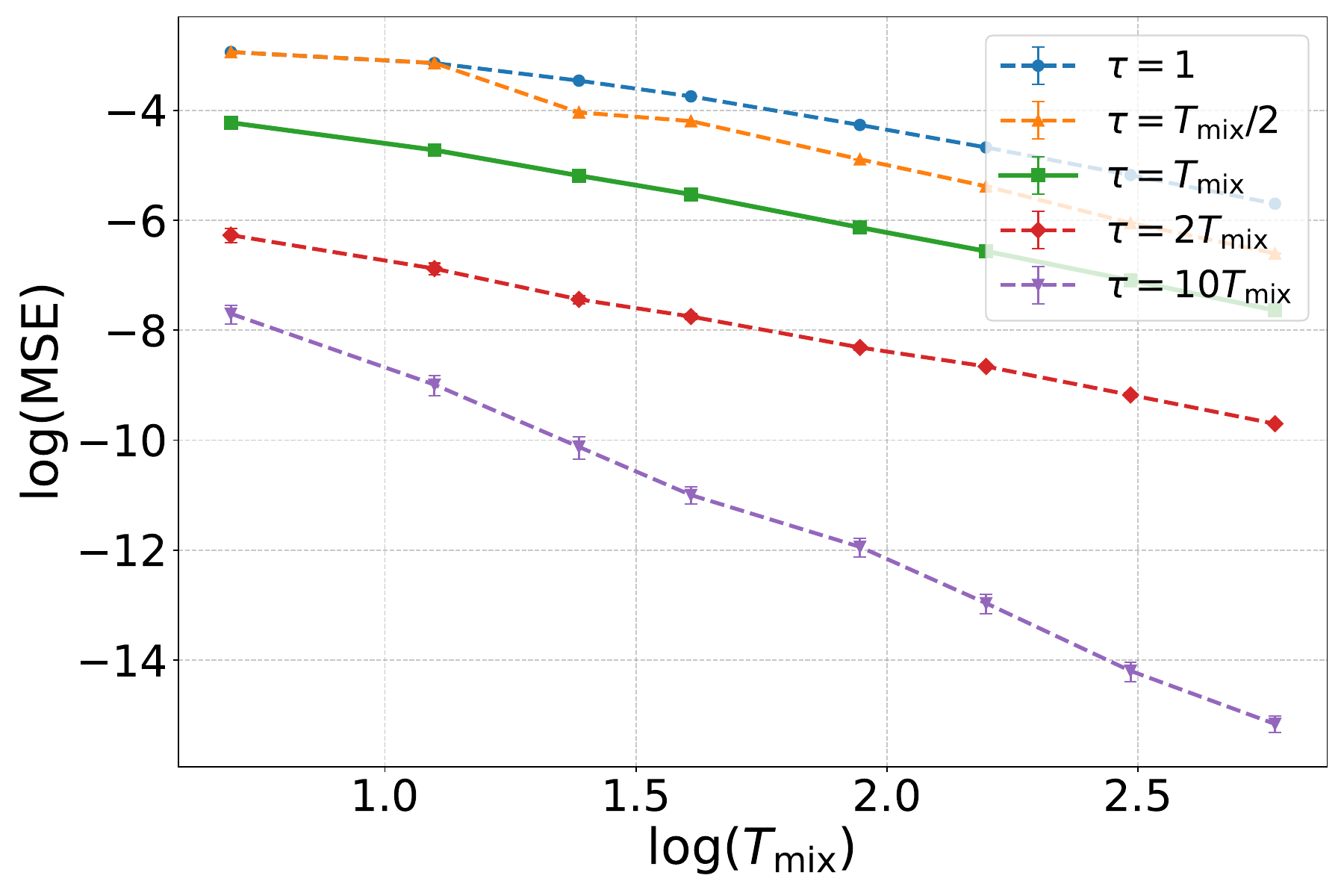}}
            \subfigure[Comparison with baselines]{\label{fig:6}\includegraphics[width=7.85cm]{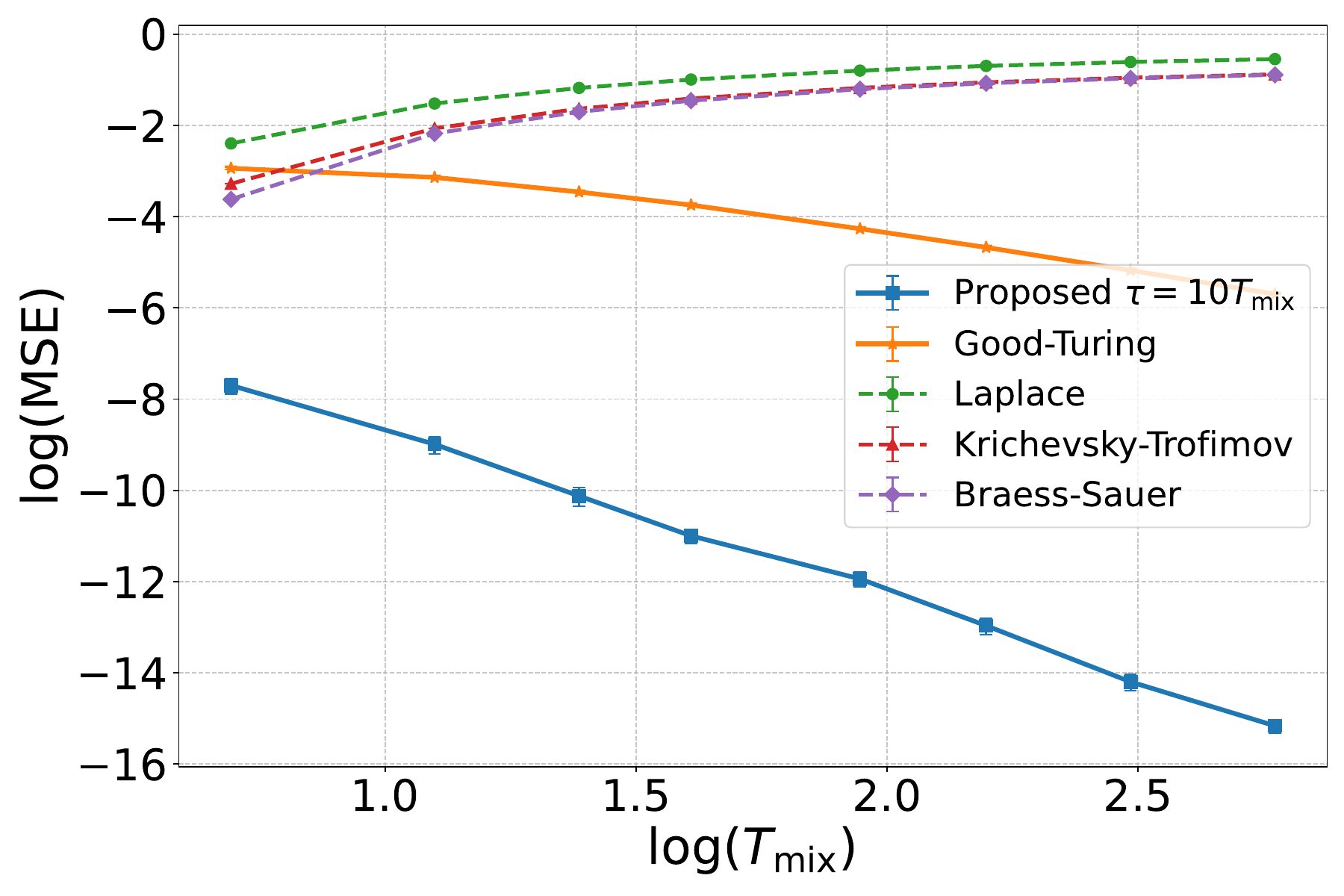}}
		\label{fig:estimation_plots3}
        \caption{Performance of the proposed estimator on the sticky Markov chain, averaged over $64$ instances.}
        \label{fig:Sticky1}
	\end{figure}

\begin{figure}
		\centering
            \subfigure[Our estimator for different choices of $\tau$]{\label{fig:7}\includegraphics[width=7.85cm]{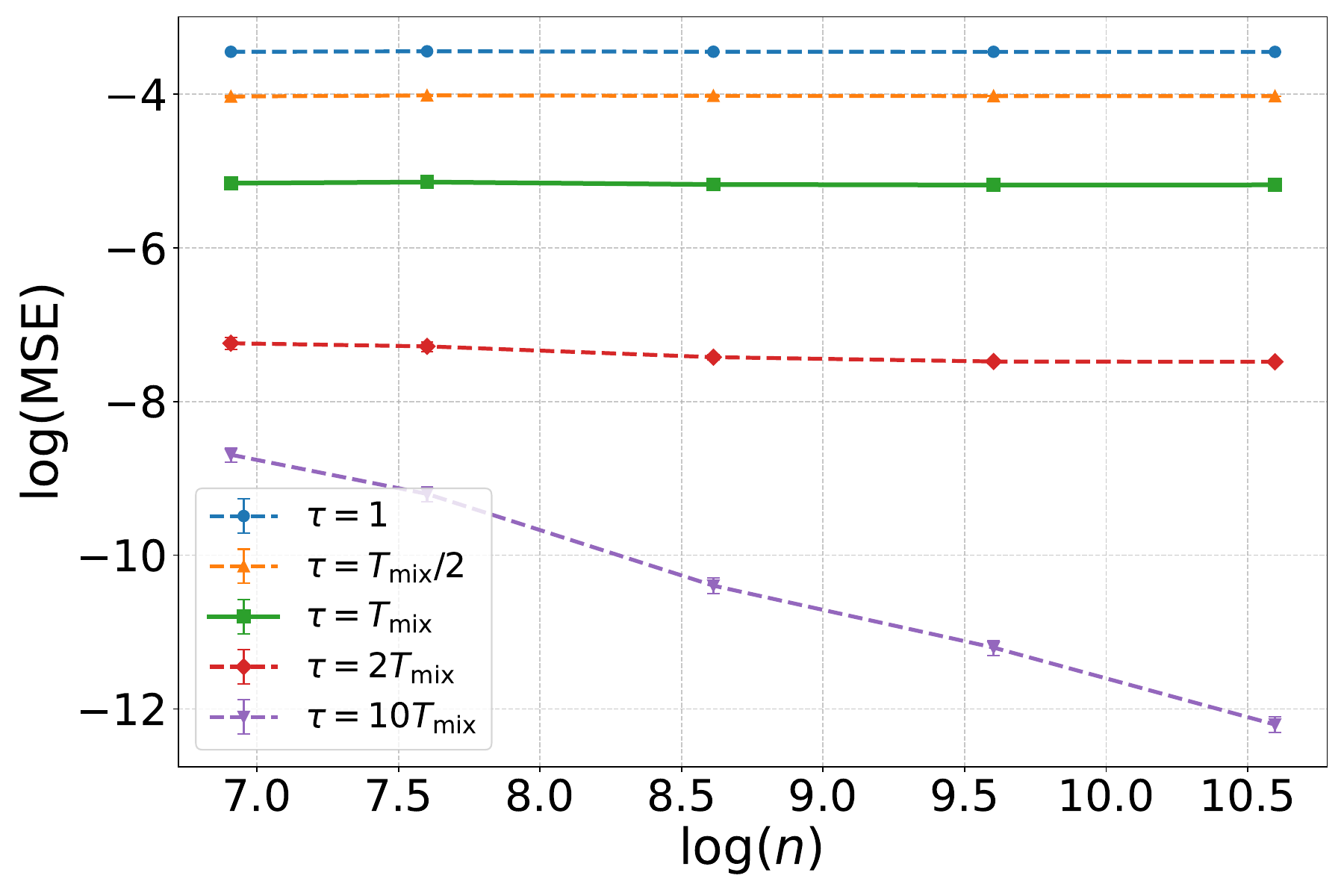}}
            \subfigure[Comparison with baselines]{\label{fig:8}\includegraphics[width=7.85cm]{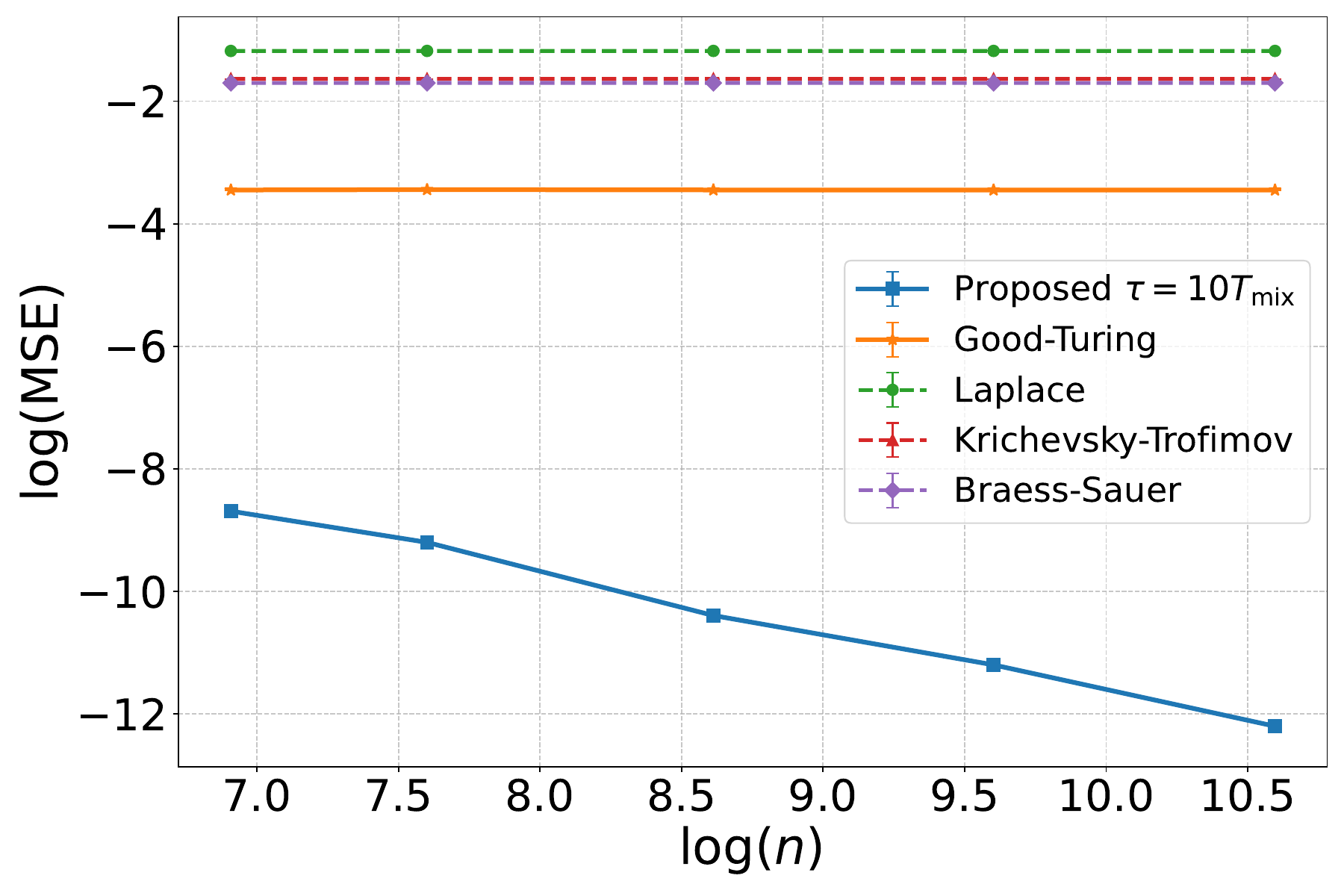}}
		\label{fig:estimation_plots4}
        \caption{Performance of the proposed estimator on the sticky Markov chain, averaged over $200$ instances. The mixing time is fixed as $\Tmix =4$.}
        \label{fig:Sticky2}
	\end{figure}

For the sticky Markov chain, Fig.~\ref{fig:5} shows the MSE of the proposed estimator, for different window sizes $\tau$, plotted against the mixing time $\Tmix \in \{2,3,4,5,7,9,12,16\}$. We set $n = 64\Tmix^3$ and $k=5n$. The performance improves as $\tau$ increases up to the dependence scale $\Tmix$. One important difference from the repeated-block chain example is that the performance of the proposed estimator improves as we increase the window size parameter beyond $\Tmix$. In Fig.~\ref{fig:6}, we compare the proposed estimator, with the best window size $\tau=10\Tmix$, with the baseline estimators as we vary mixing time $\Tmix$. The proposed estimator achieves the best performance with optimal window length. In fact, even with suboptimal window choices, the proposed estimator still performs better than the competing baselines. The add-constant estimators fail to perform consistent estimation while the Good--Turing estimator has a much slower rate than the proposed estimator.
In Fig.~\ref{fig:7}, we plot the MSE against the trajectory length $n \in \{1000, 2000,$ $ 5500, 14800,40000\}$ with $\Tmix = 4$ and $k=5n$. The $\MSE$ remains constant with $n$ until the window length parameter is set to be a multiple of $\Tmix = 4$. In Fig.~\ref{fig:8}, we compare the proposed estimator, using $\tau=10\Tmix$, with the baseline estimators, while varying the length of the trajectory $n$. All the baseline estimators fail to perform consistent estimation in this dependent regime.

\subsection{Classification test error}

We evaluate the classification  test error on two families of dependent processes:

\begin{enumerate}
    \item \textbf{Moving average (MA) process: }Fix an
order $q \ge 1$ and generate $\epsilon_1,\ldots,\epsilon_{n+q+1}
\iidsim \mathcal{N}(\mathbf{0},\mathbf{I}_d)$; the covariates are the sliding
window sums
\[
X_i \defn \sum_{j=0}^{q}\epsilon_{i+j},
\qquad i \in [n+1],
\]
and the label is the sign of the first coordinate,
$Y_i \defn \mathrm{sign}(\langle e_1, X_i\rangle)$. 
Because $X_i$ is a fixed function of the width-$(q{+}1)$ window
$(\epsilon_i,\ldots,\epsilon_{i+q})$ of an i.i.d.\ sequence, the process is
\emph{stationary} and  $\ell$-dependent with $\ell = q+1$. Thus, $\beta(\tau)=0$ for $\tau \geq \ell =q+1$.

\item \textbf{Autoregressive (AR) process: }Fix $\phi \in (0,1)$, generate
$X_0, \epsilon_1,\ldots,\epsilon_{n+1} \iidsim \mathcal{N}(\mathbf{0},\mathbf{I}_d)$,
and set
\[
X_i \defn \phi X_{i-1} + \sqrt{1-\phi^2}\,\epsilon_i,
\qquad i \in [n+1],
\]
with the label again $Y_i \defn \mathrm{sign}(\langle e_1, X_i\rangle)$. This makes the stationary distribution exactly
$\mathcal{N}(\mathbf{0},\mathbf{I}_d)$, so that the chain is started at stationarity. Unlike the MA process, an
AR chain is dependent at every lag: $\beta(\tau)$ decays
geometrically but never vanishes, so no finite window removes the dependence
exactly.
\end{enumerate}

The classification algorithm is the $k$-nearest neighbor rule with $k=3$. We compare the proposed leave-a-window-out estimator
$\testErrorhat(\tau)$ against the classical leave-one-out (LOO) estimator, which
is the special case for $\tau=1$. For both the MA and AR experiments, the target $\testError$ is computed by averaging
over $5\times 10^{4}$ trajectories, and each reported $\MSE$ is
averaged over $200$ trajectories. The MA experiment uses
$\ell \in \{2,3,4\}$ and the AR experiment uses
$\phi \in \{0.35,0.5,0.65\}$.

Before we present the results, we first give an intuitive explanation for working with the nearest-neighbor algorithm in high dimensions for an MA process. 
For the MA($1$) process, each $X_i$ is dependent on its neighbors $X_{i-1}$ and $X_{i+1}$ since it shares $\epsilon_i$ and $\epsilon_{i+1}$ with them respectively. $X_i$ is independent of all the other $X_j$'s for which $|i-j| \ge 2$. When the dimension $d$ is large, $\EE[\|X_i -X_j\|^2] \asymp 2d$ for $|i-j| =1$ and $\EE[\|X_i -X_j\|^2] \asymp 4d$ for $|i-j| \ge 2$. In high dimensions these two distances separate, thus making geometric distance equivalent to temporal dependence. Thus, the nearest-neighbor algorithm makes use of temporally correlated points for making predictions at any point $X_i$. Although the nearest-neighbor algorithm does not satisfy the uniform stability of Assumption~\ref{assmp:uniform_stability}, the equivalence between distance and temporal dependence makes it a relevant example to analyze.

\begin{figure}
    \centering
    \subfigure[Our estimator for different choices of $\tau$]{\label{fig:9}\includegraphics[width=8.0cm]{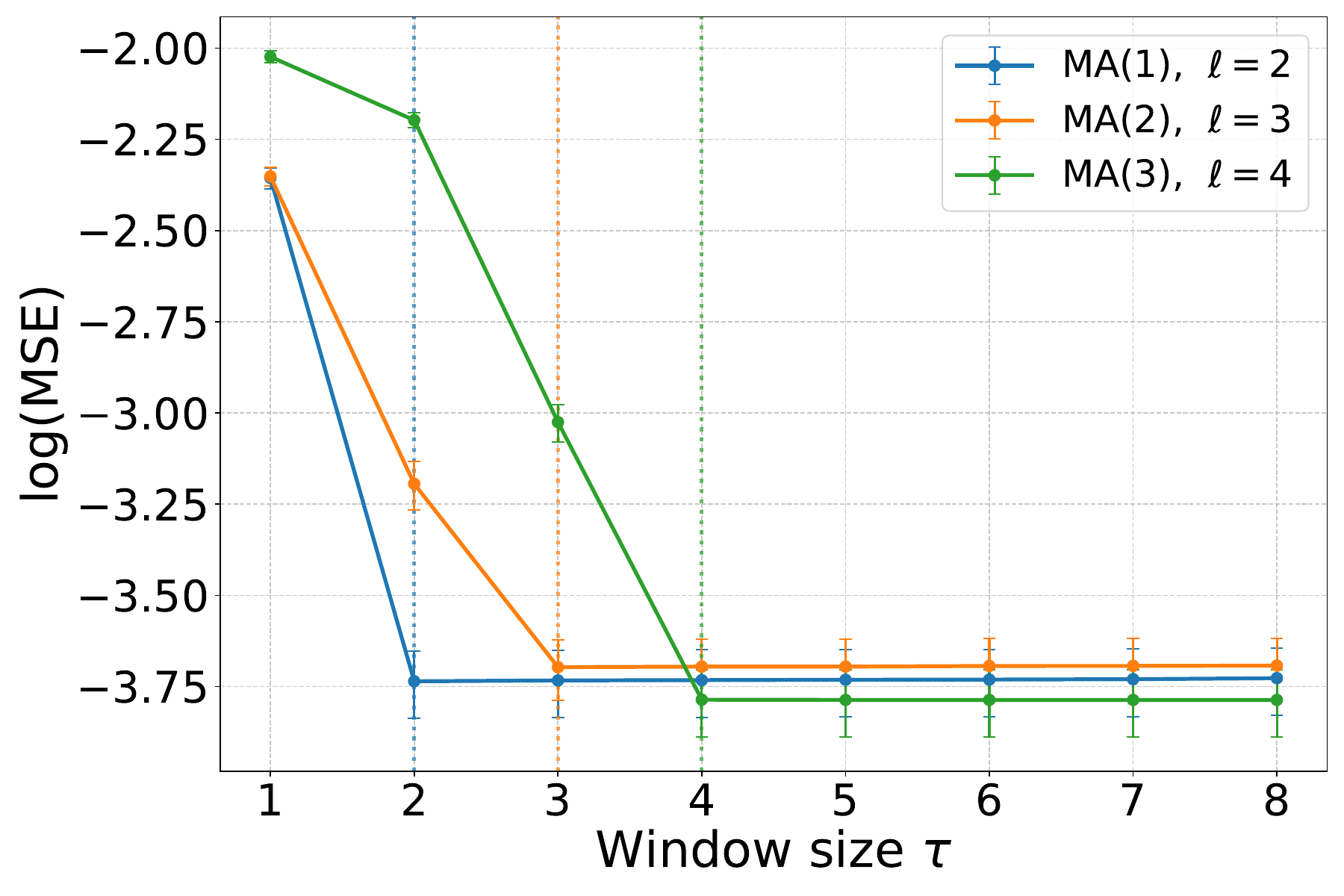}}
    \subfigure[Comparison with baseline]{\label{fig:10}\includegraphics[width=8.0cm]{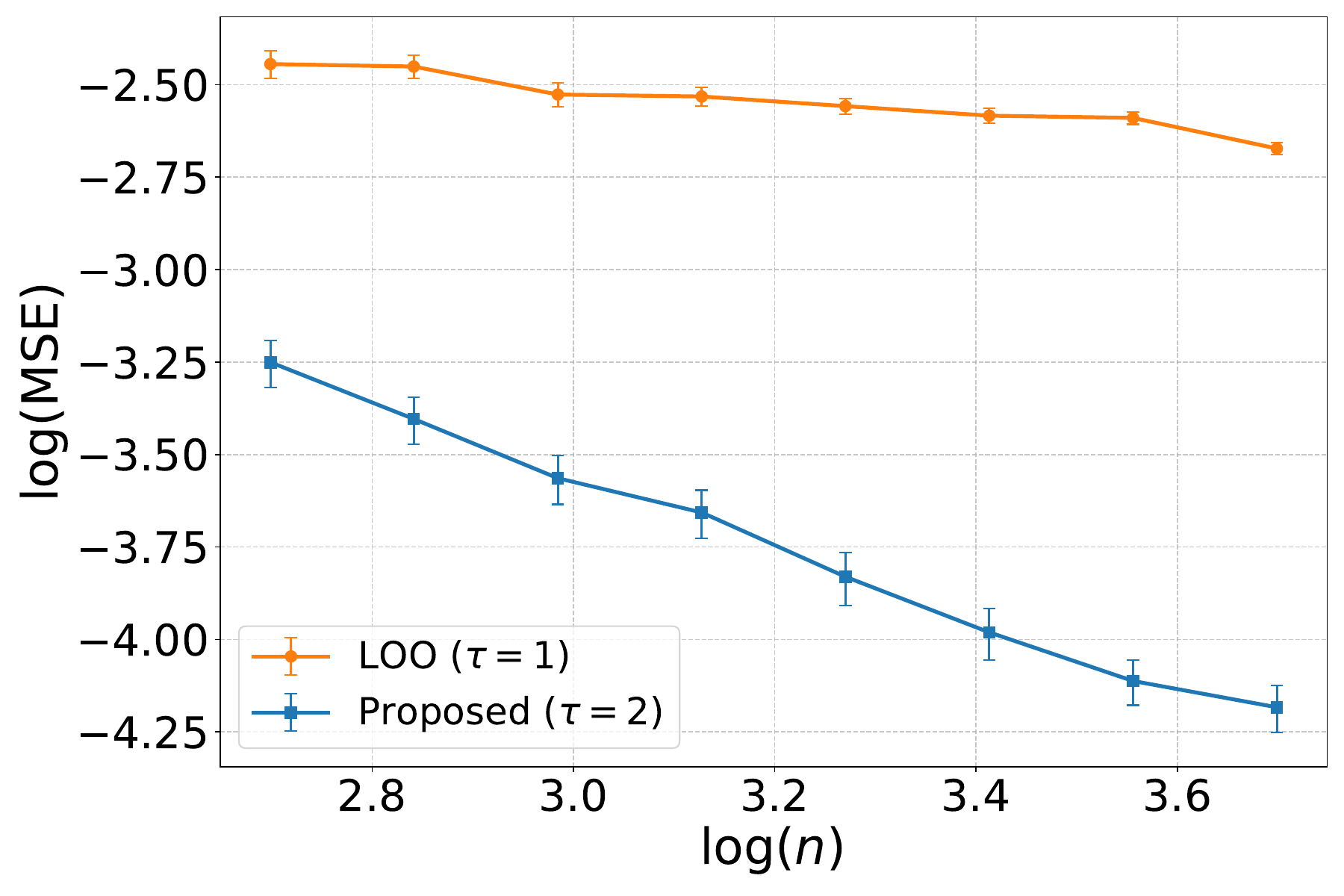}}
    \caption{Classification test error estimation on the MA process.
    (a) $\log(\MSE)$ against the window size $\tau$ at fixed $n=1500$ and $d=600$, for
    $\ell \in \{2,3,4\}$; dotted lines mark $\tau=\ell$.
    (b) $\log(\MSE)$ against $n$ for the MA($1$) process ($\ell=2$), comparing LOO
    ($\tau=1$) with the proposed estimator ($\tau=2$). \vspace{-4mm}}
    \label{fig:ClassificationTE}
\end{figure}

Let us now discuss our results.
Figure~\ref{fig:9} varies $\tau \in \{1,2,\ldots,8\}$ with $n=1500$, and $d=600$ for the MA($q$) process with $q\in \{1,2,3\}$. For each process the $\MSE$
decreases until $\tau$ reaches the dependence range and is flat
thereafter, and the elbow sits precisely at $\tau=\ell = q+1 \in\{2,3,4\}$. 
Figure~\ref{fig:10} varies $n \in \{499, 694, 965, 1341, 1863, 2589, 3598, 4999\}$ for the MA($1$) process with $\ell=2$ and $d=200$. We use the window length $\tau=2$. Recall that the classification algorithm is $3$-nearest neighbor. For each $X_i$, the LOO estimator uses the $3$ nearest neighbors in the training set which are $\{X_{i-1},X_{i+1},X_j\}$, where $X_j$ is independent of $X_i$. For the test point $X_{n+1}$ the closest points are $\{X_{n},X_j,X_k\}$, where $X_j$ and $X_k$ are independent of $X_{n+1}$. This creates a mismatch between the training points and the test point and results in a bias. Our estimator deletes the window $\Dset_i=\{i,i+1\}$ removing
the mismatch, and thus attaining consistent estimation.

\begin{figure}
    \centering
    \subfigure[Our estimator for different choices of $\tau$]{\label{fig:11}\includegraphics[width=7.5cm]{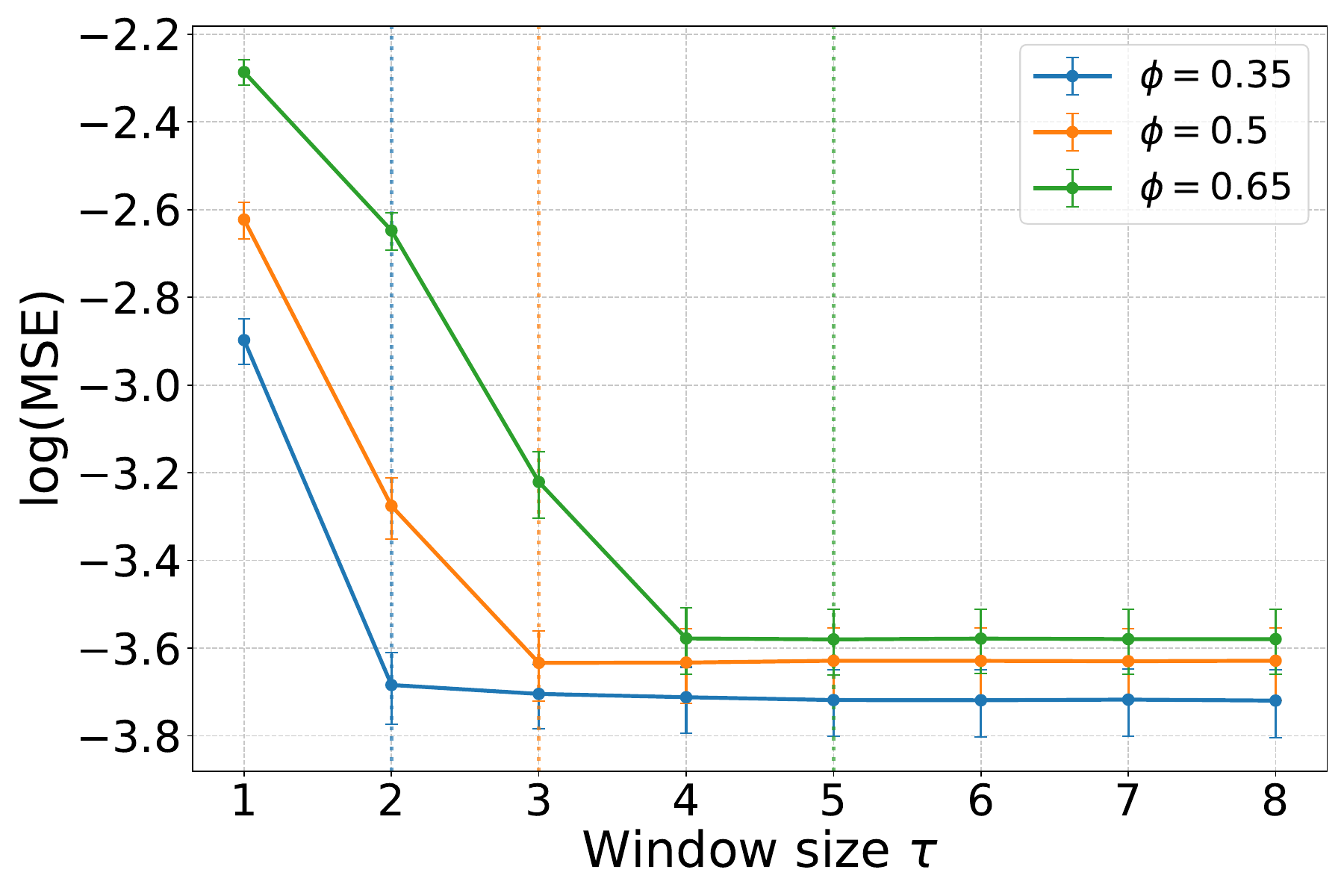}}
    \subfigure[Comparison with baseline]{\label{fig:12}\includegraphics[width=7.5cm]{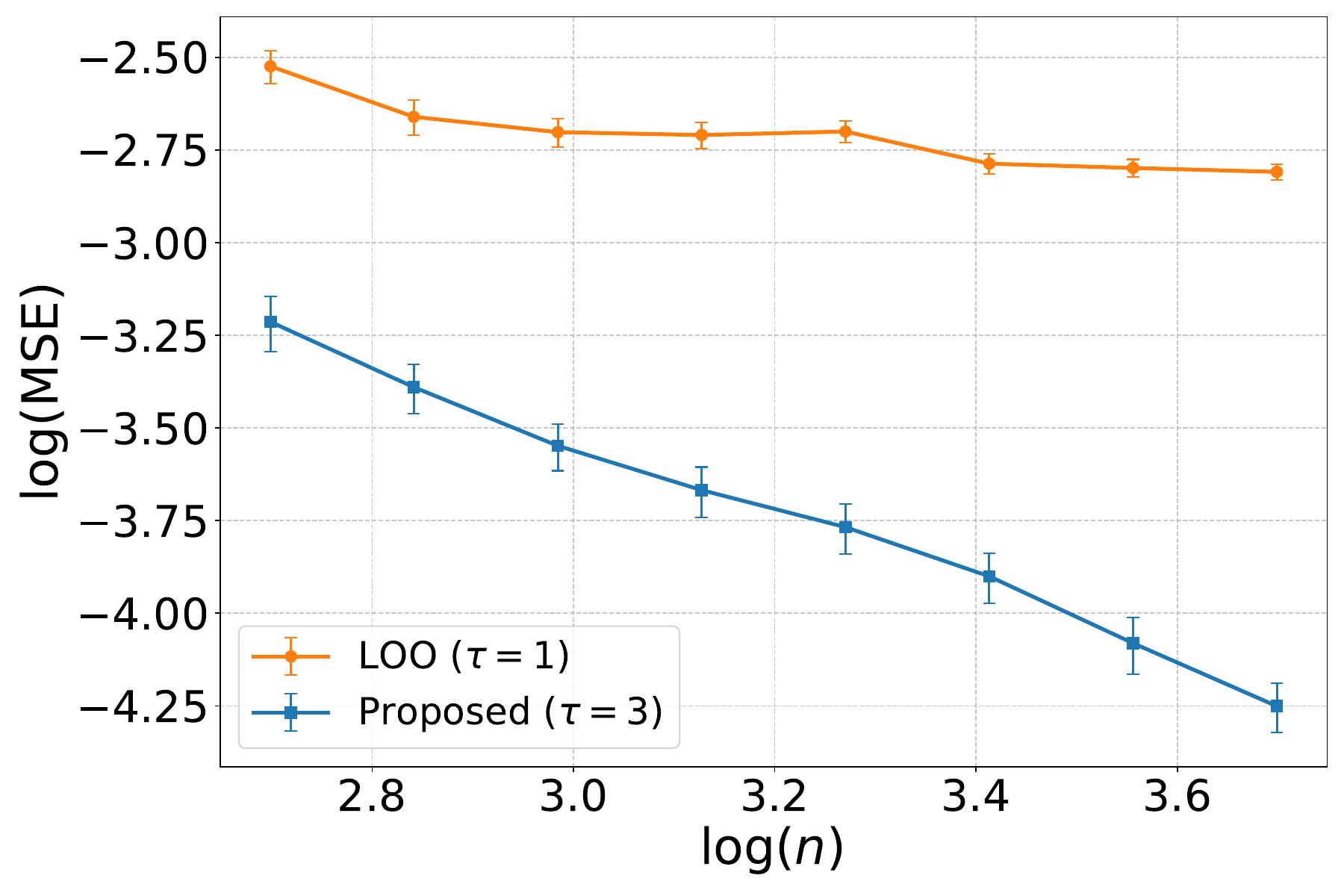}}
    \caption{Classification test error estimation on the AR($1$) process with the
    $k$-nearest neighbor rule.
    (a) $\log(\MSE)$ against the window size $\tau$ at fixed $n=1500$ and $d=600$, for
    $\phi \in \{0.35,0.5,0.65\}$; dotted lines mark the window size at which each
    curve levels off.
    (b) $\log(\MSE)$ against $n$ for $\phi=0.5$, comparing LOO ($\tau=1$) with the
    proposed estimator ($\tau=3$).}
    \label{fig:ClassificationTE_AR}
\end{figure}

Figure~\ref{fig:11} varies $\tau\in \{1,2,\ldots,8\}$ with $n=1500$ and $d=600$ for the AR process. The $\MSE$ decreases and then
levels off, and the window size at which it does so moves right as $\phi$
increases: a larger $\phi$ means a slower geometric decay of
$\beta(\tau)$, so a wider window is needed to sufficiently reduce the bias.
Figure~\ref{fig:12} varies $n \in \{499, 694, 965, 1341, 1863, 2589, 3598, 4999\}$ at $\phi=0.5$ and $d=200$ for the AR process. The window size is taken to be $\tau=3$ as predicted by Figure~\ref{fig:11} for $\phi=0.5$. The $\MSE$ for the LOO estimator plateaus, while the proposed estimator empirically attains the
$n^{-1}$ rate. Note that this is in spite of the fact that nearest-neighbors does not satisfy uniform stability  (Assumption~\ref{assmp:uniform_stability}) as required by Corollary~\ref{corollary:test_error}. 
 Appendices~\ref{sec:appendix_lr_ma} and~\ref{sec:appendix_lr_ar} repeat the
MA and AR experiments, respectively, with regularized logistic
regression in place of the $k$-nearest neighbor rule. This predictor does satisfy uniform stability.

\section{Proof of Theorem~\ref{theorem:MSE_theorem}}\label{sec:proof_MSE_theorem}

In this section we prove Theorem~\ref{theorem:MSE_theorem}.
Recall that for each index \(i\in[n]\), the local leave-a-window-out estimator corresponding to the next-token functional $f$ is given by
\[
\singleestimator(\tau) := f(X_i;\measure_{\Iset_i}),
\]
where $\Iset_i$ is defined in Eq.~\eqref{eq:index-sets}. Also, recall that our estimator of the expected next-token functional is ultimately given by
\[
\estimator(\tau) := \frac{1}{n}\sum_{i=1}^n \singleestimator(\tau).
\]
We follow a ``skipped-estimator" decomposition for the proof that was also used in~\cite{pananjady2024just,pmlr-v291-nakul25a}. For convenience, suppose that \(n\) is divisible by \(\tau\), and define the integer
\(n_0 \defn n/\tau\). For each offset \(u=0,\ldots,\tau-1\), define the ``skipped estimator"
\begin{align}
\label{eq:general_skipped_estimators}
    \estimator(\tau;u)
    \defn
    \frac{1}{n_0}\sum_{i=1}^{n_0}\widehat \theta_{\tau i-u}(\tau)
\end{align}
as the estimator that averages over the subset of local estimators that begins at index $\tau - u$ and skips over every $\tau$ indices.
With this definition in hand, we can write
\begin{align}\label{eq:skipped-estimator-decomposition}
\estimator(\tau)=\frac{1}{\tau}\sum_{u=0}^{\tau-1}\estimator(\tau;u).
\end{align}
The inequality $(a+b)^2 \leq 2(a^2 + b^2)$ and linearity of expectation yields
\begin{align*}
\MSE \big(\estimator(\tau),\estimand\big)
&=
\EE \left[\bigl(\estimator(\tau)-\estimand\bigr)^2\right] \\
&\le
2\Bigl(\estimand-\EE[\estimator(\tau)]\Bigr)^2
+
2\,\EE \left[\bigl(\estimator(\tau)-\EE[\estimator(\tau)]\bigr)^2\right].
\end{align*}
Substituting Eq.~\eqref{eq:skipped-estimator-decomposition} into the above display, we obtain
\begin{align}
\label{eq:general_bias_variance}
\MSE \big(\estimator(\tau),\estimand\big)
\le
2
\Bigl(\frac{1}{\tau}\sum_{u=0}^{\tau-1}\underbrace{\estimand-\EE[\estimator(\tau;u)]}_{\text{Bias of }\estimator(\tau;u)}\Bigr)^2
+
2\, \underbrace{\EE \left[\bigl(\estimator(\tau)-\EE[\estimator(\tau)]\bigr)^2\right]}_{\text{Variance of } \estimator(\tau)}.
\end{align}
We now bound the bias and variance terms separately.

\paragraph{Bounding the bias of \(\estimator(\tau;u)\).}
By the above definitions of the skipped estimator and the local estimator, we have
\[
\EE[\estimator(\tau;u)]
=
\frac{1}{n_0}\sum_{i=1}^{n_0}\EE \left[f(X_{\tau i - u };\measure_{\Iset_{\tau i - u}})\right].
\]
Recall that the estimand is the expected next-token functional, which is given by $\estimand = \EE[f(X_{n+1};\measure_n)]$. The triangle inequality then gives us
\begin{align}
\label{eq:bias_start_general}
\left|\estimand-\EE[\estimator(\tau;u)]\right|
\le
\frac{1}{n_0}\sum_{i=1}^{n_0}
\left|
\EE \left[f(X_{n+1};\measure_n)\right]
-
\EE \left[f(X_{\tau i-u};\measure_{\Iset_{\tau i -u}})\right]
\right|.
\end{align}
For each index \(j\in[n]\), we define the auxiliary ``switched" sequence \(\widetilde X_{\Iset_{j}}\) as below:
\begin{align*}
    \widetilde X_{\Iset_{ j}} :=
\begin{cases}
(X_{n-j+2},\ldots,X_n,X_1,\ldots,X_{n-j-\tau+1}), & \text{ if } 1\le j\le n-\tau,\\
(X_{n-j+2},\ldots,X_{n}), & \text{ if } n-\tau+1\le j\le n.
\end{cases}
\end{align*}
Let \(\widetilde\measure_{\Iset_{\tau i-u}}\) denote the empirical measure of the switched sequence \(\widetilde X_{\Iset_{\tau i-u}}\).
Then, adding and subtracting
\(\EE[f(X_{n+1};\widetilde\measure_{\Iset_{\tau i-u}})]\) in each of the summands in
Eq.~\eqref{eq:bias_start_general}, and applying the triangle inequality to each of these summands, yields
\begin{align}\nonumber
\left|\estimand-\EE[\estimator(\tau;u)]\right|
&\le
\frac{1}{n_0}\sum_{i=1}^{n_0}
\left|
\EE \left[f(X_{\tau i-u};\measure_{\Iset_{\tau i-u}})\right]
-
\EE \left[f(X_{n+1};\widetilde\measure_{\Iset_{\tau i -u}})\right]
\right| \\ \label{eq:bias_decomposition_general}
&\quad+
\frac{1}{n_0}\sum_{i=1}^{n_0}
\left|
\EE \left[f(X_{n+1};\widetilde\measure_{\Iset_{\tau i-u}})\right]
-
\EE \left[f(X_{n+1};\measure_n)\right]
\right|.
\end{align}
We first handle each of the summands in the first line of
Eq.~\eqref{eq:bias_decomposition_general}. By the definition of the total variation distance, we have
\begin{align*}
\left|
\EE \left[f(X_{\tau i-u};\measure_{\Iset_{\tau i-u}})\right]
-
\EE \left[f(X_{n+1};\widetilde\measure_{\Iset_{\tau i-u}})\right]
\right|
&\le 2\|f\|_{\infty}\,\cdot \TV\Bigl((X_{\Iset_{\tau i - u}},X_{\tau i-u}), (\widetilde X_{\Iset_{\tau i - u}},X_{n+1})
\Bigr) \\
&\overset{\1}{\le}
4 \|f\|_{\infty} \cdot \beta(\tau),
\end{align*}
where step $\1$ leverages~\citet[Proposition 1]{barber2025predictive}, which bounds \newline$\TV\Bigl((X_{\Iset_{\tau i - u}},X_{\tau i-u}), (\widetilde X_{\Iset_{\tau i - u}},X_{n+1})\Bigr)$ (defined there as the ``switch coefficient") by twice the $\beta$-mixing coefficient. Note that $(X_{\Iset_{\tau i - u}},X_{\tau i-u})$ is the sequence with point $X_{\tau i-u}$ reinstated in $X_{\Iset_{\tau i - u}}$ at its location.
Substituting this bound into Eq.~\eqref{eq:bias_decomposition_general} yields
\begin{align}
\label{eq:bias_bound_general}
\left|\estimand-\EE[\estimator(\tau;u)]\right|
\le
4 \|f\|_{\infty} \beta(\tau) + \frac{1}{n_0}\sum_{i=1}^{n_0}
\left|
\EE \left[f(X_{n+1};\widetilde\measure_{\Iset_{\tau i-u}})\right]
-
\EE \left[f(X_{n+1};\measure_n)\right]
\right|.
\end{align}
Ultimately, the bias term in Ineq.~\eqref{eq:general_bias_variance} is bounded as
\begin{align}\nonumber
    2
\Bigl(\frac{1}{\tau}\sum_{u=0}^{\tau-1}\estimand-\EE[\estimator(\tau;u)]\Bigr)^2 &\leq 2 \Bigl(4 \|f\|_{\infty} \beta(\tau)  +\frac{1}{n_0 \tau} \sum_{u=0}^{\tau-1}\sum_{i=1}^{n_0}
\left|
\EE \left[f(X_{n+1};\widetilde\measure_{\Iset_{\tau i-u}})\right]
-
\EE \left[f(X_{n+1};\measure_n)\right]
\right|\Bigr)^2 \\ \nonumber
&= 2 \Bigl(4 \|f\|_{\infty} \beta(\tau)  +\frac{1}{n } \sum_{j=1}^{n}
\left|
\EE \left[f(X_{n+1};\widetilde\measure_{\Iset_{j}})\right]
-
\EE \left[f(X_{n+1};\measure_n)\right]
\right|\Bigr)^2\\ \nonumber
&= 2 \Bigl(4 \|f\|_{\infty} \beta(\tau)  +\frac{1}{n } \sum_{j=1}^{n-\tau}
\left|
\EE \left[f(X_{n+1};\widetilde\measure_{\Iset_{j}})\right]
-
\EE \left[f(X_{n+1};\measure_n)\right]
\right| \\ \nonumber
&\qquad \qquad \qquad \qquad+\frac{1}{n } \sum_{j=n-\tau+1}^{n}
\left|
\EE \left[f(X_{n+1};\widetilde\measure_{\Iset_{j}})\right]
-
\EE \left[f(X_{n+1};\measure_n)\right]
\right|\Bigr)^2\\ \nonumber
&\overset{\1}{\leq} 2 \Bigl(4 \|f\|_{\infty} \beta(\tau)  +\frac{1}{n} \sum_{j=1}^{n-\tau}
\left|
\EE \left[f(X_{n+1};\measure_{\Iset_{j}})\right]
-
\EE \left[f(X_{n+1};\measure_n)\right]
\right|
+\frac{2\tau \|f\|_{\infty}}{n }\Bigr)^2\\ 
&\overset{\2}{\leq}2 \Bigl(4 \|f\|_{\infty} \Bigl( \beta(\tau)+ \frac{\tau}{n}\Bigr)+r_f(\tau,n)\Bigr)^2,\label{eq:final_bias}
\end{align}
where step $\1$ follows since $\widetilde \Ical_j = \Ical_{n-j-\tau+2}$ for $1 \le j \le n-\tau$, along with the trivial bound $\left|
\EE \left[f(X_{n+1};\widetilde\measure_{\Iset_{j}})\right]
-
\EE \left[f(X_{n+1};\measure_n)\right]
\right| \le 2\|f\|_{\infty}$  and step $\2$ substitutes Assumption~\ref{assmp:stability}.

\paragraph{Bounding the variance of \(\estimator(\tau)\).}
For the variance term we invoke Assumption~\ref{assmp:bd} together with the
McDiarmid-type inequality for dependent random variables that was shown by~\citet[Theorem~2.9]{paulin2015concentration}. 
We restate the theorem here for ease of reference.

\begin{theorem}[{\citep[Theorem 2.9]{paulin2015concentration}}]
\label{thm:paulin_mcdiarmid}
    Let $X^n = (X_1,\ldots,X_n)$ be a sequence of random variables, and let $\widehat{X}^m = (\widehat{X}_1,\ldots,\widehat{X}_m)$ be a partition of this sequence. Suppose we have a Marton coupling for $\widehat{X}^m$ with mixing matrix given by $\Gamma$. Let $h:\Xcal^n \to \mathbb{R}$ be a function such that
    \begin{align*}
        h(x) - h(y) \leq \sum_{i=1}^n c_i \ind{x_i \neq y_i} \text{ for all }x,y\in \Xcal^n,
    \end{align*}
    where we term $\{c_i\}_{i=1}^n$ ``bounded-differences constants".
    Now, define the cumulative bounded differences constant based on the partition $\widehat{X}^m$ as follows:
    \begin{align*}
        C_i(c) = \sum_{X_j \in \widehat{X}_i} c_j \text{ for } i \in [m].
    \end{align*}
    Define $C(c) \defn [C_1(c), \ldots, C_m(c)]^T \in \mathbb{R}^m$. Then we have, for any $t\geq 0$,
    \begin{align*}
        \Prob(|h(X^n) - \EE[h(X^n)]| \geq t) \leq 2 \exp\big( \frac{-2t^2}{\vecnorm{\Gamma \cdot C(c)}{2}^2} \big).
    \end{align*}
    
\end{theorem}
We will apply Theorem~\ref{thm:paulin_mcdiarmid} with the choice $h := \estimator(\tau)$ (for ease of notation, we henceforth drop the argument $X^n$).
For this choice, Assumption~\ref{assmp:bd} implies that for every \(i\in[n]\), the bounded-differences constants are exactly
\(
c_i = \frac{B_f}{n}.
\)
Since we have assumed that \(X^n\) is \((\gamma,s)\)-Martonizable, Definition~\ref{defn:martonizable}
guarantees that the uniform block partition
\(\widehat X^m=(\widehat X_1,\ldots,\widehat X_m)\) into \(m=\lceil n/s\rceil\)
contiguous blocks of size at most \(s\) admits a Marton coupling whose mixing
matrix \(\Gamma\) satisfies \(\opnorm{\Gamma}\le \gamma\). Therefore, the cumulative
bounded-differences constants are given by
\[
C_i(c)=\sum_{X_j\in \widehat X_i} c_j
\leq
s(\widehat X_i) \cdot \frac{B_f}{n} \text{ for all } i\in[m],
\]
where $s(\widehat X_i)$ denotes the size of the block $\widehat X_i$.
Applying Theorem~\ref{thm:paulin_mcdiarmid} to the function $h := \estimator(\tau)$ yields
\[
\Prob \big(
\left|\estimator(\tau)-\EE[\estimator(\tau)]\right|\ge t
\big)
\le
2\exp \big(
-\frac{2 t^2}{\vecnorm{\Gamma \cdot C(c)}{2}^2}
\big).
\]
Integrating the above tail bound from $0$ to $\infty$ gives us
\begin{align}
\label{eq:variance_bound_general}
\EE \left[
\big(\estimator(\tau)-\EE[\estimator(\tau)]\big)^2
\right]
\le
\vecnorm{\Gamma \cdot C(c)}{2}^2 \overset{\1}{\leq} \gamma^2 \cdot \frac{B_f^2 \cdot s}{n},
\end{align}
where step $\1$ uses \(\vecnorm{\Gamma \cdot C(c)}{2} \le \opnorm{\Gamma} \cdot \vecnorm{C(c)}{2} \le \gamma \vecnorm{C(c)}{2}\)
together with the fact that the blocks of the uniform partition are disjoint, of size at most $s$ and
cover \([n]\), so that \(\sum_{i=1}^m s(\widehat X_i)=n\) and hence
\[
\vecnorm{C(c)}{2}^2
=
\frac{B_f^2}{n^2}\sum_{i=1}^{m} s(\widehat X_i)^2
\le
\frac{B_f^2 \cdot s}{n^2}\,\sum_{i=1}^{m} s(\widehat X_i)
=
\frac{B_f^2 \cdot s}{n}.
\]

\paragraph{Completing the proof.}
Substituting Eqs.~\eqref{eq:final_bias} and
\eqref{eq:variance_bound_general} into Eq.~\eqref{eq:general_bias_variance}, we obtain
\begin{align*}
\MSE \big(\estimator(\tau),\estimand\big)
&\le
2\left(4\|f\|_{\infty}\,\Bigl( \beta(\tau)+ \frac{\tau}{n}\Bigr)+r_f(\tau,n)\right)^2
+
\frac{2 \cdot \gamma^2 \cdot B_f^2 \cdot s}{n},
\end{align*}
as claimed.
\qed

\section{Discussion}\label{sec:discussion}
We have provided a flexible estimator and analysis for next-token functionals of a sequence of dependent random variables, where the last random variable is unobserved. 
We also establish a minimax lower bound over a class of stationary finite-state ergodic Markov chains, which reveals that our estimation error rate
is sharp in its dependence on $n$ and the mixing time of the stochastic process, $\Tmix$.
Finally, our approach isolates bias and variance: the bias is controlled via the $\beta$-mixing coefficient (Definition~\ref{defn:beta_mixing}) and deletion stability, while the variance is controlled via Marton-coupling concentration for bounded-difference functionals, through the Martonizability parameters $(\gamma,s)$ (Definition~\ref{defn:martonizable}).
An interesting takeaway from our results is that in many cases, our guarantee is relatively agnostic to the choice of window size parameter $\tau$, provided it is chosen large enough.

Several open questions remain, and we elaborate on a few of them to conclude. First, proving the bound for the classification test error relied on a uniform stability assumption (Assumption~\ref{assmp:uniform_stability}). As seen in the numerical experiments, this assumption is sufficient rather than necessary. It remains open to provide sharper assumptions on the training algorithm under which our estimator succeeds. Second, we study the nearest-neighbor tail probability functional when the state space is real-valued (i.e. $\Xcal \subset \mathbb{R}$). A natural question to ask is whether we can study this functional in high dimensions. Third, obtaining the optimal dependence on the parameter $\zeta$ for the count surprise probability functional is left for future work.
Finally, we believe that our decoupling device of leaving a window out could be more broadly useful in stochastic optimization~\citep[e.g.][]{mou2024optimal,nakul2026multiscale} and property testing~\citep[e.g.][]{beran1992statistical} with dependent data.

\subsection*{Acknowledgments and declaration of AI use}

This work was supported in part by the National Science Foundation through grants CCF-2107455, CCF-2239151 and IIS-2212182, and a Google Research Scholar Award. 

Anthropic's Opus 5 was used to set up the experiments and to polish the writing in Section~\ref{sec:numerical_results}, Appendix~\ref{sec:appendix_lr} and for the plots in Section~\ref{sec:estimator}. OpenAI's GPT 5.6 Sol was used to proofread the paper. The
authors assume full responsibility for the contents of the paper.

\small
\bibliographystyle{abbrvnat}
\bibliography{reference-template}

\normalsize
\appendix

\section{Proofs}\label{sec:proofs}

In this section, we present the proofs of all our remaining results.

\subsection{Proof of Corollary~\ref{corollary:surprise_probability}}\label{sec:proof_corr_surprise_probability}

In this section we provide the proof of Corollary~\ref{corollary:surprise_probability}, which is Theorem~\ref{theorem:MSE_theorem} applied to the surprise probability functional defined in Section~\ref{sec:surprise} as
\(
    f(X_{n+1};\measure_n) := \ind{X_{n+1} \notin \bX_{[n]}}.
\)
Recall that we denoted by $\surprise = \Prob(X_{n+1} \notin \bX_{[n]})$ the estimand (which is the expected surprise probability functional), and by $\Shat(\tau)$ our surprise probability estimator.
Being an indicator, the surprise probability functional satisfies $\|f\|_{\infty} = 1$. We claim that, for this functional, Assumption~\ref{assmp:stability} holds with $r_{f}(\tau,n) = \frac{\tau}{n}$ and Assumption~\ref{assmp:bd} holds with $B_f = 3$. Therefore, we can invoke Theorem~\ref{theorem:MSE_theorem} to obtain
\begin{align*}
        \MSE\big(\Shat(\tau),\surprise\big) \leq 64 \cdot \Bigl( \beta(\tau)+ \frac{\tau}{n}\Bigr)^2 + 4 \cdot \left( \frac{\tau}{n} \right)^2 + \frac{18 \cdot \gamma^2 \cdot s}{n},
    \end{align*}
where we used the inequality $(a+b)^2 \leq 2(a^2 +b^2)$, which yields $2\left(4\Bigl( \beta(\tau)+ \frac{\tau}{n}\Bigr)+r_f(\tau,n)\right)^2 \le 64\Bigl( \beta(\tau)+ \frac{\tau}{n}\Bigr)^2 + 4 r_f(\tau,n)^2$.
This is exactly the bound claimed in Corollary~\ref{corollary:surprise_probability}.
It remains to verify Assumptions~\ref{assmp:stability} and~\ref{assmp:bd}, which we do below.

\subsubsection{Verification of Assumption~\ref{assmp:stability}}

Recall that Assumption~\ref{assmp:stability} states that
\[
\frac{1}{n}\sum_{i=1}^{n}
\left|
\EE \left[f(X_{n+1};\measure_n)\right]
-
\EE \left[f(X_{n+1};\measure_{\Iset_i})\right]
\right|
\le r_f(\tau,n).
\]
For the surprise probability functional, we have
\[
\frac{1}{n}\sum_{i=1}^{n}
\left|
\EE \left[f(X_{n+1};\measure_n)\right]
-
\EE \left[f(X_{n+1};\measure_{\Iset_i})\right]
\right|
= \frac{1}{n}\sum_{i=1}^n\left|\Prob(X_{n+1} \notin \bX_{\Iset_i}) - \Prob(X_{n+1} \notin \bX_{[n]}) \right|,
\]
where the sets $\Dset_i$ and $\Iset_i$ are defined in Eq.~\eqref{eq:index-sets}.
From here, the proof will resemble the steps in~\cite[Section 7.2.1]{pananjady2024just}.
Recall that we are working with skipped estimators with offset $u \in \{0,\ldots,\tau-1\}$ as in the proof of Theorem~\ref{theorem:MSE_theorem}, and that we defined $n_0 = n/\tau$ as shorthand. The triangle inequality then gives us
\begin{align}\label{eq:intermediate_step}
\frac{1}{n}\sum_{i=1}^n\left|\Prob(X_{n+1} \notin \bX_{\Iset_i}) - \Prob(X_{n+1} \notin \bX_{[n]}) \right|
\leq \frac{1}{\tau} \sum_{u=0}^{\tau-1} \frac{1}{n_0} \sum_{i=1}^{n_0} \left|\Prob(X_{n+1} \notin \bX_{\Iset_{\tau i-u}}) - \Prob(X_{n+1} \notin \bX_{[n]}) \right|.
\end{align}
Using \citet[Lemma 11]{pananjady2024just}, for any $i \in [n]$ we have
\begin{align*}
    \left|\Prob(X_{n+1} \notin \bX_{\Iset_i}) - \Prob(X_{n+1} \notin \bX_{[n]}) \right| \leq \EE\left[ \ind{X_{n+1} \notin \bX_{\Iset_i}}\ind{X_{n+1} \in \bX_{\Dset_i}} \right].
\end{align*}
Substituting the above display into Eq.~\eqref{eq:intermediate_step}, we obtain
\begin{align}\label{eq:step2}
\frac{1}{n}\sum_{i=1}^n\left|\Prob(X_{n+1} \notin \bX_{\Iset_i}) - \Prob(X_{n+1} \notin \bX_{[n]}) \right| \leq \frac{1}{\tau} \sum_{u=0}^{\tau-1} \frac{1}{n_0} \sum_{i=1}^{n_0} \EE\left[ \ind{X_{n+1} \notin \bX_{\Iset_{\tau i - u}}}\ind{X_{n+1} \in \bX_{\Dset_{\tau i - u}}} \right].
\end{align}
Since we are working with skipped estimators, the blocks $\{\Dset_{\tau i - u}\}_{i=1}^{n_0}$ are non-overlapping, so that for any fixed $i \in [n]$ and $u \in \{0,\ldots,\tau-1\}$, we have
\begin{align}
    \label{eq:subset}
    \bigcup_{i' \in [n_0] \setminus i} \Dset_{\tau i'-u} \subset \Iset_{\tau i - u}.
\end{align}
Now, suppose for some $i \in [n_0]$ we have $\ind{X_{n+1} \notin \bX_{\Iset_{\tau i - u}}}\ind{X_{n+1} \in \bX_{\Dset_{\tau i - u}}} =1$. 
(If no such index exists, we are done.)
This means that $X_{n+1} \notin \bX_{\Iset_{\tau i - u}}$. Thus, we have
\begin{align*}
    \sum_{i' \in [n_0] \setminus i} \ind{X_{n+1} \notin \bX_{\Iset_{\tau i' - u}}}\ind{X_{n+1} \in \bX_{\Dset_{\tau i' - u}}} &\leq \sum_{i' \in [n_0] \setminus i} \ind{X_{n+1} \in \bX_{\Dset_{\tau i' - u}}} \\\\
    &\overset{\1}{\leq} \ind{X_{n+1} \in \bX_{\Iset_{\tau i - u}}} = 0,
\end{align*}
where step $\1$ follows from Eq.~\eqref{eq:subset}.
Hence, $\sum_{i=1}^{n_0} \EE\left[ \ind{X_{n+1} \notin \bX_{\Iset_{\tau i - u}}}\ind{X_{n+1} \in \bX_{\Dset_{\tau i - u}}} \right] \leq 1$ for each offset $u \in \{0,\ldots,\tau-1\}$.
Substituting this inequality into Ineq.~\eqref{eq:step2} yields
\begin{align*}
    \frac{1}{n}\sum_{i=1}^n\left|\Prob(X_{n+1} \notin \bX_{\Iset_i}) - \Prob(X_{n+1} \notin \bX_{[n]}) \right| \leq \frac{1}{\tau}\sum_{u=0}^{\tau-1} \frac{1}{n_0} = \frac{1}{n_0} = \frac{\tau}{n}.
\end{align*}
Thus, we have shown that Assumption~\ref{assmp:stability} is satisfied with $r_{f}(\tau,n) = \frac{\tau}{n}$.

\subsubsection{Verification of Assumption~\ref{assmp:bd}}
Next, we verify Assumption~\ref{assmp:bd}.
Recall that $X^n = (X_1, \ldots, X_n)$ is the original sequence and $X^{(k)}(a) = (X_1,\ldots,X_{k-1},a,X_{k+1},\ldots,X_n)$ is the sequence obtained after replacing the $k$-th symbol $X_k$ with the symbol $a \in \Xcal$. 
We overload notation and use $\Shat(X^n), \Shat(X^{(k)}(a)),\Shat(X^{(k)}(b))$ to refer to the surprise probability estimator when deployed on the original sequence $X^n$ and the modified sequences $X^{(k)}(a)$ and $X^{(k)}(b)$ respectively (note that we have dropped the window size $\tau$ for clarity).
At times, we drop the sequence argument and simply refer to $\Shat$ or $\Shat_{(j)}$ when the sequence on which the estimator is applied is clear from context.
Then, we have the following bounded differences-type lemma for our estimator of the surprise probability functional.
\begin{lemma}\label{lemma:bdd_lemma_surprise}
    For any sequence $X^n \in \Xcal^n$ and any window size $\tau \in \{1,\ldots, n-1\}$, we have
    \begin{align*}
        \sup_{a,b \in \Xcal}\sup_{k \in [n]}\left|\Shat(X^{(k)}(a))-\Shat(X^{(k)}(b)) \right| \leq \frac{3}{n}.
    \end{align*}
\end{lemma}
Lemma~\ref{lemma:bdd_lemma_surprise} directly implies that the surprise probability estimator $\Shat(\tau)$ satisfies Assumption~\ref{assmp:bd} with $B_f = 3$, which is precisely our claim. The remainder of this section proves this lemma.
\paragraph{Proof of Lemma~\ref{lemma:bdd_lemma_surprise}.}
    Let us define the following quantity for some fixed sequence $X^n$, pair of symbols $a,b \in \Xcal$ and index $k \in [n]$:
    \begin{align*}
        V^{(k)}(a,b) &:= \sum_{j =1}^{n} \ind{\Shat_{(j)}(X^{(k)}(a))\neq \Shat_{(j)}(X^{(k)}(b))}, \text{ where we recall that } 
        \Shat_{(j)}(X^n) &= \ind{X_{j} \notin \bX_{\Iset_{j}}}.
    \end{align*}
    We then have
    \begin{align*}
        \left|\Shat(X^{(k)}(a))-\Shat(X^{(k)}(b)) \right| &\leq \frac{1}{n} \sum_{j=1}^n \left|\Shat_{(j)}(X^{(k)}(a)) - \Shat_{(j)}(X^{(k)}(b))\right| \\
        &= \frac{1}{n} \sum_{j=1}^n \ind{\Shat_{(j)}(X^{(k)}(a))\neq \Shat_{(j)}(X^{(k)}(b))} =: \frac{V^{(k)}(a,b)}{n}.
    \end{align*}
   Thus, it suffices to prove that $V^{(k)}(a,b) \leq 3$ for any pair of symbols $a,b \in \Xcal$ and any index $k \in [n]$.
   We will do this by enumerating the maximum possible number of indices $j \in [n]$ for which $\Shat_{(j)}(X^{(k)}(a)) \neq \Shat^{(j)}(X^{(k)}(b))$. Observe that, for any sequence, $\Shat_{(j)} =1$ necessitates that the symbol $X_j$ occurs for the first time at position $j$. For the second occurrence and beyond (i.e. considering all indices $j' > j$ such that $X_{j'} = X_j$), we always have $\Shat_{(j')} = \ind{X_{j'} \notin \bX_{\Iset_{j'}}} = 0$. This is because $\Dset_{j'} = \{j',\ldots,(j'+\tau-1)\wedge n\}$ implies that $j \in \Iset_{j'}$, so that $X_{j'} = X_j \in \bX_{\Iset_{j'}}$.
    We claim that if we replace one occurrence of $a$ by $b$ (in going from the original sequence $X^{(k)}(a)$ to the modified sequence $X^{(k)}(b)$) then at most $3$ indicators among $\{\Shat_{(j)}\}_{j=1}^n$ can change: two corresponding to the first and second occurrence of $a$ and one corresponding to the first occurrence of $b$. 
    To prove this claim, we analyze the following four cases, which together encompass all possibilities for the index $k \in [n]$, the pair of symbols $(a,b)$ and the original sequence $X^n$.
    \begin{enumerate}
        \item \textbf{Case I:} Let $k$ be the index of the \emph{first} occurrence of $a$ in the original sequence and the \emph{first} occurrence of $b$ in the modified sequence, i.e. $X_j \notin \{a,b\}$ for all $j < k$ and $X_k = a$.
        Essentially, in this case we are replacing the first occurrence of $a$ in $X^{(k)}(a)$ by the first occurrence of $b$ in $X^{(k)}(b)$. In this case, there are only $3$ indicators that can change:
    \begin{enumerate}
        \item The indicator for the first occurrence of $a$ in the original sequence can change, i.e. we can have $\Shat_{(k)}(X^{(k)}(a)) \neq \Shat_{(k)}(X^{(k)}(b))$.
        \item The indicator for the second occurrence of $a$ in the original sequence can change, since in the modified sequence $X^{(k)}(b)$ it is the first occurrence of $a$.
        \item The indicator for the first occurrence of $b$ in the original sequence can change, since in the modified sequence $X^{(k)}(b)$ it is the second occurrence of $b$.
    \end{enumerate}
    The indicators corresponding to the other indices will not change in this case, i.e. we will have $\Shat_{(j)}(X^{(k)}(a)) = \Shat_{(j)}(X^{(k)}(b))$.
    To argue this, we enumerate the two alternative possibilities for the indices below.
    \begin{enumerate}
        \item The index $j$ is such that $X_j \notin \{a,b\}$. Then, changing $X_k$ from $a$ to $b$ does not affect the indicator $\ind{X_j \notin \bX_{\Iset_j}}$, implying that $\Shat_{(j)}(X^{(k)}(a)) = \Shat_{(j)}(X^{(k)}(b))$.
        \item The index $j > k$ is a third-or-later occurrence of the symbol $a$, or a second-or-later occurrence of the symbol $b$, in the original sequence. Then, we have $\Shat_{(j)}(X^{(k)}(a)) = \Shat_{(j)}(X^{(k)}(b)) = 0$.
    \end{enumerate}
    Thus, $V^{(k)}(a,b) \leq 3$ in this case.
        \item \textbf{Case II:} Let $k$ be the index of the \emph{second} occurrence of $a$ in the original sequence and the \emph{first} occurrence of $b$ in the modified sequence; this means that $X_j \neq b$ for all $j < k$ and $X_k = a$.
        Essentially, in this case we are replacing the second occurrence of $a$ by the first occurrence of $b$. In this case, there are again only $3$ indicators that can change:
        \begin{enumerate}
            \item The indicator for the first occurrence of $a$ in the original sequence can change. In particular, this event can occur in the situation where there are only two occurrences of $a$ in the original sequence $X^{(k)}(a)$.
            \item The indicator for the second occurrence of $a$ in the original sequence can change, i.e. we can have $\Shat_{(k)}(X^{(k)}(a)) \neq \Shat_{(k)}(X^{(k)}(b))$. This is because the index $k$ is the first occurrence of $b$ in the modified sequence $X^{(k)}(b)$.
            \item The indicator for the first occurrence of $b$ in the original sequence can change, since in the modified sequence $X^{(k)}(b)$ it is the second occurrence of $b$.
        \end{enumerate}
        The indicators corresponding to the other indices will not change in this case, i.e.~we will have $\Shat_{(j)}(X^{(k)}(a)) = \Shat_{(j)}(X^{(k)}(b))$.
        To argue this, we enumerate the two alternative possibilities for the indices below.
        \begin{enumerate}
            \item The index $j$ is such that $X_j \notin \{a,b\}$. Then, changing $X_k$ from $a$ to $b$ does not affect the indicator $\ind{X_j \notin \bX_{\Iset_j}}$, implying that $\Shat_{(j)}(X^{(k)}(a)) = \Shat_{(j)}(X^{(k)}(b))$.
            \item The index $j > k$ is a third-or-later occurrence of the symbol $a$, or a second-or-later occurrence of the symbol $b$, in the original sequence. Then, we have $\Shat_{(j)}(X^{(k)}(a)) = \Shat_{(j)}(X^{(k)}(b)) = 0$.
        \end{enumerate}
        Thus, $V^{(k)}(a,b) \leq 3$ in this case.
        \item \textbf{Case III:} Let $k$ be the index of the \emph{first} occurrence of $a$ in the original sequence and the \emph{second} occurrence of $b$ in the modified sequence; this means that $X_j \neq a$ for all $j < k$, $X_{j_1} = b$ for exactly one index $j_1 < k$, and $X_k = a$. Essentially, in this case we are replacing the first occurrence of $a$ by the second occurrence of $b$. In this case, there are again only $3$ indicators that can change:
        \begin{enumerate}
            \item The indicator for the first occurrence of $b$ in the original sequence can change. In particular, if $j_1$ was the only occurrence of $b$ in the original sequence, we could have $\Shat_{(j_1)}(X^{(k)}(a)) = 1$ but $\Shat_{(j_1)}(X^{(k)}(b)) = 0$.
            \item The indicator for the first occurrence of $a$ in the original sequence can change,~i.e. we can have $\Shat_{(k)}(X^{(k)}(a)) \neq \Shat_{(k)}(X^{(k)})(b)$. This is because the index $k$ is the second occurrence of $b$ in the modified sequence $X^{(k)}(b)$.
            \item The indicator for the second occurrence of $a$ in the original sequence can change, since in the modified sequence it is the first occurrence of $a$.
        \end{enumerate}
        The indicators corresponding to the other indices will not change in this case, i.e.~we will have $\Shat_{(j)}(X^{(k)}(a)) = \Shat_{(j)}(X^{(k)}(b))$.
        To argue this, we enumerate the two alternative possibilities for the indices below.
        \begin{enumerate}
            \item The index $j$ is such that $X_j \notin \{a,b\}$. Then, changing $X_k$ from $a$ to $b$ does not affect the indicator $\ind{X_j \notin \bX_{\Iset_j}}$, implying that $\Shat_{(j)}(X^{(k)}(a)) = \Shat_{(j)}(X^{(k)}(b))$.
            \item The index $j > k$ is a third-or-later occurrence of the symbol $a$, or a second-or-later occurrence of the symbol $b$. Then, we have $\Shat_{(j)}(X^{(k)}(a)) = \Shat_{(j)}(X^{(k)}(b)) = 0$.
        \end{enumerate}
        Thus, $V^{(k)}(a,b) \leq 3$ in this case.
        \item \textbf{Case IV:} Finally, let $k$ be the index of a second or later occurrence of $a$ in the original sequence and the index of a second or later occurrence of $b$ in the modified sequence; this means that $X_{j_a} = a$ for at least one $j_a < k$ and $X_{j_b} = b$ for at least one $j_b < k$. Essentially, in this case we are replacing a second-or-later occurrence of $a$ by a second-or-later occurrence of $b$. In this case only the following two indicators can change:
        \begin{enumerate}
            \item The indicator for the first occurrence of $a$ can change. In particular, if $j_a$ and $k$ were the only two occurrences of $a$ in the original sequence, we could have $\Shat_{(j_a)}(X^{(k)}(a)) = 0$ but $\Shat_{(j_a)}(X^{(k)}(b)) = 1$.
            \item The indicator for the first occurrence of $b$ can change. In particular, if $j_b$ was the only occurrence of $b$ in the original sequence, we could have $\Shat_{(j_b)}(X^{(k)}(a)) = 1$ but $\Shat_{(j_b)}(X^{(k)}(b)) = 0$.
            \end{enumerate}
        The indicators corresponding to the other indices will not change in this case, i.e.~we will have $\Shat_{(j)}(X^{(k)}(a)) = \Shat_{(j)}(X^{(k)}(b))$.
        To argue this, we enumerate the three alternative possibilities for the indices below.
        \begin{enumerate}
            \item The index $j$ is such that $X_j \notin \{a,b\}$. Then, changing $X_k$ from $a$ to $b$ does not affect the indicator $\ind{X_j \notin \bX_{\Iset_j}}$, implying that $\Shat_{(j)}(X^{(k)}(a)) = \Shat_{(j)}(X^{(k)}(b))$.
            \item For the index $k$, we have $\Shat_{(k)}(X^{(k)}(a)) = \Shat_{(k)}(X^{(k)}(b)) = 0$.
            \item The index $j \neq k$ is a second or later occurrence of either $a$ or $b$ in the original sequence.
            In this case, we again have $\Shat_{(j)}(X^{(k)}(a)) = \Shat_{(j)}(X^{(k)}(b)) = 0$.
        \end{enumerate}
        Thus, $V^{(k)}(a,b) \leq 2$ in this case.
    \end{enumerate}
    Combining the four cases, we have $V^{(k)}(a,b) \leq 3$ for any $a,b \in \Xcal$. This completes the proof of the lemma, and therefore the verification of Assumption~\ref{assmp:bd}.

\qed
\subsection{Proof of Proposition~\ref{thm:minimax_risk}}
\label{sec:proof_thm_minimax}

We prove the lower bound in a series of steps. We first construct a
family of i.i.d.\ distributions inspired by the construction in \citet{rajaraman2017minimax} whose surprise probabilities are hard to estimate. Second, from any such i.i.d. distribution from this family, we construct a stationary ergodic Markov chain with mixing time $\Tmix$; this induces a corresponding family of Markov chains. Third, we relate the surprise probability of this Markov chain to the surprise probability of the corresponding i.i.d. distribution. Finally, we apply Le Cam's two-point method to obtain a minimax lower bound on the rate of estimation of the surprise probability over the constructed family of Markov chains.

\paragraph{Step 1: A hard family of i.i.d.\ surprise probabilities.}

Consider the finite alphabet $\Ycal:=\{0,1,\ldots,k\} \text{ with } k:=16n$.
For any $q \in [0,1]$, we can define the distribution $\pi_q \in \Delta(\Ycal)$ as follows:
\[
    \pi_q(0)=q,
        \text{ and }
        \pi_q(y)=\frac{1-q}{k}
        \text{ for } y\in[k].
\]
This is a ``spike-and-uniform" distribution that puts mass equal to $q$ on $0$ and distributes the remaining mass equal to $1 - q$ uniformly on the rest of the alphabet, i.e.~$\{1,\ldots,k\}$.
Then, we define the family of such ``spike-and-uniform" distributions parameterized by $q$ as 
\(
    \mathcal P
    :=
    \left\{
        \pi_q\in\Delta(\Ycal):
        q\in[1/2,3/4]
    \right\}.
\)
For $m\in\{0,\ldots,n\}$, let
\(
    Y_1,\ldots,Y_{m+1}
    \stackrel{\mathrm{i.i.d.}}{\sim}\pi_q
\)
and define the i.i.d. surprise probability as
\(
    \Siid(m;q)
    :=
    \Prob \big(
        Y_{m+1}\notin\bY_{[m]}
    \big).
\)
A direct calculation gives us
\begin{align}
    \Siid(m;q)
    &=
    \sum_{y\in\Ycal}
    \pi_q(y)\left(1-\pi_q(y)\right)^m
    \nonumber\\
    &=
    q(1-q)^m
    +
    (1-q)
    \left(
        1-\frac{1-q}{k}
    \right)^m.
    \label{eq:Siid}
\end{align}
The following lemma provides a lower bound on the difference in the i.i.d. surprise probabilities for two choices of parameter $p > q$ that is proportional to the difference in the parameters, i.e. $(p-q)$.

\begin{lemma}
\label{lemma:rate_q}
For $\tfrac12\le q<p\le\tfrac34$ and $1\le m\le n$, we have
\[
    \Siid(m;q)-\Siid(m;p)
    \ge
    \frac9{10}(p-q).
\]
\end{lemma}
We prove this lemma in Appendix~\ref{sec:pf_lemma_rate_q}.

\paragraph{Step 2: Construction of an ergodic Markov chain from an i.i.d. distribution.}

Fix an i.i.d. distribution $\pi_q\in\mathcal P$ and an integer-valued parameter $\ell \geq 2$. Consider the Markov chain whose state at step $t$ is given by the two-dimensional tuple
\[
    X_t=(Y_t,L_t)
    \in
    \Xcal:=\Ycal\times[\ell].
\]
Above, the first entry in the tuple $Y_t$ takes values in the finite alphabet $\Ycal$ that we defined in Step 1 of the proof.
The second entry corresponds to a ``block length counter"; in particular, $L_t$ denotes the remaining length of the current block, and the event $L_t = 1$ corresponds to the start of a new block at step $t+1$.
The transition rule of this Markov chain is given by
\[
    X_{t+1}
    =
    \begin{cases}
        (Y_t,L_t-1), & L_t>1,\\[1mm]
        (Y',L'), & L_t=1,
    \end{cases}
\]
where
\(
    Y'\sim\pi_q,
    L'\sim\UNIF([\ell]),
\)
and $Y' \perp L'$. In words, every new block begins with an independent label from
$\pi_q$ and an independent length uniformly distributed on $[\ell]$.
Observe that the label remains fixed within the block while the counter deterministically
counts down to $1$.
The following lemma states the salient properties of this construction that we will utilize in subsequent steps.

\begin{lemma}[Ergodic block Markov chain]
\label{lemma:block_MC}
For every integer $\ell\ge2$ and $\pi_q\in\mathcal P$, the Markov chain above is
finite-state, irreducible, and aperiodic. Its stationary distribution is
\[
    \mu_q(y,r)
    =
    \pi_q(y)\alpha_r,
    \text{ where }
    \alpha_r
    :=
    \frac{2(\ell-r+1)}{\ell(\ell+1)},
\]
and its mixing time satisfies
\(
    \Tmix \leq \ell.
\)
\end{lemma}

We prove Lemma~\ref{lemma:block_MC} in Appendix~\ref{sec:pf_lemma_block_MC}. 

Finally, we specify the initialization process for the chain. 
As in the rest of this paper, we initialize
the chain as $X_1 \sim \mu_q$, which ensures that the resulting process is stationary.

\paragraph{Step 3: Relating the Markov-chain surprise probability to the
i.i.d.\ surprise probability.}

Let us define the index of the block that contains step $t$ as 
\[
B_t \defn 1 + \sum_{s=1}^{t-1} \ind{L_s = 1}.
\]
We define the random variable $\niid \defn B_n$ as the total number of blocks intersecting the trajectory $X^n$ of length $n$.
According to this notation, each block $b$ carries an independent label
\(
    Y_b\sim\pi_q,
\)
and the observed state can be alternatively written as
\(
    X_t=(Y_{B_t},L_t).
\)
For each block counter $r\in[\ell]$, define
\[
    \Ical_r
    :=
    \{t\in[n]:L_t=r\},
    \text{ and }
    N_r
    :=
    |\Ical_r|
    =
    \sum_{t=1}^n\ind{L_t=r}.
\]
Observe that $\Ical_r$ denotes the subset of steps whose states have block counter equal to $r$, while $N_r$ denotes the size of this subset.
These will be central objects in our argument to relate the Markov chain surprise probability to the i.i.d. surprise probability.

First, we claim that the following two events are equivalent:
\begin{align}\label{eq:equivalence}
    \{X_{n+1}\notin\bX_{[n]}\}
    \equiv
    \left\{
        Y_{B_{n+1}}
        \notin
        \{Y_{B_t}:t\in\Ical_{L_{n+1}}\}
    \right\}.
\end{align}
To see the equivalence, first suppose that
\(
    X_{n+1}\notin\bX_{[n]}.
\)
Consider any $t\in\Ical_{L_{n+1}}$. By definition of
$\Ical_{L_{n+1}}$, we have
\(
    L_t=L_{n+1}.
\)
Since $X_t \neq X_{n+1}$ is equivalent to $(Y_{B_t},L_t) \neq (Y_{B_{n+1}},L_{n+1})$, these steps have to satisfy
\(
    Y_{B_t}\neq Y_{B_{n+1}}
    \text{ for every }t\in\Ical_{L_{n+1}}.
\)
This implies that 
\[
    Y_{B_{n+1}}
    \notin
    \{Y_{B_t}:t\in\Ical_{L_{n+1}}\}.
\]
Conversely, suppose that
\(
    Y_{B_{n+1}}
    \notin
    \{Y_{B_t}:t\in\Ical_{L_{n+1}}\}.
\)
Consider any $t\in[n]$. If
$t\in\Ical_{L_{n+1}}$, then
$L_t=L_{n+1}$, but by assumption
\(
    Y_{B_t}\neq Y_{B_{n+1}},
\)
so we must have $X_t\neq X_{n+1}$. If instead
$t\notin\Ical_{L_{n+1}}$, then
\(
    L_t\neq L_{n+1},
\)
and therefore $X_t\neq X_{n+1}$. This implies that $X_t\neq X_{n+1}$ for every $t\in[n]$, which gives us
\[
    X_{n+1}\notin\bX_{[n]}.
\]
This establishes both directions of the equivalence and proves the claim~\eqref{eq:equivalence}.

Note that $Y_{B_{n+1}}$ and $\{Y_{B_t}:t\in \Ical_{L_{n+1}}\}$ belong to different blocks and thus, are $N_{L_{n+1}}+1$ i.i.d. draws from $\pi_q$. Recall from Step~1 that $\Siid(m;q)$ is the surprise probability for
$m+1$ independent labels drawn from $\pi_q$. Therefore, writing as shorthand
\(
    \surprise(n;q)
    :=
    \Prob(X_{n+1}\notin\bX_{[n]}),
\)
we obtain
\begin{align}
    \surprise(n;q)
    =
    \EE \left[
        \Siid(N_{L_{n+1}};q)
    \right].
    \label{eq:MC_surprise_representation}
\end{align}
Note that since the block counters $\{L_t\}_{t\in [n]}$ are independent of the labels $\{Y_t\}_{t\in [n]}$, the distribution of $N_{L_{n+1}}$ is also independent of
$q$.

The ultimate goal of this step is to lower bound the difference between the Markov chain surprise probabilities corresponding to parameters $p > q$ by a quantity proportional to the difference $p - q$, in a manner analogous to Lemma~\ref{lemma:rate_q}.
We will establish this in the forthcoming Eq.~\eqref{eq:MC_surprise_separation}.
The equivalence to an i.i.d. surprise probability established in Eq.~\eqref{eq:MC_surprise_representation} almost enables us to do this, but the corresponding i.i.d. surprise probability is over the \emph{random sequence length} $N_{L_{n+1}}$, and we need to ensure that this sequence length is more likely than not to be non-zero.
(Indeed, if $m = 0$, we trivially have $\Siid(m;q) = \Siid(m;p) = 1$.)
In particular, we show in the following lemma that the probability that $N_{L_{n+1}} = 0$ (i.e. that $L_{n+1}$ has not appeared at all in the first $n$ observations) is at most $1/2$.
\begin{lemma}\label{lemma:probability_bound}
    The probability that $L_{n+1}$ has not appeared in the first $n$ observations is bounded as
    \[
    \Prob(N_{L_{n+1}} = 0) \leq \frac{1}{2}.
    \]
\end{lemma}
We prove Lemma~\ref{lemma:probability_bound} in Appendix~\ref{sec:proof_probability_bound}.
Combining Eq.~\eqref{eq:MC_surprise_representation},
Lemma~\ref{lemma:rate_q}, Lemma~\ref{lemma:probability_bound}, and the fact that
$\Siid(0;q)=\Siid(0;p)=1$, we obtain
\begin{align}
    \surprise(n;q)-\surprise(n;p)
    &=
    \EE \left[
        \Siid(N_{L_{n+1}};q)
        -
        \Siid(N_{L_{n+1}};p)
    \right]
    \nonumber\\
    &\ge
    \frac9{10}(p-q)
    \Prob(N_{L_{n+1}}\ge1)
    \nonumber\\
    &\ge
    \frac9{20}(p-q).
    \label{eq:MC_surprise_separation}
\end{align}
This shows that the difference in the surprise probabilities of the Markov chains parameterized by $p > q$ is lower bounded by a positive quantity that is proportional to the difference in parameters $(p - q)$.

\paragraph{Step 4: Applying Le Cam's method.}

Denote the probability distribution on the trajectory $X^n$ generated according to the Markov chain construction of Step 2, parameterized by $q$, as $\Lcal_q^n$.
Consider two choices of parameter $p > q$ satisfying $p, q \in [1/2,3/4]$.
In order to apply Le Cam's method to these two instances, we need to upper bound the total variation distance between the probability distributions of these instances, i.e.~we need to upper bound $\TV(\Lcal_q^n,\Lcal_p^n)$.

Pinsker's inequality states that $\TV(\Lcal_q^n,\Lcal_p^n) \leq \sqrt{\frac{1}{2} \KL(\Lcal_q^n\,\|\,\Lcal_p^n)}$, so it will suffice to upper bound the KL-divergence.
First, we apply the chain rule of KL-divergence~\citep[Section~2.5]{cover_thomas_2006_eit} by conditioning with respect to the block length process $\{L_1,\ldots,L_n\}$.
Noting that $\{L_1,\ldots,L_n\}$ is identically distributed under the probability distributions $\Lcal_q^n$ and $\Lcal_p^n$, we obtain
\begin{align} \nonumber
    \KL(\Lcal_q^n\,\|\,\Lcal_p^n)
    &= \EE\left[ \KL(\Lcal_q^n|L_1,\ldots,L_n\,\|\,\Lcal_p^n|L_1,\ldots,L_n)\right] \\ 
    &\overset{\1}{=} \EE\left[ \niid \cdot \KL(\pi_q \|\ \pi_p)\right]
    =\EE[\niid]\,
    \KL(\pi_q\,\|\,\pi_p).
    \label{eq:MC_KL}
\end{align}
Step $\1$ in the above follows because, conditioned on the block length process $\{L_1,\ldots,L_n\}$, the trajectory $X^n$ is determined by $\niid$ independent block labels that are drawn from distribution $\pi_q$ and $\pi_p$ respectively; moreover, $\niid$ is a deterministic function of $\{L_1,\ldots,L_n\}$.

Working from Eq.~\eqref{eq:MC_KL}, we need to characterize $\EE[\niid]$, which is simply the expected number of blocks intersecting the observed trajectory $X^n$.
By definition, a new block begins immediately after any step $t$ satisfying $L_t=1$. Consequently, we have
\(
    \niid
    =
    1+
    \sum_{t=1}^{n-1}\ind{L_t=1}.
\)
By Lemma~\ref{lemma:block_MC},
\(
    \Prob(L_t=1)
    =
    \alpha_1
    =
    \frac{2}{\ell+1}.
\)
Hence, by linearity of expectation, we have
\begin{align}
    \EE[\niid]
    &=
    1+\frac{2(n-1)}{\ell+1}
    \le
    \frac{5n}{2\ell},
    \label{eq:Cn_MC}
\end{align}
where the last inequality uses our assumption that $n\ge2\ell$.
Combining Eqs.~\eqref{eq:MC_KL} and~\eqref{eq:Cn_MC} then yields
\begin{align}\label{eq:MC_KL_intermediate}
    \KL(\Lcal_q^n\,\|\,\Lcal_p^n) \leq \frac{5n}{2\ell} \KL(\pi_q\,\|\,\pi_p).
\end{align}
Next, we describe a standard choice of $p > q$ that will completely specify our two instances.
We will pick
\[
    q=\frac12 \text{ and }
    p=\frac12+\delta, \text{ where }
    \delta=\sqrt{\frac{\ell}{8n}}.
\]
Since $n\ge2\ell$, we have $\delta\le1/4$, so $p\le3/4$. It is easy to verify that
\begin{align*}
    \KL(\pi_q\,\|\,\pi_p)
    &=
    \KL \big(
        \mathrm{Ber}(q)\,\big\|\,\mathrm{Ber}(p)
    \big)
    =
    -\frac12\log(1-4\delta^2)
    \le
    \frac83\delta^2,
\end{align*}
where the last inequality follows from the identity $-\log(1-x)\le x/(1-x)$ for $0\le x<1$.
Combining this with Eq.~\eqref{eq:MC_KL_intermediate} gives us
\[
    \KL(\Lcal_q^n\,\|\,\Lcal_p^n)
    \le
    \frac{5n}{2\ell}
    \cdot \frac83\delta^2
    =
    \frac56.
\]
Pinsker's inequality therefore yields
\(
    \TV(\Lcal_q^n,\Lcal_p^n)
    \le
    \sqrt{\frac5{12}}
    <
    \frac23.
\)

On the other hand, Eq.~\eqref{eq:MC_surprise_separation} gives
\(
    \Delta_S
    :=
    |\surprise(n;q)-\surprise(n;p)|
    \ge
    \frac{9\delta}{20}.
\)
Now, we will use the shorthand $\EE_u\left[\cdot\right]$ to denote expectation with respect to the distribution $\Lcal_u^n$ for $u \in \{q,p\}$.
Then, directly applying Le Cam's two-point method~\citep[Section~15.2]{wainwright2019high} yields
\begin{align*}
    \inf_{\widetilde\surprise}
    \max_{u\in\{q,p\}}
    \EE_u \left[
        \big(
            \widetilde\surprise(X^n)-\surprise(n;u)
        \big)^2
    \right]
    &\ge
    \frac{\Delta_S^2}{8}
    \big(
        1-\TV(\Lcal_q^n,\Lcal_p^n)
    \big)
    \ge
    \frac18
    \left(\frac{9\delta}{20}\right)^2
    \frac13\\
    &=
    \frac{27}{3200}\delta^2=
    \frac{27\ell}{25600n}
    \ge
    \frac{\ell}{1024n}.
\end{align*}
By Lemma~\ref{lemma:block_MC}, all the Markov chains from this family, including these two instances, mix in time $\Tmix \leq \ell$. Therefore, we have
\[
    \inf_{\widetilde\surprise}
    \sup_{X^{n+1}\in\Pcal_{\mathrm{MC}}(\Tmix)}
    \EE \left[
        \big(
            \widetilde\surprise(X^n)-\surprise
        \big)^2
    \right]
    \ge
    \inf_{\widetilde\surprise}
    \max_{u\in\{q,p\}}
    \EE_u \left[
        \big(
            \widetilde\surprise(X^n)-\surprise(n;u)
        \big)^2
    \right]
    \ge
    \frac{\Tmix}{1024n}.
\]
This is the desired statement, and completes the proof of the proposition.
\qed

\subsubsection{Proof of Lemma~\ref{lemma:rate_q}}\label{sec:pf_lemma_rate_q}

Differentiating Eq.~\eqref{eq:Siid} with respect to $q$ yields
\begin{align*}
    \frac{\partial}{\partial q} \Siid(m;q)
    &=
    (1-q)^{m-1}\bigl(1-(m+1)q\bigr)\\
    &\qquad
    -
    \left(1-\frac{1-q}{k}\right)^{m-1}
    \left(
        1-\frac{(m+1)(1-q)}{k}
    \right).
\end{align*}
The first term in the above display is non-positive because $q\ge1/2$ and $m\ge1$. For the second term, since
$1-q\le1/2$, $m\le n$, $k=16n$, and $\frac{1-q}{k} < 1$, we have
\[
    \left(1-\frac{1-q}{k}\right)^{m-1}
    \ge
    1-\frac{(m-1)(1-q)}{k}
    \ge
    \frac{31}{32}.
\]
Moreover, we only need to show the statement of the lemma for $n\ge 1$.
Therefore, we have
\[
    1-\frac{(m+1)(1-q)}{k}
    \ge
    1-\frac{n+1}{32n}
    \ge
    \frac{15}{16}.
\]
Putting these inequalities together gives us
\[
    \frac{\partial}{\partial q} \Siid(m;q)
    \le
    -\frac{31}{32}\frac{15}{16}
    \le
    -\frac9{10} \text{ for all } q.
\]
The lemma then follows by applying the mean-value theorem together with the above bound.
\qed

\subsubsection{Proof of Lemma~\ref{lemma:block_MC}}
\label{sec:pf_lemma_block_MC}

Recall that we defined the candidate stationary distribution as
\[
    \mu_q(y,r)
    =
    \pi_q(y)\alpha_r,
    \text{ where }
    \alpha_r
    =
    \frac{2(\ell-r+1)}{\ell(\ell+1)}.
\]
We first verify stationarity. Since $\sum_{r=1}^{\ell}\alpha_r
    =
    \frac{2}{\ell(\ell+1)}
    \sum_{r=1}^{\ell}(\ell-r+1)
    =
    1$, we have $\sum_{y \in \mathcal{Y}} \sum_{r=1}^{\ell} \mu_q(y,r) = \sum_{y \in \mathcal{Y}} \pi_q(y) = 1$.
    Consequently, $\mu_q$ is a probability distribution.
Now, let $P_q$ denote the transition kernel.
Overloading notation, we denote $\mu_q$ to be the vector whose elements comprise of $\{\mu_q(y,r)\}_{y \in \Ycal, r \in [\ell]}$.
To verify stationarity, it suffices to show that $\mu_q^\top P_q = \mu_q^\top$.
For any $1\le r<\ell$, the state
$(y,r)$ can be reached either deterministically from $(y,r+1)$ or through
a reset from a state with remaining length $1$. Therefore, a direct calculation gives us
\begin{align*}
    (\mu_q^\top P_q)(y,r)
    =
    \pi_q(y)\alpha_{r+1}
    +
    \sum_{y' \in \Ycal}\mu_q(y',1)\frac{\pi_q(y)}{\ell}&=
    \pi_q(y)\alpha_{r+1}
    +
    \alpha_1\frac{\pi_q(y)}{\ell}\\
    &=
    \pi_q(y)
    \left(
        \frac{2(\ell-r)}{\ell(\ell+1)}
        +
        \frac{2}{\ell(\ell+1)}
    \right)\\
    &=
    \pi_q(y)\alpha_r
    =
    \mu_q(y,r).
\end{align*}
For the special case $r=\ell$, the state $(y,\ell)$ can only be reached through a reset.
Again, direct calculation gives us
\[
    (\mu_q^\top P_q)(y,\ell)
    =\sum_{y' \in \Ycal}\mu_q(y',1)\frac{\pi_q(y)}{\ell}=
    \alpha_1\frac{\pi_q(y)}{\ell}
    =
    \frac{2\pi_q(y)}{\ell(\ell+1)}
    =
    \mu_q(y,\ell).
\]
This calculation covers all cases, and so we have shown that
\(
    \mu^{\top} P_q=\mu_q^{\top}.
\)

Next, we show irreducibility.
Note that $\pi_q$ has full support, i.e. $\pi_q(y) > 0$ for all $y \in \Ycal$. 
Moreover, from every state the chain reaches a reset in the number of steps it takes for the block counter to become $1$, which is at most equal to $\ell$.
Consequently, after at most $\ell$ steps, every state in
$\Ycal\times[\ell]$ can be reached with positive probability. 
This implies that the chain is irreducible.

Finally, it remains to compute the mixing time. 
In particular, we will show the desired statement that $\Tmix \leq \ell$.
Denote by $P_q^{\ell}(x)$ the conditional distribution of $X_{\ell + 1}$ given that $X_1 = x$.
Recalling the definition of mixing time (Eq.~\eqref{defn:mixing_time}), it suffices to show that $\TV(P_q^{\ell}(x),\mu_q) < 1/4$.

Define $\nu_q(y,r)
    :=
    \frac{\pi_q(y)}{\ell}$
to be the distribution of the tuple of an independently sampled block label and block length.
Suppose that we start from the state $x = (y,r)$.
Then, after exactly $r$ steps, the initial block
has ended, which yields $P_q^r(x)=\nu_q$.
Applying this together with the fact that $\mu_q$ is the stationary distribution, we have
\begin{align*}
    \TV(P_q^\ell(x),\mu_q)
    &=
    \TV \big(
        \nu_q^\top P_q^{\ell-r},
        \mu_q^\top P_q^{\ell-r}
    \big)\\
    &\overset{\1}{\le}
    \TV(\nu_q,\mu_q),
\end{align*}
where step $\1$ uses the contraction property of the total variation distance.
It remains to compute $\TV(\nu_q,\mu_q)$.
Direct calculation yields
\begin{align*}
    \TV(\nu_q,\mu_q)
    =
    \frac12
    \sum_{y\in\Ycal}\sum_{r=1}^{\ell}
    \left|
        \frac{\pi_q(y)}{\ell}
        -
        \pi_q(y)
        \frac{2(\ell-r+1)}{\ell(\ell+1)}
    \right| 
    &=
    \frac{1}{2\ell(\ell+1)}
    \sum_{r=1}^{\ell}|2r-\ell-1|\\
    &=
    \begin{cases}
        \displaystyle
        \frac{\ell}{4(\ell+1)},
        & \ell \text{ even},\\[2mm]
        \displaystyle
        \frac{\ell-1}{4\ell},
        & \ell \text{ odd},
    \end{cases}\\
    &<
    \frac14.
\end{align*}
Putting these steps together yields
\(
    \Tmix\le\ell.
\)
This completes the proof of the lemma.
\qed

\subsubsection{Proof of Lemma~\ref{lemma:probability_bound}}\label{sec:proof_probability_bound}

Recall that we need to bound $\Prob(N_{L_{n+1}}=0)$. First, let $\mathsf{BL}_1,\ldots,\mathsf{BL}_{\niid}$ denote the lengths of the consecutive blocks that intersect with the trajectory $X^n$; note that $\mathsf{BL}_1 = L_1$ and $\mathsf{BL}_j \overset{\text{i.i.d.}}{\sim} \UNIF([\ell])$ for $j \in \{2,\ldots,\niid-1\}$, where $\mathsf{BL}_{\niid-1}$ is last complete block length. Since we have assumed that $n \geq 2\ell$, we have
\begin{align*}
\Prob(N_{L_{n+1}} = 0) &\leq \Prob\big( L_{n+1} \text{ does not appear in the first two blocks} \big) \\
&=  \Prob\big( L_{n+1}> \mathsf{BL}_1, L_{n+1}> \mathsf{BL}_2 \big)=  \Prob\big( L_{n+1}> \max\{\mathsf{BL}_1, \mathsf{BL}_2\}\big) \\ 
&= \EE \left[ \Prob\big( L_{n+1}> \max\{\mathsf{BL}_1, \mathsf{BL}_2\} | \mathsf{BL}_1,\mathsf{BL}_2\big)\right].
\end{align*}
Note that in the above display, we are conditioning on the length of the first two blocks and $L_{n+1}$ represents the remaining length of the current block. Thus, for any $r \in [\ell]$ we have $\Prob\big( L_{n+1}> r \big) \leq \Prob( \mathsf{BL}_{2} > r) = \frac{\ell-r}{\ell}$, since $\mathsf{BL}_{2} \sim \UNIF([\ell])$. Thus, we obtain
\begin{align*}
    \Prob(N_{L_{n+1}} = 0) &\leq \EE \left[ \Prob\big( L_{n+1}> \max\{\mathsf{BL}_1, \mathsf{BL}_2\} | \mathsf{BL}_1,\mathsf{BL}_2\big)\right] \leq \EE \left[ \frac{\ell - \max\{\mathsf{BL}_1, \mathsf{BL}_2\}}{\ell}\right] \leq \EE \left[ \frac{\ell -  \mathsf{BL}_2}{\ell}\right] \leq \frac{1}{2}.
\end{align*}
Thus, we have obtained the desired statement that
\(
    \Prob(N_{L_{n+1}}\ge1)\ge\frac12.
\)

\subsection{Proof of Corollary~\ref{corollary:count_surprise_probability}}\label{sec:proof_corr_count_surprise_probability}

In this section we provide the proof of Corollary~\ref{corollary:count_surprise_probability}, which is Theorem~\ref{theorem:MSE_theorem} applied to the count surprise probability functional defined in Section~\ref{sec:count_surprise} as \(
    f(X_{n+1};\measure_n) := \ind{N_{X_{n+1}}(X^n) \leq \zeta}.
\)
Recall that we denoted by $\Ncal := \Prob(N_{X_{n+1}}(X^n) \leq \zeta)$ the estimand (which is the expected count surprise probability functional), and by $\Nhat(\tau)$ our count surprise probability estimator.
Being an indicator, the count surprise probability functional satisfies $\|f\|_{\infty} = 1$. We claim that, for this functional, Assumption~\ref{assmp:stability} holds with $r_{f}(\tau,n) = \frac{(\zeta+1)\tau}{n}$ and Assumption~\ref{assmp:bd} holds with $B_f = 2\zeta+5$. 
Therefore, we can invoke Theorem~\ref{theorem:MSE_theorem} to obtain
\begin{align*}
        \MSE\big(\Nhat(\tau),\Ncal\big) \leq 64 \cdot \Bigl( \beta(\tau)+ \frac{\tau}{n}\Bigr)^2 + 4 \cdot \Bigl( \frac{(\zeta+1)\tau}{n} \Bigr)^2 + \frac{2 \cdot (2\zeta+5)^2 \cdot \gamma^2 \cdot s}{n},
    \end{align*}
where we again used the inequality $(a+b)^2 \leq 2(a^2 +b^2)$, which yields $2\left(4\Bigl( \beta(\tau)+ \frac{\tau}{n}\Bigr)+r_f(\tau,n)\right)^2 \le 64\Bigl( \beta(\tau)+ \frac{\tau}{n}\Bigr)^2 + 4 r_f(\tau,n)^2$. This is exactly the bound claimed in Corollary~\ref{corollary:count_surprise_probability}. It remains to verify Assumptions~\ref{assmp:stability} and~\ref{assmp:bd}, which we do below.

\subsubsection{Verification of Assumption~\ref{assmp:stability}}

For the count surprise probability functional, we have
\[
\frac{1}{n}\sum_{i=1}^{n}
\left|
\EE \left[f(X_{n+1};\measure_n)\right]
-
\EE \left[f(X_{n+1};\measure_{\Iset_i})\right]
\right|
= \frac{1}{n}\sum_{i=1}^n\left|\Prob(N_{X_{n+1}}(X_{\Iset_i}) \leq \zeta) - \Prob(N_{X_{n+1}}(X^n) \leq \zeta) \right|,
\]
where the sets $\Dset_i$ and $\Iset_i$ are defined in Eq.~\eqref{eq:index-sets}.
From here, the proof will resemble the steps in~\cite[Section 7.4.1]{pananjady2024just}.
Recall that we are working with skipped estimators with offset $u \in \{0,\ldots,\tau-1\}$ as in the proof of Theorem~\ref{theorem:MSE_theorem}, and that we defined $n_0 = n/\tau$ as shorthand. The triangle inequality then gives us
\begin{align}\nonumber
&\frac{1}{n}\sum_{i=1}^n\left|\Prob(N_{X_{n+1}}(X_{\Iset_i}) \leq \zeta) - \Prob(N_{X_{n+1}}(X^n) \leq \zeta) \right|  \\ \label{eq:intermediate_step_n}
&\qquad\leq \frac{1}{\tau} \sum_{u=0}^{\tau-1} \frac{1}{n_0} \sum_{i=1}^{n_0} \left|\Prob(N_{X_{n+1}}(X_{\Iset_{\tau i -u}}) \leq \zeta) - \Prob(N_{X_{n+1}}(X^n) \leq \zeta) \right|.
\end{align}
Using \citet[Lemma 12]{pananjady2024just}, for any $i \in [n]$ we have
\begin{align*}
    \left|\Prob(N_{X_{n+1}}(X_{\Iset_{i}}) \leq \zeta) - \Prob(N_{X_{n+1}}(X^n) \leq \zeta) \right| \leq \EE\left[ \ind{N_{X_{n+1}}(X_{\Iset_{i}}) \leq \zeta}\ind{X_{n+1} \in \bX_{\Dset_i}} \right].
\end{align*}
Substituting the above display into Eq.~\eqref{eq:intermediate_step_n}, we obtain
\begin{align}\label{eq:step2-count}
&\frac{1}{n}\sum_{i=1}^n\left|\Prob(N_{X_{n+1}}(X_{\Iset_i}) \leq \zeta) - \Prob(N_{X_{n+1}}(X^n) \leq \zeta) \right| \nonumber \\ 
&\qquad \leq \frac{1}{\tau} \sum_{u=0}^{\tau-1} \frac{1}{n_0} \sum_{i=1}^{n_0} \EE\left[ \ind{N_{X_{n+1}}(X_{\Iset_{\tau i - u}}) \leq \zeta}\ind{X_{n+1} \in \bX_{\Dset_{\tau i - u}}} \right].
\end{align}
Now suppose for some $i \in [n_0]$ we have $\ind{N_{X_{n+1}}(X_{\Iset_{\tau i -u}}) \leq \zeta}\ind{X_{n+1} \in \bX_{\Dset_{\tau i - u}}} =1$. (If no such index exists, we are done.) This means that $N_{X_{n+1}}(X_{\Iset_{\tau i - u}}) \leq \zeta$. Thus, we have
\begin{align*}
    \sum_{i' \in [n_0] \setminus i} \ind{N_{X_{n+1}}(X_{\Iset_{\tau i' -u}}) \leq \zeta}\ind{X_{n+1} \in \bX_{\Dset_{\tau i' - u}}} &\leq \sum_{i' \in [n_0] \setminus i} \ind{X_{n+1} \in \bX_{\Dset_{\tau i' - u}}} \\\\
    &\overset{\1}{\leq} N_{X_{n+1}}(X_{\Iset_{\tau i -u}}) \leq \zeta,
\end{align*}
where step $\1$ follows from Eq.~\eqref{eq:subset}.
Hence, $\sum_{i=1}^{n_0} \EE\left[ \ind{N_{X_{n+1}}(X_{\Iset_{\tau i - u}}) \leq \zeta}\ind{X_{n+1} \in \bX_{\Dset_{\tau i - u}}} \right] \leq \zeta +1$ for each offset $u \in \{0,\ldots,\tau-1\}$.
Substituting this inequality into Ineq.~\eqref{eq:step2-count} yields
\begin{align*}
    \frac{1}{n}\sum_{i=1}^n\left|\Prob(N_{X_{n+1}}(X_{\Iset_i}) \leq \zeta) - \Prob(N_{X_{n+1}}(X^n) \leq \zeta) \right| \leq \frac{1}{\tau}\sum_{u=0}^{\tau-1} \frac{(\zeta+1)\tau}{n} = \frac{(\zeta+1)\tau}{n}.
\end{align*}
Thus, we have shown that Assumption~\ref{assmp:stability} is satisfied with $r_{f}(\tau,n) = \frac{(\zeta+1)\tau}{n}$.

\subsubsection{Verification of Assumption~\ref{assmp:bd}}

Recall, $X^n = (X_1, \ldots, X_n)$ is the original sequence and $X^{(k)}(a) := (X_1,\ldots,X_{k-1},a,X_{k+1},\ldots,X_n)$ is the sequence obtained after replacing the $k$-th symbol $X_k$ with the symbol $a \in \Xcal$. 
We overload notation and use $\Nhat(X^n), \Nhat(X^{(k)}(a)),\Nhat(X^{(k)}(b))$ to refer to the surprise probability estimator when deployed on the original sequence $X^n$ and the modified sequences $X^{(k)}(a)$ and $X^{(k)}(b)$ respectively (note that we have dropped the window parameter $\tau$ for clarity).
At times, we drop the sequence argument and simply refer to $\Nhat$ or $\Nhatlocalj$ when the sequence on which the estimator is applied is clear from context.
Then, we have the following bounded differences-type lemma for our estimator of the count surprise probability functional.
\begin{lemma}\label{lemma:bdd_lemma_surprise_n}
    For any sequence $X^n \in \Xcal^n$ and any window size $\tau \in \{1,\ldots, n-1\}$ we have
    \begin{align*}
        \sup_{a,b \in \Xcal}\sup_{k \in [n]}\left|\Nhat(X^{(k)}(a))-\Nhat(X^{(k)}(b)) \right| \leq \frac{(2\zeta+5)}{n}.
    \end{align*}
\end{lemma}
Lemma~\ref{lemma:bdd_lemma_surprise_n} directly implies that the count surprise probability estimator $\Nhat(\tau)$ satisfies Assumption~\ref{assmp:bd} with $B_f = 2\zeta+5$. We now prove this lemma.

\paragraph{Proof of Lemma~\ref{lemma:bdd_lemma_surprise_n}.}
    Fix the sequence $X^n$, pair of symbols $a,b \in \Xcal$ and index $k \in [n]$. Recall that we defined $\Nhatlocalj(X^n) = \ind{N_{X_j}(X_{\Iset_{j}}) \leq \zeta}$, where $N_x(\cdot)$
    denotes the number of occurrences of $x$ in the sequence argument. In parallel to the
    proof of Lemma~\ref{lemma:bdd_lemma_surprise}, we define
    \begin{align*}
        V^{(k)}(a,b) = \sum_{j=1}^n \ind{\Nhatlocalj(X^{(k)}(a)) \neq \Nhatlocalj(X^{(k)}(b))}.
    \end{align*}
    An identical series of steps as in the proof of Lemma~\ref{lemma:bdd_lemma_surprise} yields
    \[
    \bigl|\Nhat(X^{(k)}(a))-\Nhat(X^{(k)}(b))\bigr| \le \frac{V^{(k)}(a,b)}{n}.
    \]
    Thus, it suffices to prove that $V^{(k)}(a,b) \le 2\zeta+5$ for any pair of symbols $a,b \in \Xcal$ and any index $k \in [n]$. We will prove this via a looser\footnote{Observe that specializing this bound to $\zeta = 0$ yields $V^{(k)}(a,b) \leq 5$ for the case of the surprise probability functional, which is looser than the claimed bound $V^{(k)}(a,b) \leq 3$ in the proof of Lemma~\ref{lemma:bdd_lemma_surprise}.}, but simpler and more general argument as compared to the one presented in the proof of Lemma~\ref{lemma:bdd_lemma_surprise}.

    For an index $j \in [n]$, the local estimator $\Nhatlocalj$ can change only if switching the symbol at index
    $k$ from $a$ to $b$ alters the value of $\ind{N_{X_j}(X_{\Iset_j}) \le \zeta}$.
    In the worst case, the indicator corresponding to the index $j = k$ could change.
    For $j \neq k$, the count $N_{X_j}(X_{\Iset_j})$ changes, and therefore the corresponding local estimator can change only when $k \in \Iset_j$ and $X_j \in \{a,b\}$. Writing $A$ (respectively, $B$)
    for the number of changed indicators at positions $j \neq k$ with $X_j = a$
    (respectively, $X_j = b$), we therefore have
    \begin{align}\label{eq:V-A-B-bound}
    V^{(k)}(a,b) \le A + B + 1.
    \end{align}
    We first show that $A \le \zeta+2$. Consider an index $j \neq k$ such that $X_j = a$ and $k \in \Iset_j$. Switching $X_k$
    from $a$ to $b$ removes exactly one occurrence of $a$ from $\Iset_j$, lowering
    $N_a(X_{\Iset_j})$ by exactly one.
    Hence, the local estimator $\Nhatlocalj$ changes (from $0$ to $1$) only if this count
    dips below the threshold $\zeta$, which can only happen if $N_a\bigl(X_{\Iset_j}\bigr) = \zeta+1$ (here and in all uses below, $X_{\Iset_j}$ is used as shorthand to denote the original decorrelated sequence corresponding to index $j$).
    
    We will show that the number of such indices for which this can happen is at most $\zeta + 1$.
    Let $M$ be the total number of occurrences of $a$ in the original sequence $X^{(k)}(a)$.
    If $M \leq \zeta + 1$, we are done.
    If $M > \zeta + 1$, let $j_1 < \cdots < j_M$ be the occurrences of $a$ in $X^{(k)}(a)$. For $j = j_i \neq k$, we have $\Iset_{j_i} = [n]\setminus[j_i,j_i+\tau-1]$. This yields
    \[
    N_a\bigl(X_{\Iset_{j_i}}\bigr) \geq i - 1,
    \]
    meaning that $N_a\bigl(X_{\Iset_{j_i}}\bigr) = \zeta + 1$ necessitates $i - 1 \le \zeta + 1 \implies i \leq \zeta + 2$.
    Essentially, only the local estimators corresponding to the first $\zeta+2$ occurrences
    $j_1,\ldots,j_{\zeta+2}$ of $a$ can change, giving $A \le \zeta+2$.
    A similar argument applied to the symbol $b$ (and only considering indices for which the count $N_b\bigl(X_{\Iset_j}\bigr) = \zeta$) yields $B \le \zeta + 2$. Substituting the bounds on $A$ and $B$ into Eq.~\eqref{eq:V-A-B-bound} yields the desired statement $V^{(k)}(a,b) \leq 2 \zeta + 5$ and completes the proof of the lemma.
\qed

\subsection{Proof of Corollary~\ref{corollary:nearest_surprise}}\label{sec:proof_corr_nearest_surprise}
In this section we provide the proof of Corollary~\ref{corollary:nearest_surprise} for the nearest-neighbor tail probability functional defined in Section~\ref{sec:nns} as
\(
f(X_{n+1};\measure_n)
    =
    \ind{\min_{i\in[n]} |X_{n+1}-X_i|>\delta}.
\)
Recall that we denoted by $\NS(\delta) := \Prob(\min_{i\in[n]} |X_{n+1}-X_i|>\delta)$ the estimand (which is the expected nearest-neighbor tail probability functional), and by $\NShat(\tau)$ our nearest-neighbor tail probability estimator.
Being an indicator, the nearest-neighbor tail probability functional satisfies $\|f\|_{\infty} = 1$. We claim that, for this functional, Assumption~\ref{assmp:stability} holds with $r_{f}(\tau,n) = \frac{\tau}{n}$ and Assumption~\ref{assmp:bd} holds with $B_f = 5$. Therefore, we can invoke Theorem~\ref{theorem:MSE_theorem} to obtain
\[
\MSE \big(\NShat(\tau), \NS(\delta)\big)
\le
64 \cdot \Bigl( \beta(\tau)+ \frac{\tau}{n}\Bigr)^2
+
4 \cdot \big(\frac{\tau}{n}\big)^2
+
\frac{50 \cdot \gamma^2 \cdot s}{n},
\]
where we again used the inequality $(a+b)^2 \leq 2(a^2 +b^2)$, which yields $2\left(4\Bigl( \beta(\tau)+ \frac{\tau}{n}\Bigr)+r_f(\tau,n)\right)^2 \le 64\Bigl( \beta(\tau)+ \frac{\tau}{n}\Bigr)^2 + 4 r_f(\tau,n)^2$.
This is exactly the bound claimed in Corollary~\ref{corollary:nearest_surprise}. 
It remains to verify Assumptions~\ref{assmp:stability} and~\ref{assmp:bd}, which we do below.

\subsubsection{Verification of Assumption~\ref{assmp:stability}}

We will follow the same argument that we used to verify Assumption~\ref{assmp:stability} for the original surprise probability functional with minor adaptations to the nearest-neighbor tail probability functional.
For the nearest-neighbor tail probability functional, we have
\begin{align*}
&\frac{1}{n}\sum_{i=1}^{n}
\left|
\EE \left[f(X_{n+1};\measure_n)\right]
-
\EE \left[f(X_{n+1};\measure_{\Iset_i})\right]
\right|\\
&\qquad = \frac{1}{n}\sum_{i=1}^n\left|\Prob\big(\min_{j\in \Iset_i} |X_{n+1}-X_j|>\delta\big) - \Prob\big(\min_{j\in[n]} |X_{n+1}-X_j|>\delta\big) \right|,
\end{align*}
where the sets $\Dset_i$ and $\Iset_i$ are defined in Eq.~\eqref{eq:index-sets}.
As before, we work with skipped estimators with offset $u \in \{0,\ldots,\tau-1\}$.
Recall that we defined $n_0 = n/\tau$ as shorthand.
Then, the skipped estimator with offset $u$ (defined in Eq.~\eqref{eq:general_skipped_estimators}), specialized to the nearest-neighbor tail probability, is given by $\NShat(\tau;u) \defn \frac{1}{n_0}\sum_{i=1}^{n_0}\NShat_{(\tau i-u)}(\tau)$.
The triangle inequality then gives us
\begin{align}\nonumber
&\frac{1}{n}\sum_{i=1}^n\left|\Prob\big(\min_{j\in \Iset_i} |X_{n+1}-X_j|>\delta\big) - \Prob\big(\min_{j\in[n]} |X_{n+1}-X_j|>\delta\big) \right| \\ \nonumber &\qquad\leq \frac{1}{\tau} \sum_{u=0}^{\tau-1} \frac{1}{n_0} \sum_{i=1}^{n_0} \left|\Prob\big(\min_{j\in \Iset_{\tau i -u}} |X_{n+1}-X_j|>\delta\big) - \Prob\big(\min_{j\in[n]} |X_{n+1}-X_j|>\delta\big) \right| \\ \nonumber
&\qquad = \frac{1}{\tau} \sum_{u=0}^{\tau-1} \frac{1}{n_0} \sum_{i=1}^{n_0} \left|\EE\left[\ind{\min_{j\in \Iset_{\tau i -u}} |X_{n+1}-X_j|>\delta} - \ind{\min_{j\in[n]} |X_{n+1}-X_j|>\delta} \right]\right|\\ \label{eq:intermediate_step_ns}
&\qquad \overset{\1}{=} \frac{1}{\tau} \sum_{u=0}^{\tau-1} \frac{1}{n_0} \sum_{i=1}^{n_0} \left|\EE\left[\ind{\min_{j\in \Iset_{\tau i -u}} |X_{n+1}-X_j|>\delta} \ind{\min_{j\in \Dset_{\tau i -u}} |X_{n+1}-X_j|\le\delta} \right]\right|,
\end{align}
where step $\1$ follows because the two indicators can differ only if \(X_{n+1}\) has no neighbor within
distance \(\delta\) among $\{X_j : j \in \Iset_{\tau i -u}\}$, but does have such a neighbor among
$\{X_j : j \in \Dset_{\tau i -u}\}$. (Note that step $\1$ is an analog of~\cite[Lemma 11]{pananjady2024just} adapted to these nearest-neighbor-based indicator functions.)

Recall from Eq.~\eqref{eq:subset} that $\bigcup_{i' \in [n_0] \setminus i} \Dset_{\tau i'-u} \subset \Iset_{\tau i - u}$.
Now, suppose that for some \(i \in [n_0]\), we have
\(
\ind{\min_{j \in \Iset_{\tau i-u}} |X_{n+1}-X_j|>\delta}
\ind{\min_{j \in \Dset_{\tau i -u}} |X_{n+1}-X_j|\le\delta}
=1.
\)
This means that no element in the subsequence \( X_{\Iset_{\tau i -u}}\) can be within distance $\delta$ of \(X_{n+1}\). Thus, we have
\begin{align*}
&\sum_{i'\in[n_0]\setminus i}
\ind{\min_{j \in \Iset_{\tau i'-u}} |X_{n+1}-X_j|>\delta}
\ind{\min_{j \in \Dset_{\tau i' -u}} |X_{n+1}-X_j|\le\delta} \\
&\qquad\le
\sum_{i'\in[n_0]\setminus i}
\ind{\min_{j \in \Dset_{\tau i' -u}} |X_{n+1}-X_j|\le\delta} \\
&\qquad\overset{\1}{\le}
\ind{\min_{j \in \Iset_{\tau i -u}} |X_{n+1}-X_j|\le\delta}
=0,
\end{align*}
where step \(\1\) follows from Eq.~\eqref{eq:subset}.
Hence, for each offset $u \in \{0,\ldots,\tau-1\}$, we have \newline $\sum_{i=1}^{n_0} \left|\EE\left[\ind{\min_{j\in \Iset_{\tau i -u}} |X_{n+1}-X_j|>\delta} \ind{\min_{j\in \Dset_{\tau i -u}} |X_{n+1}-X_j|\le\delta} \right]\right| \leq 1$.
Substituting this inequality into Ineq.~\eqref{eq:intermediate_step_ns} yields
\begin{align*}
&\frac{1}{n}\sum_{i=1}^n\left|\Prob\big(\min_{j\in \Iset_i} |X_{n+1}-X_j|>\delta\big) - \Prob\big(\min_{j\in[n]} |X_{n+1}-X_j|>\delta\big) \right| \leq \frac{1}{\tau} \sum_{u=0}^{\tau-1} \frac{1}{n_0} = \frac{1}{n_0} = \frac{\tau}{n}.
\end{align*}
Thus, we have shown that Assumption~\ref{assmp:stability} is satisfied with $r_{f}(\tau,n) = \frac{\tau}{n}$.

\subsubsection{Verification of Assumption~\ref{assmp:bd}}

Recall that $X^n = (X_1, \ldots, X_n)$ is the original sequence and $X^{(k)}(a) := (X_1,\ldots,X_{k-1},a,X_{k+1},\ldots,X_n)$ is the sequence obtained after replacing the $k$-th symbol $X_k$ with the symbol $a \in \Xcal$. 
We overload notation and use $\NShat(X^n), \NShat(X^{(k)}(a)),\NShat(X^{(k)}(b))$ to refer to the nearest-neighbor tail probability estimator when deployed on the original sequence $X^n$ and the modified sequences $X^{(k)}(a)$ and $X^{(k)}(b)$ respectively (note that we have dropped the window parameter $\tau$ for clarity).
Then, we have the following bounded differences-type lemma, which verifies Assumption~\ref{assmp:bd} for $\NShat(\tau)$.

\begin{lemma}
\label{lemma:bdd_lemma_ns}
Suppose \(\Xcal \subset \mathbb{R}\). For any sequence \(X^n \in \Xcal^n\) and any window size
\(\tau \in \{1,\ldots,n-1\}\), we have
\[
\sup_{a,b\in\Xcal}\sup_{k\in[n]}
\left|
\NShat(X^{(k)}(a))-\NShat(X^{(k)}(b))
\right|
\le \frac{5}{n}.
\]
\end{lemma}
Lemma~\ref{lemma:bdd_lemma_ns} directly implies that the nearest-neighbor tail probability estimator $\NShat(\tau)$ satisfies Assumption~\ref{assmp:bd} with $B_f = 5$. We now prove this lemma.

\paragraph{Proof of Lemma~\ref{lemma:bdd_lemma_ns}.}
Let us define the following quantity for some fixed sequence $X^n$, pair of symbols $a,b \in \Xcal$ and index $k \in [n]$:
\[
V^{(k)}(a,b) := \sum_{j=1}^n \ind{\NShat_{(j)}(X^{(k)}(a)) \neq \NShat_{(j)}(X^{(k)}(b))}.
\]
An identical series of steps as in the proof of Lemma~\ref{lemma:bdd_lemma_surprise} yields $\bigl|\NShat(X^{(k)}(a))-\NShat(X^{(k)}(b))\bigr| \le V^{(k)}(a,b)/n$. Thus, it suffices to prove that $V^{(k)}(a,b)\le 5$ for any pair of symbols $a, b \in \Xcal$ and any index $k \in [n]$.
We first consider all indices $j \neq k$, i.e. indices that remain unchanged between the original and modified sequence.
Here, the local estimator $\NShat_{(j)}$ can change only when $k\in\Iset_j$.
In this case, denoting as shorthand
\[
M_j := \min_{i\in\Iset_j\setminus\{k\}} |X_j - X_i|,
\]
we can write
\[
\NShat_{(j)}\bigl(X^{(k)}(c)\bigr) = \ind{M_j > \delta \ \text{ and }\ |X_j - c| > \delta},
\text{ for } c\in\{a,b\}.
\]
Thus, the local estimator $\NShat_{(j)}$ changes \emph{iff} $M_j > \delta$ and exactly one of $a,b$ lies within distance $\delta$ of $X_j$; i.e. either $\{|X_j - a| \leq \delta \text{ and } |X_j - b| > \delta\}$ or $\{|X_j - a| > \delta \text{ and } |X_j - b| \leq \delta \}$. Accordingly, define the (possibly larger) sets of \emph{type-$a$} and \emph{type-$b$} indices as below:
\begin{align*}
\Tcal_a &:= \bigl\{\, j \neq k \ :\ k\in\Iset_j,\ \ M_j > \delta,\ \ \text{ and } |X_j - a| \le \delta \,\bigr\}\\
\Tcal_b &:= \bigl\{\, j \neq k \ :\ k\in\Iset_j,\ \ M_j > \delta,\ \ \text{ and } |X_j - b| \le \delta \,\bigr\}.
\end{align*}
Clearly, every changed index $j\neq k$ must lie in $\Tcal_a$ or $\Tcal_b$ or both.
Including the index $j = k$, we then have
\begin{align}\label{eq:nns_V_split}
V^{(k)}(a,b) \;\le\; 1 + |\Tcal_a| + |\Tcal_b|.
\end{align}
We will now show that $|\Tcal_a| \le 2$; the bound $|\Tcal_b| \le 2$ can similarly be shown by an identical argument with $b$ in place of $a$.
To show this, we first claim that any pair of distinct indices $j \neq j' \in \Tcal_a$ must be far apart, meaning that $|X_j - X_{j'}| > \delta$.
Without loss of generality, let $j < j'$. Since $\Dset_{j'} = \{j',\ldots,(j'+\tau-1)\wedge n\}$, and $j < j'$, we get $j \in \Iset_{j'}$. Since we are considering indices $j \neq k$, we further have $j \in \Iset_{j'}\setminus\{k\}$.
Therefore, we obtain
\[
|X_{j'} - X_j| \;\ge\; \min_{i\in\Iset_{j'}\setminus\{k\}} |X_{j'} - X_i| \;=\; M_{j'} \;>\; \delta,
\]
where the last inequality uses the fact that $j' \in \Tcal_a$.
This proves the claim.

Consequently, the points $\{X_j : j \in \Tcal_a\}$ have pairwise distance greater than $\delta$, i.e. $|X_j - X_{j'}| > \delta$ for all $j \neq j' \in \Tcal_a$. 
On the other hand, the definition of $\Tcal_a$ implies that $X_j \in [a - \delta, a + \delta]$ for all $j \in \Tcal_a$, meaning that the points all lie in an interval of length $2\delta$.
Putting these statements together yields $|\Tcal_a| \le 2$, and likewise $|\Tcal_b| \le 2$. Substituting these bounds into Equation~\eqref{eq:nns_V_split}, we obtain $V^{(k)}(a,b) \le 5$, and so we have completed the proof of the lemma.
\qed

\subsection{Proof of Corollary~\ref{corollary:test_error}}\label{sec:proof_corr_test_error}

In this section, we provide the proof of Corollary~\ref{corollary:test_error}, which is Theorem~\ref{theorem:MSE_theorem} applied to the classification test error functional defined in Section~\ref{sec:cte} as $f(Z_{n+1}; \measure_n^Z) = \ell(\widehat g_{Z^n}, Z_{n+1})$.
Recall that we denoted by $\testError$ the estimand (which is the population test error) and by $\testErrorhat(\tau)$ our test error estimator.
Because we have assumed a bounded loss function (i.e.~$\ell(\widehat g_{Z^n}, z) \in [0,1]$ for all $z \in \Zcal$), the test error functional satisfies $\|f\|_{\infty} = 1$. We claim, for this functional, that Assumption~\ref{assmp:stability} is satisfied with $r_{f}(\tau,n) = \frac{C\tau}{n-\tau+1}$, and Assumption~\ref{assmp:bd} is satisfied with $B_f=4C+1$ provided that $\tau \leq n/2$. Therefore, we can invoke Theorem~\ref{theorem:MSE_theorem} to obtain
\begin{align*}
    \MSE(\testErrorhat(\tau),\testError) \le 64\left( \beta(\tau)+ \frac{\tau}{n}\right)^2+4\left(\tfrac{C\tau}{n-\tau+1}\right)^2+\tfrac{2(4C+1)^2\gamma^2 s}{n},
    \end{align*}
where we used the inequality $(a+b)^2 \leq 2(a^2 +b^2)$, which yields $2\left(4\Bigl( \beta(\tau)+ \frac{\tau}{n}\Bigr)+r_f(\tau,n)\right)^2 \le 64\Bigl( \beta(\tau)+ \frac{\tau}{n}\Bigr)^2 + 4 r_f(\tau,n)^2$.
This is exactly the bound claimed in Corollary~\ref{corollary:test_error}.
It remains to verify Assumptions~\ref{assmp:stability} and~\ref{assmp:bd}, which we do below.

\subsubsection{Verification of Assumption~\ref{assmp:stability}}

For the classification test error functional, we have
\[
\frac{1}{n}\sum_{i=1}^{n}
\left|
\EE \left[f(Z_{n+1};\measure_n^Z)\right]
-
\EE \left[f(Z_{n+1};\measure_{\Iset_i}^Z)\right]
\right|
=
\frac{1}{n}\sum_{i=1}^{n}
\left|
\EE \left[\ell(\widehat g_{Z^n},Z_{n+1})\right]
-
\EE \left[\ell(\widehat g_{Z_{\Iset_i}},Z_{n+1})\right]
\right|,
\]
where the sets $\Dset_i$ and $\Iset_i$ are defined in Eq.~\eqref{eq:index-sets}.
Because $|\Dset_i| \leq \tau$, the training sets $Z^n$ and $Z_{\Iset_i}$ differ by at most $\tau$ data points. 
Therefore, we can use a telescoping sum along with the triangle inequality to obtain
\begin{align}
    &\left|
\EE \left[\ell(\widehat g_{Z^n},Z_{n+1})\right]
-
\EE \left[\ell(\widehat g_{Z_{\Iset_i}},Z_{n+1})\right]
\right| \leq \left|\EE \left[\ell(\widehat g_{Z^n},Z_{n+1})\right]
-
\EE \left[\ell(\widehat g_{Z_{[n]\setminus \{i\}}},Z_{n+1})\right]\right|\nonumber\\
&\qquad + \sum_{j = 0}^{\tau-2} \left|\EE \left[\ell(\widehat g_{Z_{[n] \setminus \{i,\ldots,i + j\}}},Z_{n+1})\right]
-
\EE \left[\ell(\widehat g_{Z_{[n] \setminus \{i,\ldots,i + j + 1\}}},Z_{n+1})\right]\right| \cdot \ind{i + j + 1 \leq n}\label{eq:intermediate_step_cte}
\end{align}
Now, we invoke the assumption of uniform stability that we have made on the classifier (Assumption~\ref{assmp:uniform_stability}).
In particular, for each $j \in \{0,\ldots,\tau-2\}$ that satisfies $i + j + 1 \leq n$, we invoke Assumption~\ref{assmp:uniform_stability} with respect to the dataset $Z_{[n] \setminus \{i,\ldots,i+j\}}$ to obtain
\[
\left|\EE \left[\ell(\widehat g_{Z_{[n] \setminus \{i,\ldots,i + j\}}},Z_{n+1})\right]
-
\EE \left[\ell(\widehat g_{Z_{[n] \setminus \{i,\ldots,i + j + 1\}}},Z_{n+1})\right]\right| \leq \frac{C}{|Z_{[n] \setminus \{i,\ldots,i + j\}}|} = \frac{C}{n - j - 1}.
\]
Similarly, invoking Assumption~\ref{assmp:uniform_stability} with respect to the original dataset $Z^n$ yields
\[
\left|\EE \left[\ell(\widehat g_{Z^n},Z_{n+1})\right]
-
\EE \left[\ell(\widehat g_{Z_{[n]\setminus \{i\}}},Z_{n+1})\right]\right| \leq \frac{C}{n}.
\]
Substituting these inequalities into Ineq.~\eqref{eq:intermediate_step_cte} gives us, for all $i \in [n]$,
\begin{align*}
\left|
\EE \left[\ell(\widehat g_{Z^n},Z_{n+1})\right]
-
\EE \left[\ell(\widehat g_{Z_{\Iset_i}},Z_{n+1})\right]
\right| \leq \sum_{k = 0}^{\tau - 1} \frac{C}{n - k} \leq \frac{C\tau}{n - \tau + 1}.
\end{align*}
Ultimately, we have 
\[
\frac{1}{n}\sum_{i=1}^{n}
\left|
\EE \left[f(Z_{n+1};\measure_n^Z)\right]
-
\EE \left[f(Z_{n+1};\measure_{\Iset_i}^Z)\right]
\right|
=
\frac{1}{n}\sum_{i=1}^{n}
\left|
\EE \left[\ell(\widehat g_{Z^n},Z_{n+1})\right]
-
\EE \left[\ell(\widehat g_{Z_{\Iset_i}},Z_{n+1})\right]
\right| \leq \frac{C \tau}{n - \tau + 1}.
\]
Thus, we have shown that Assumption~\ref{assmp:stability} is satisfied with $r_f(\tau,n) = \frac{C \tau}{n - \tau + 1}$.

\subsubsection{Verification of Assumption~\ref{assmp:bd}}
Recall that $Z^n$ denotes the original trajectory and $Z_{[n]\setminus k}$ represents the trajectory after removing the $k$-th data point. Following the replacement-vector notation of Assumption~\ref{assmp:bd}, for $k \in [n]$ and $z,\tilde{z} \in \Zcal$ we write
\[
Z^{(k)}(z) := (Z_1,\ldots,Z_{k-1},z,Z_{k+1},\ldots,Z_n),
\]
and we abbreviate $Z^n := Z^{(k)}(z)$ and $\widetilde{Z}^{n} := Z^{(k)}(\tilde{z})$, so that the two trajectories differ only at index $k$. Recalling the definition of the test error estimator (Eq.~\eqref{eq:test-error-estimator}) and using the triangle inequality, we obtain
\begin{align}\label{eq:te_bd_start}
    |\testErrorhat(Z^n) - \testErrorhat(\widetilde{Z}^{n})| &\leq \frac{1}{n} \sum_{j=1}^n |\ell(\widehat g_{Z_{\Iset_j}}, Z_j)-\ell(\widehat g_{\widetilde{Z}_{\Iset_j}}, \widetilde{Z}_j)|.
\end{align}
We use two elementary properties of the index sets of Eq.~\eqref{eq:index-sets}. First, since $j \in \Dset_j$, the $j$-th local estimator never trains on the $j$-th data point, i.e.
\begin{align}
    \label{eq:te_j_notin_Ij}
    j \notin \Iset_j .
\end{align}
Second, since $|\Dset_j| \le \tau$, every local classifier is trained on at least $n-\tau$ points, i.e.
\begin{align}
    \label{eq:te_Iset_size}
    |\Iset_j| = n - |\Dset_j| \;\geq\; n-\tau
    \qquad \text{ for every } j \in [n].
\end{align}
We now bound the $j$-th summand of Eq.~\eqref{eq:te_bd_start} by separating indices into two cases.
\begin{enumerate}
    \item \textbf{Case 1: $j = k$.} Since $\ell \in [0,1]$, we have $|\ell(\widehat g_{Z_{\Iset_k}}, Z_k)-\ell(\widehat g_{\widetilde{Z}_{\Iset_k}}, \widetilde{Z}_k)| \leq 1$.
    \item \textbf{Case 2: $j \neq k$.} Here the evaluation point is unchanged, i.e. $Z_j = \widetilde{Z}_j$, and only the classifier can change. This happens only if $k \in \Iset_j$, in which case the training sets $Z_{\Iset_j}$ and $\widetilde{Z}_{\Iset_j}$ differ only at data point $k$. The triangle inequality then gives us
    \begin{align*}
        |\ell(\widehat g_{Z_{\Iset_j}}, Z_j)-\ell(\widehat g_{\widetilde{Z}_{\Iset_j}}, Z_j)| &\leq |\ell(\widehat g_{Z_{\Iset_j}}, Z_j)-\ell(\widehat g_{Z_{\Iset_j\setminus k}}, Z_j)| + |\ell(\widehat g_{Z_{\Iset_j\setminus k}}, Z_j)-\ell(\widehat g_{\widetilde{Z}_{\Iset_j}}, Z_j)| \\
        &\overset{\1}{\leq} \frac{2C}{|\Iset_j|} \overset{\2}{\leq} \frac{2C}{n - \tau},
    \end{align*}
    where step $\1$ applied the uniform stability assumption (Assumption~\ref{assmp:uniform_stability}) twice and step $\2$ used Eq.~\eqref{eq:te_Iset_size} to lower bound the size of the training set $|\Iset_j|$.
\end{enumerate}
Putting these cases together yields
\begin{align*}
    |\testErrorhat(Z^n) - \testErrorhat(\widetilde{Z}^n)|
    \;\leq\; \frac{1}{n}\left( (n-1) \cdot \frac{2C}{n-\tau} + 1 \right)
    \;\overset{\1}{\leq}\; \frac{4C+1}{n},
\end{align*}
where step $\1$ uses the standing condition $\tau \leq n/2$, which gives $\frac{n-1}{n-\tau} \leq \frac{n}{n-\tau} \leq 2$.
Thus, Assumption~\ref{assmp:bd} is satisfied with $B_f=4C+1$.
\qed

\section{Failure of Good--Turing}\label{example:1}
We give a simple dependent-process example showing that the Good--Turing estimator need
not consistently estimate the surprise probability. This is the repeated block chain from Section~\ref{sec:numerical_results}. Fix $\ell\ge3$ and
consider the repeated-block process,
where the i.i.d.\ base sequence $(\widetilde X_i)_{i\ge1}$ is drawn uniformly
from a finite state space $\Xcal$. For each of these labels we draw $L_i$ which denotes the number of times $\widetilde X_i$ is repeated until the total trajectory length is $n$. $L_1$ is drawn from a law to ensure stationarity and the remaining block lengths
$(L_i)_{i\ge2}$ are drawn independently as
\(
    L_i\sim\UNIF(\{2,\ldots,\ell\}).
\)
The block lengths are independent of the base sequence, and the process is
initialized at stationarity. We assume that $|\Xcal|$ is much larger than the
number of samples, i.e., $|\Xcal|\gg n$ and $n \geq 2\ell$. Now $X_{n+1}$ is unseen among
$X_1,\dots,X_n$ only when $n$ is the last position of its current block and
the next block draws a symbol not previously observed. The probability that $n$ is the last position of its current block
is $1/\EE[L_i]$ for $i \ge 2$, so the surprise probability satisfies
\[
    \surprise
    \defn
    \Prob(X_{n+1}\notin\bX_{[n]})
    \geq
    \frac{2}{\ell+2}
    \left(
        \frac{|\Xcal|-n}{|\Xcal|}
    \right)
    =
    \frac{2}{\ell+2}
    -
    \frac{2n}{(\ell+2)|\Xcal|}.
\]
On the other hand, every complete block has length at least $2$, so a singleton
can arise only from the first block or the block truncated by the endpoint of the
observed trajectory. Hence the singleton count is at most $2$, and the Good--Turing
estimator satisfies
\[
    \widehat S_{\mathrm{GT}}\le\frac{2}{n}.
\]
Thus
\[
    \surprise-\EE[\widehat S_{\mathrm{GT}}]
    \geq
    \frac{2}{\ell+2}
    -
    \frac{2n}{(\ell+2)|\Xcal|}
    -
    \frac{2}{n}.
\]
Since we assume the size of the state space to be much larger than $n$, and
$n \geq 2\ell$, we obtain a bias of order $1/\ell$. In
particular, for fixed $\ell$, the Good--Turing estimator exhibits a non-vanishing bias
on this dependent process as $n$ grows.

\section{Classification test error with logistic regression}\label{sec:appendix_lr}

The experiments of Section~\ref{sec:numerical_results} use the $k$-nearest
neighbor rule with $k=3$. We repeat them here with
$L^2$-regularized logistic regression.
We take $d=200$ throughout, and normalize each process so that the
marginal law of $X_i$ is exactly $\mathcal{N}(\mathbf{0},\mathbf{I}_d)$; the labels
are unchanged, $Y_i=\mathrm{sign}(\langle e_1, X_i\rangle)$. The target $\testError$ is computed from $2000$ independent
trainings, and each reported $\MSE$ is averaged over $200$ trajectories
for the $\tau$ sweeps and $150$ for the sweeps over $n$.

\subsection{Moving-average process}\label{sec:appendix_lr_ma}

We now discuss the numerical results for the MA process of Section~\ref{sec:numerical_results} with logistic regression.

\begin{figure}
    \centering
    \subfigure[Our estimator for different choices of $\tau$]{\label{fig:13}\includegraphics[width=7.0cm]{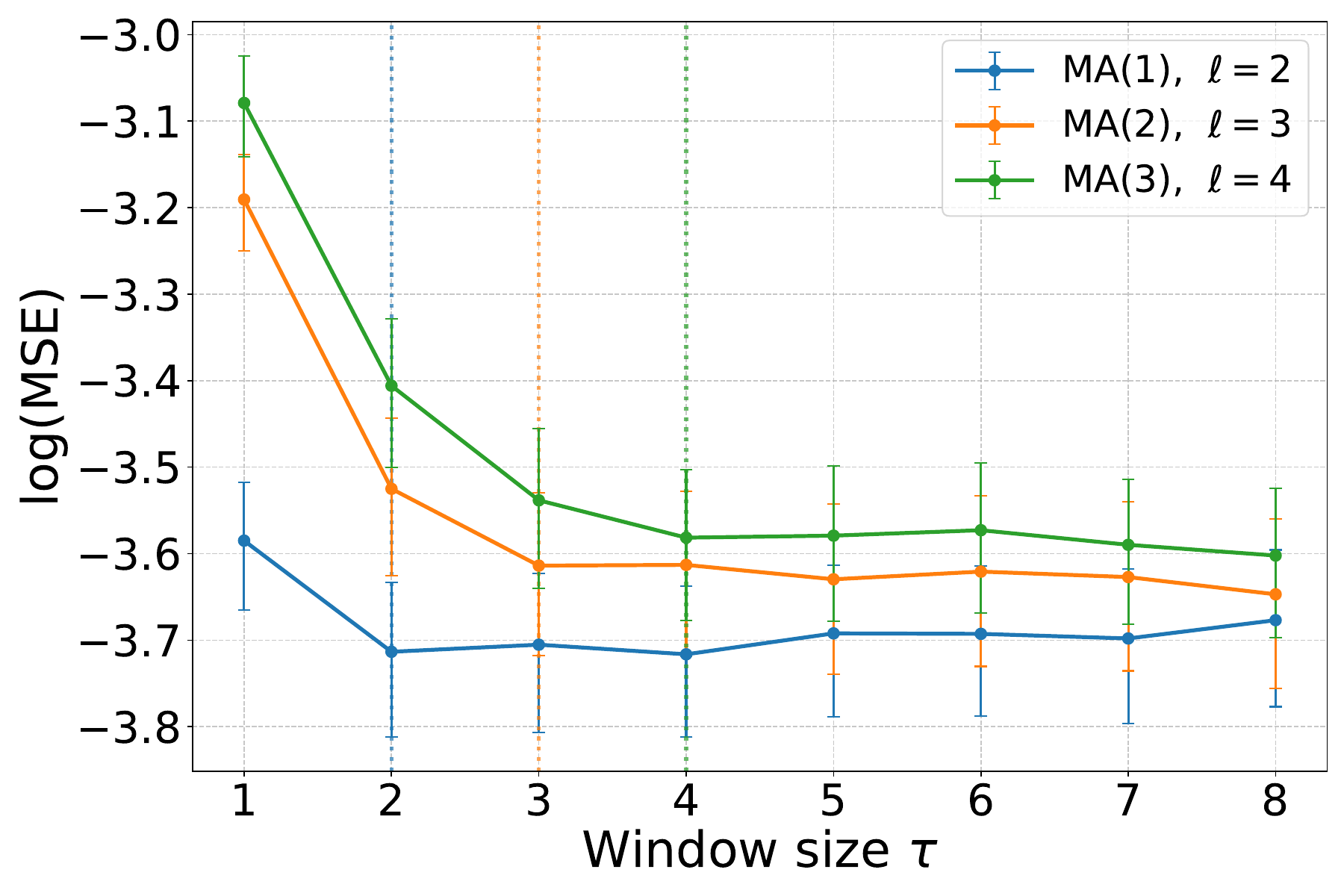}}
    \subfigure[Comparison with baseline]{\label{fig:14}\includegraphics[width=7.0cm]{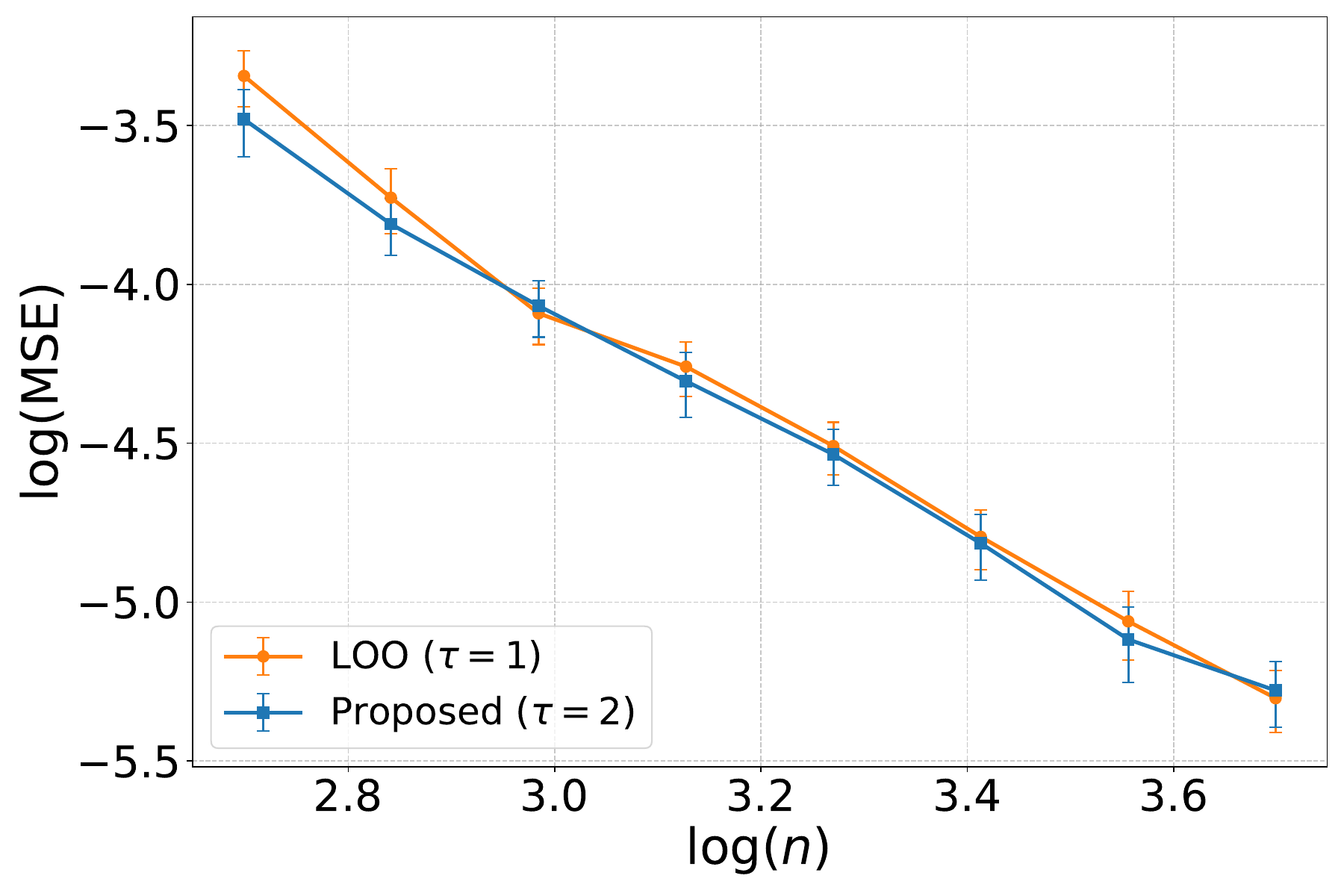}}
    \caption{Classification test error estimation on the MA process with logistic
    regression.
    (a) $\log(\MSE)$ against the window size $\tau$ at fixed $n=600$ and $d=200$, for
    $\ell \in \{2,3,4\}$;
    (b) $\log(\MSE)$ against $n$ for the MA($1$) process ($\ell=2$) with $d=200$, comparing LOO
    ($\tau=1$) with the proposed estimator ($\tau=2$). \vspace{-4mm}}
    \label{fig:ClassificationTE_MA_LR}
\end{figure}

Figure~\ref{fig:13} plots the $\MSE$ of the MA process, for different orders of the process $q \in \{1,2,3\}$, as we vary $\tau \in \{1,2,\ldots,8\}$, with $n=600$ and $d=200$. The dependence parameter varies across
$\ell = q+1 \in \{2,3,4\}$. The plot  shows the same elbow at $\tau=\ell$ as for the $k$-nearest
neighbor rule. The dependence vanishes once the window size $\tau \ge \ell$ since the process is exactly $\ell$-dependent.

Figure~\ref{fig:14} plots the MSE as we vary $n \in \{499,694,965,1341,1863,2589,3598,4999\}$ for the MA($1$) process with $\tau=2$. The plot shows that for logistic regression, the LOO estimator and the proposed estimator overlap. As discussed earlier, this shows that uniform stability helps in reducing the bias even when using the simple plug-in error of the training data points. Thus, Assumption~\ref{assmp:uniform_stability} is sufficient but not necessary.

\subsection{Autoregressive process}\label{sec:appendix_lr_ar}

We now discuss the results for the AR process of Section~\ref{sec:numerical_results} with logistic regression.

\begin{figure}
    \centering
    \subfigure[Our estimator for different choices of $\tau$]{\label{fig:15}\includegraphics[width=7.0cm]{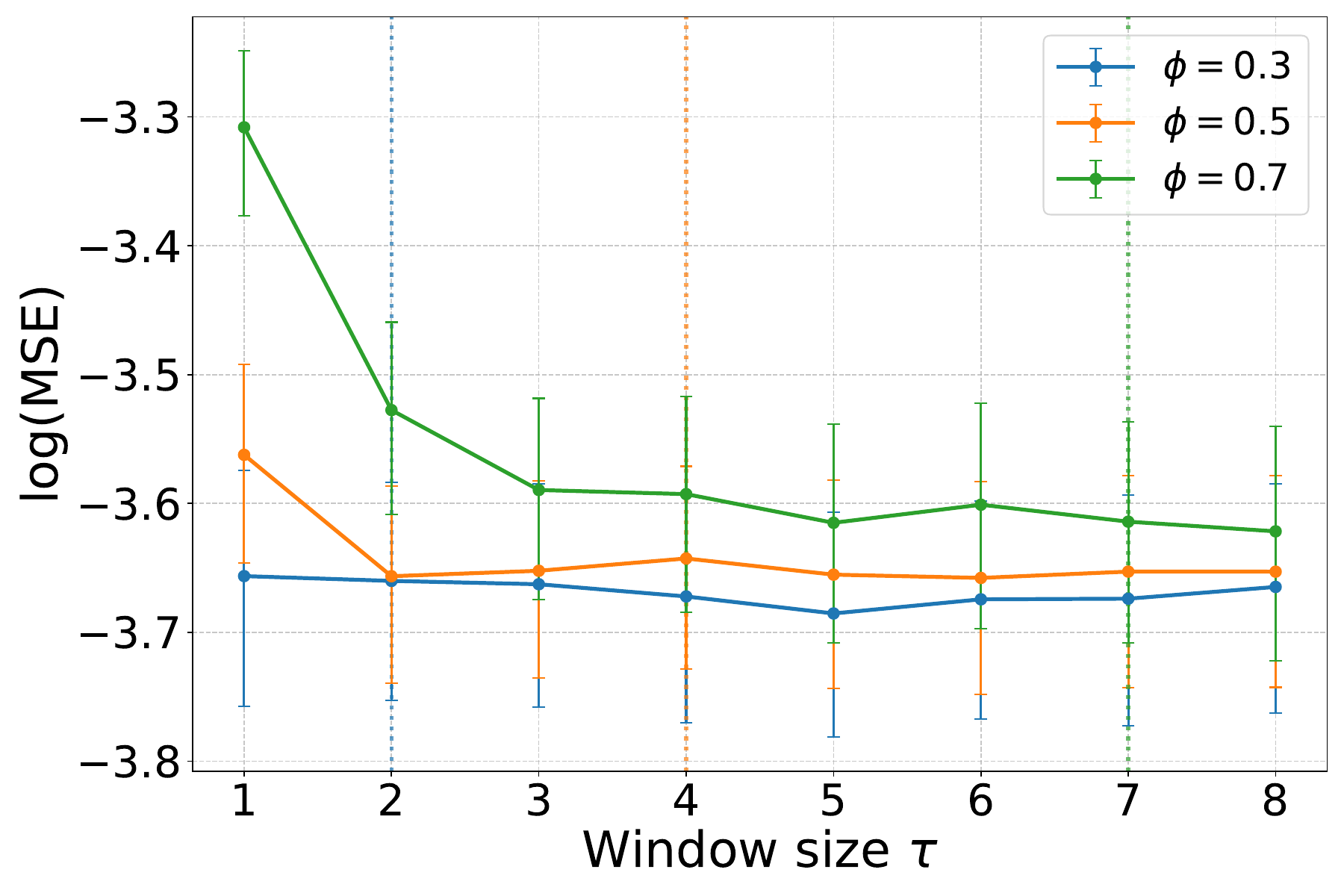}}
    \subfigure[Comparison with baseline]{\label{fig:16}\includegraphics[width=7.0cm]{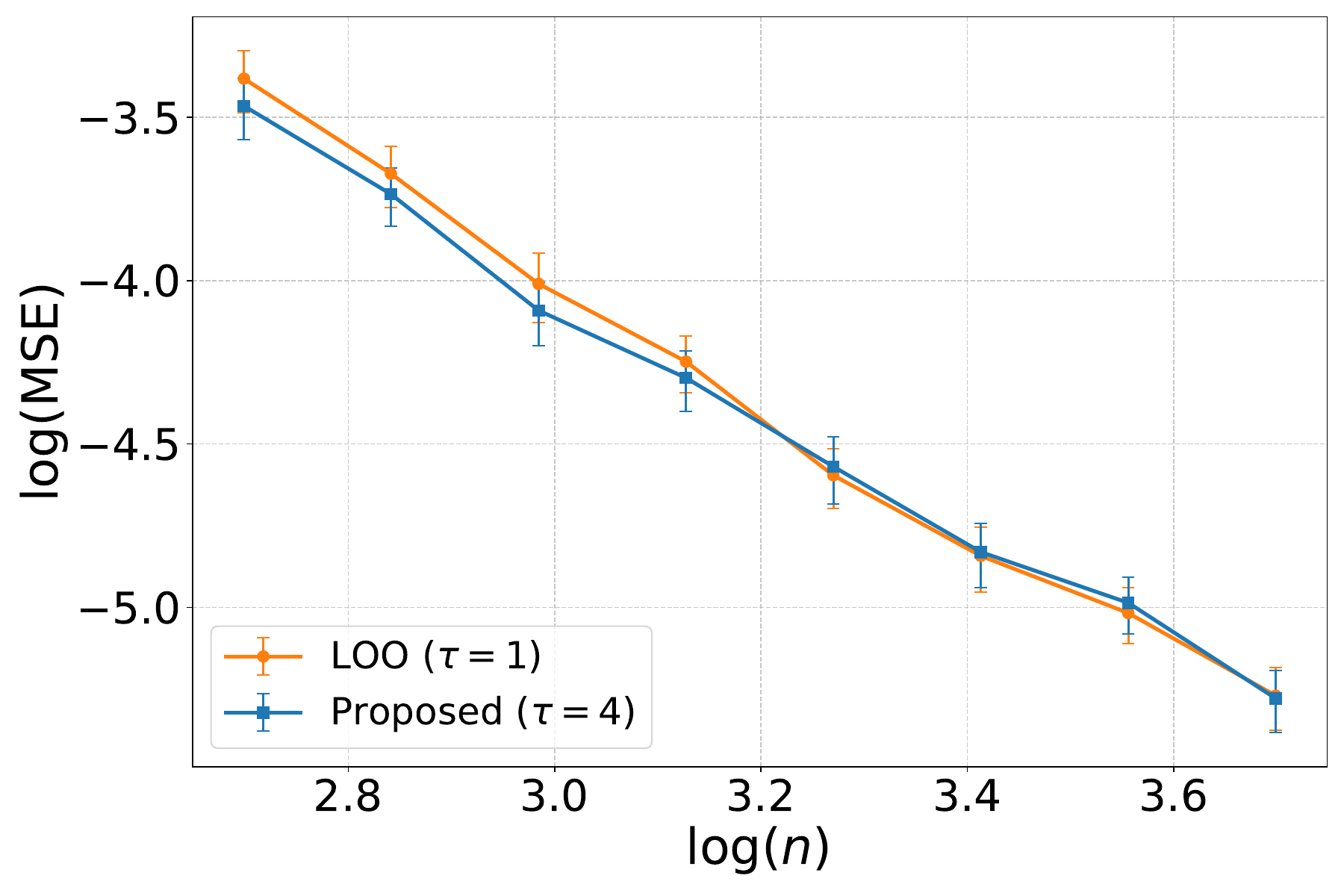}}
    \caption{Classification test error estimation on the AR process with
    logistic regression.
    (a) $\log(\MSE)$ against the window size $\tau$ at fixed $n=600$ and $d=200$, for
    $\phi \in \{0.3,0.5,0.7\}$.
    (b)~$\log(\MSE)$ against $\log n$ for $\phi=0.5$ and $d=200$, comparing LOO ($\tau=1$) with the
    proposed estimator ($\tau=4$).}
    \label{fig:ClassificationTE_AR_LR}
\end{figure}

Figure~\ref{fig:15} plots the $\MSE$ of the AR process, for different values $\phi \in \{0.3,0.5,0.7\}$, as we vary $\tau \in \{1,2,\ldots,8\}$, with $n=600$ and $d=200$. The plot  shows the same elbow as for the $k$-nearest
neighbor rule. The dependence decreases geometrically with the window size but does not exactly vanish beyond a certain window size.

Figure~\ref{fig:16} plots the MSE as we vary $n \in \{499,694,965,1341,1863,2589,3598,4999\}$ for the AR process with $\tau=4$. Again the plot shows that with logistic regression the LOO estimator and the proposed estimator overlap. This is due to uniform stability of regularized logistic regression.

\end{document}